%% file: main.tex
\pdfoutput=1
\pdftrailerid{}\pdfinfoomitdate=1\pdfsuppressptexinfo=-1
\documentclass[10pt,a4paper]{article}
\usepackage[T1]{fontenc}
\usepackage[utf8]{inputenc}
\usepackage{amsmath,amsthm}
\usepackage{newtxtext,newtxmath}
\usepackage{microtype}
\usepackage[margin=2.5cm]{geometry}
\usepackage{booktabs}
\usepackage{array}
\usepackage{dcolumn}
\usepackage{graphicx}
\usepackage{xcolor}
\usepackage[numbers,sort&compress]{natbib}
\usepackage{caption}
\usepackage{enumitem}
\usepackage{float}
\usepackage{placeins}
\usepackage{titlesec}
\usepackage{needspace}
\usepackage{etoolbox}
\usepackage{hyperref}
\definecolor{linkblue}{rgb}{0.10,0.20,0.42}
\hypersetup{colorlinks,linkcolor=linkblue,citecolor=linkblue,urlcolor=linkblue,bookmarksnumbered,
  pdftitle={Evidence or Marginal? What Moment-Vector Propagation Computes in Polynomial Density Neurons},pdfauthor={Przemys\l aw Spyra},
  pdfsubject={Uncertain evidence in polynomial density neurons},
  pdfkeywords={virtual evidence, Jeffrey's rule, Hierarchical Correlation Reconstruction, copula density, orthogonal series, pre-registration}}
\setlist{itemsep=1.5pt,topsep=3pt,parsep=0pt}
\titleformat{\section}{\large\bfseries}{\thesection}{0.7em}{}
\titleformat{\subsection}{\normalsize\bfseries}{\thesubsection}{0.7em}{}
\titleformat{\paragraph}[runin]{\normalsize\bfseries}{}{0pt}{}
\titlespacing*{\section}{0pt}{11pt plus 2pt minus 1pt}{4pt plus 1pt}
\titlespacing*{\subsection}{0pt}{8pt plus 2pt minus 1pt}{3pt plus 1pt}
\titlespacing*{\paragraph}{0pt}{4pt plus 1pt}{0.6em}
\preto\section{\Needspace{5.3\baselineskip}}
\preto\subsection{\Needspace{4.8\baselineskip}}
\graphicspath{{figures/}}
\makeatletter\def\NAT@spacechar{~}\makeatother 
\newcommand{\doi}[1]{doi: \href{https://doi.org/#1}{\nolinkurl{#1}}}

\input{generated/numbers.tex}
\newcommand{\V}[1]{\ifcsname v@#1\endcsname\csname v@#1\endcsname\else\errmessage{unknown number #1}\fi}

\newcommand{\pcd}{p_{\mathrm{CD}}}
\newcommand{\peu}{p_{\mathrm{EU}}}
\newcommand{\pec}{p_{\mathrm{EC}}}
\newcommand{\pg}{p_{\mathrm{G}}}
\newcommand{\dden}{d_{\mathrm{denom}}}
\newcommand{\dneg}{d_{\mathrm{neg}}}
\newcommand{\dmu}{d_{\mathrm{model,uncal}}}
\newcommand{\dmc}{d_{\mathrm{model,cal}}}
\newcommand{\dtot}{d_{\mathrm{total}}}
\newcommand{\ddc}{d_{\mathrm{dc}}}
\newcommand{\rs}{\rho_S}
\newcommand{\Eq}{\mathrm{E}_q}
\newcommand{\Eu}{\mathrm{E}_u}
\newcommand{\EW}{\mathrm{E}_W}
\newcommand{\Var}{\mathrm{Var}}
\newcommand{\Cov}{\mathrm{Cov}}
\newcommand{\sd}{\mathrm{sd}}
\newcommand{\TV}{\mathrm{TV}}
\newcommand{\CV}{\mathrm{CV}_q}
\newcommand{\sobs}{\sigma_{\mathrm{obs}}}
\newcommand{\se}{\sigma_{\mathrm{e}}}
\newcommand{\so}{\sigma_{\mathrm{o}}}
\newcommand{\xe}{\xi_{\mathrm{e}}}
\newcommand{\xo}{\xi_{\mathrm{o}}}
\newcommand{\yo}{y_{\mathrm{o}}}
\newcommand{\ym}{y_{\mathrm{m}}}
\newcommand{\verdict}[1]{\emph{#1}}

\newtheoremstyle{upright}{6pt plus 1pt}{6pt plus 1pt}{\normalfont}{}{\bfseries}{.}{0.5em}{}
\theoremstyle{upright}
\newtheorem{theorem}{Theorem}
\newtheorem{proposition}[theorem]{Proposition}
\newtheorem{corollary}[theorem]{Corollary}

\title{\vspace{-1.2em}Evidence or Marginal?\\[3pt] \large What Moment-Vector Propagation Computes in Polynomial Density Neurons}
\author{Przemys{\l}aw Spyra\\ AGH University of Krakow\\ \texttt{pspyra@agh.edu.pl}}
\date{}

\begin{document}
\maketitle
\vspace{-2.2em}

\begin{abstract}
\noindent When a neuron's input is lost, or known through a belief $q$, what should it output? For a neuron storing a joint polynomial density $N$ with input density $D$, two formulas compete. $\pcd=\Eq[N]/\Eq[D]$ costs one matrix\nobreakdash--vector product; Hierarchical Correlation Reconstruction introduces it as a constant-denominator approximation to $\peu=\Eq[N/D]$. For a density model both are exact answers to different questions: $\pcd$ is Bayesian updating with the belief as virtual evidence; $\peu$ installs the belief as a new marginal (Jeffrey's rule). With a missing input $\pcd$ returns the model's conditional; $\peu$ generally does not. This post hoc reading is proved, and illustrated outside the pair family up to $k=8$ inputs; it has no pre-registered test, only one exploratory real-data pilot. Post hoc, closed forms fall within a factor 2 of all \V{lead.n} measured gaps, and negative density begins near input correlation $1/k$. In a pre-registered synthetic study (up to 32 inputs), \V{hyp.sup} of \V{hyp.total} hypotheses are supported, \V{hyp.inc} inconclusive and \V{hyp.not} not supported, including all three on transfer beyond equicorrelated Gaussians. Post hoc, where $\peu$ is defined, truncation usually outweighs the gap for Beta beliefs in pair neurons; triple neurons under flat and data beliefs are exceptions.
\end{abstract}

\section{Introduction}

A neuron with two inputs loses one of them at test time. If the neuron stores the joint density of its inputs and output, the right output seems obvious: condition on the input that remains. Now write the loss as a belief $q$ over the inputs, a point mass on the first and flat on the second, and pass $q$ through the neuron. There are two natural formulas: average the conditional $N/D$ under $q$, or average numerator and denominator separately and divide. They disagree, and only the second returns the conditional the model implies. Change the story, and let the second input really be uniform and independent of the first in the population the neuron now serves. The same two formulas now swap roles, and the first is right.

This is not a numerical accident. Normalising after averaging looks like a cheap approximation to normalising before. This paper shows that the two are \emph{exact answers to different questions}. In a density model, $\pcd=\Eq[N]/\Eq[D]$ is Bayesian updating with the belief as evidence (Pearl's virtual evidence, one sum-product message), and $\peu=\Eq[N/D]$ installs the belief as a new marginal (Jeffrey's rule). Which is right depends on where the belief came from.

The question arises in density neurons, which store the joint density of inputs and output as coefficients of a polynomial basis. With many inputs the basis keeps only terms that couple at most two variables (a \emph{pair} neuron) or three (a \emph{triple} neuron), and a truncated expansion of this kind can turn negative. Hierarchical Correlation Reconstruction (HCR) is one such model~\citep{duda2024biology,duda2018hierarchical}. Holding a density, not only a function, such a neuron can take a density over its inputs and return one over its output. The natural formula integrates the conditional against the belief, an integral of a ratio of polynomials. HCR's Eq.~10~\citep{duda2024biology} instead averages numerator and denominator separately; the belief then enters only through its moment vector, and propagation is one matrix--vector product. The HCR paper~\citep{duda2024biology} states the integral of the conditional as the density to propagate, and presents this step as an approximation with a constant denominator. We ask what that step computes, how far it lies from the integral, and when either exists.

\begin{figure}[t]
\centering
\includegraphics{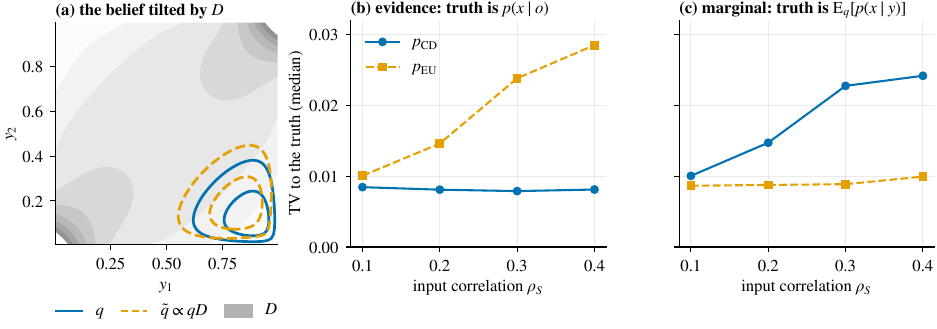}
\caption{Two questions, two exact answers (illustration, not a study result; $k=2$, full tensor model, $m=4$, population coefficients). (a) Under $\pcd$ the belief $q$ is multiplied by the neuron's input density $D$, giving $\tilde q$; the two inputs here report conflicting evidence. (b) If the belief is per-input evidence about an observation $o$, $\pcd$ matches $p(x\mid o)$ up to truncation. (c) If the inputs really follow $q$, $\peu$ matches. Medians over 200 observations at noise level $\sobs=1$; the flat floor near \V{sem.evidence.cd.0.4.1.0} is the truncation error of the degree-4 model.}
\label{fig:idea}
\end{figure}

\paragraph{Contributions.}
\begin{itemize}
\item \textbf{Meaning} (Section~\ref{sec:semantics}). We identify HCR's Eq.~10 as virtual evidence and its target as Jeffrey's rule, and draw the consequence for missing inputs. The mathematics of uncertain evidence is classical~\citep{jeffrey1965logic,pearl1988probabilistic,chan2005revision}, and any model that normalises after averaging has the same structure; what we add is this identification, not a new~update.
\item \textbf{Size} (Section~\ref{sec:gap}). An exact expansion explains why the gap is second order in the pair basis and first order in the triple basis, and closed forms explain why the pre-registered single power law fits only one of its four cells (H1c, \emph{inconclusive}).
\item \textbf{Existence} (Section~\ref{sec:fail}). A positive semidefinite form of the pair denominator gives a convex bound on the negativity onset, an exact corner formula covers the triple basis, a formula for $\Eq[D]$ explains why pair messages survive where triple messages do not, and we give the sample size needed under noise.
\item \textbf{Measurement} (Section~\ref{sec:study}). All \V{hyp.total} pre-registered outcomes are in Table~\ref{tab:hyp}, many of them unfavourable. Post hoc analyses of the same data show when truncation, not the denominator, sets the error, and when it does not (Section~\ref{sec:trunc}).
\end{itemize}

We do not claim that either form is right in general, nor that the contrast of Figure~\ref{fig:idea} persists at large $k$. It appears in full and triple models up to $k=8$, the largest case we tried, and the pair family loses it, because its rows cannot hold the conditional (Section~\ref{sec:illus}). The study's reference and metric names were fixed before the semantic reading was worked out, and its design tests only one of the two semantics.

\section{Setting}\label{sec:setting}

\paragraph{The neuron.} HCR works on variables normalised to $[0,1]$, for instance by their marginal distribution functions, so the object modelled is a copula density~\citep{sklar1959fonctions,nelsen2006introduction}. Let $f_0\equiv 1$ and $f_p(u)=\sqrt{2p+1}\,P_p(2u-1)$ be the orthonormal shifted Legendre polynomials on $[0,1]$, and $f_{\mathbf{j}}(y)=\prod_l f_{j_l}(y_l)$ for a multi-index $\mathbf{j}$. A neuron with output $x$ and inputs $y=(y_1,\dots,y_k)$ models
\begin{equation}\label{eq:neuron}
N(x,y)=\sum_{i=0}^{m} f_i(x)\,A_i(y),\qquad A_i(y)=\sum_{\mathbf{j}} a_{i\mathbf{j}}\,f_{\mathbf{j}}(y),
\end{equation}
with degree at most $m$ per variable and $a_{00}=1$. The coefficients are mixed moments, estimated by $\hat a_{i\mathbf{j}}=\frac1n\sum_t f_i(x_t)f_{\mathbf{j}}(y_t)$~\citep{duda2024biology}. With exact moments of a known ground truth in place of estimates we speak of \emph{population coefficients}. A full tensor is too large for big $k$, so a neuron keeps only multi-indices with few nonzero entries among the $k+1$ variables: at most two (\emph{pair}) or three (\emph{triple}). For $k=32$, $m=4$ this gives \V{coef.pair} and \V{coef.triple} coefficients. Integrating over $x$ removes all rows $i>0$, so $D(y):=\int_0^1 N(x,y)\,dx=A_0(y)$ is the neuron's model of the input density, and $h_x(y)=N(x,y)/D(y)$ its conditional of $x$ given $y$.

\paragraph{Two updates.} Given a belief $q$ on the inputs, define
\begin{equation}\label{eq:exact}
\peu(x)=\Eq\!\left[\frac{N(x,y)}{D(y)}\right],
\qquad
\pcd(x)=\frac{\Eq[N(x,y)]}{\Eq[D(y)]}=\frac{\sum_i f_i(x)\sum_{\mathbf{j}} a_{i\mathbf{j}} b_{\mathbf{j}}}{\sum_{\mathbf{j}} a_{0\mathbf{j}} b_{\mathbf{j}}},\quad b_{\mathbf{j}}=\Eq[f_{\mathbf{j}}(y)].
\end{equation}
$\pcd$ is Eq.~10 of~\citep{duda2024biology}. It sees $q$ only through the moments $b_{\mathbf{j}}$, costs one matrix--vector product, and integrates to $1$ whenever $\Eq[D]\neq0$. $\peu$ needs $N/D$ to be integrable under $q$, which is more than $D\neq0$: at a regular sign change of $D$ inside the support, $\Eq|N/D|=\infty$. For a general belief it needs an integral over $y$. If the density of $q$ lies in the span of the basis, $\peu$ is also linear in the moments, with output coefficients $c_i=\sum_{\mathbf{j}}M_{i\mathbf{j}}b_{\mathbf{j}}$ and $M_{i\mathbf{j}}=\int f_{\mathbf{j}}A_i/D\,dy$, but $M$ is a $k$-dimensional integral that needs $D>0$ on the cube. So the difference in cost lies in obtaining the matrix, not in applying it. The label CD abbreviates \emph{constant denominator}, and EU \emph{exact, uncalibrated}, the protocol's name for this form, where ``exact'' means only that the integral is evaluated without replacing the denominator. We keep it as a label; in the text $\peu$ is the \emph{marginal-preserving} form, for the reason given by Theorem~\ref{thm:sem}.

\paragraph{Negativity and calibration.} A truncated orthogonal expansion of a density need not be nonnegative. Orthogonal-series density estimation has a long history~\citep{cencov1962evaluation,schwartz1967estimation,kronmal1968estimation,efromovich2010orthogonal}, and corrections that turn an estimator into a bona fide density are known~\citep{gajek1986improving,glad2003correction}. The HCR paper proposes a calibration, for example $\max(\rho,0.1)$ or a softplus followed by normalisation, and says that in neural use it is mostly neglected, as the resulting approximations should be compensated during training; likelihood evaluation needs a positive density~\citep[Secs.~II-A, IV]{duda2024biology}. This is separate from the constant denominator. The same paper states that step as an approximation~\citep[Sec.~II-C]{duda2024biology}, and notes that the denominator of its prediction formula can be removed given further inter-layer normalisation~\citep[Sec.~II-B]{duda2024biology}. We keep the two issues apart throughout.

\section{What the two forms compute}\label{sec:semantics}

What does each form compute when $N$ is a genuine density? Each is an exact Bayesian update, for a different reading of $q$. This reading is post hoc: it was worked out after the study was designed and run, and it changes no verdict. Proofs are in Appendix~\ref{app:proofs}. The signed case is treated after the theorem.

The intuition is one line of algebra. Averaging $N=h_xD$ under $q$ and dividing by $\Eq[D]$ averages the conditional $h_x$ under $q$ reweighted by $D$. The denominator has not been approximated away: it has moved into the belief.

\begin{theorem}[What $\pcd$ and $\peu$ compute]\label{thm:sem}
Let $N\geq0$ be a joint density and $q$ a bounded density with $D>0$ on its support, and write $\tilde q=qD/\Eq[D]$.
\begin{enumerate}[label=(\alph*)]
\item \emph{Evidence.} Add a binary variable $E$ with $P(E=1\mid x,y)=q(y)/\sup q$, conditionally independent of $x$ given $y$. Under the model $N$, $p(x\mid E=1)=\pcd(x)$ and $p(y\mid E=1)=\tilde q(y)$.
\item \emph{Marginal.} $q(y)h_x(y)$ is the unique joint density with input marginal $q$ and the model's conditional. Its output marginal is $\peu$.
\item \emph{Same conditional, different belief.} $\pcd=\int h_x\,\tilde q\,dy$ and $\peu=\int h_x\,q\,dy$.
\item \emph{Conversion.} If $\Eq[1/D]<\infty$, then $\pcd[q/D]=\peu[q]$ and $\peu[qD]=\pcd[q]$, with beliefs normalised.
\end{enumerate}
\end{theorem}

Part (a) is Pearl's virtual evidence~\citep{pearl1988probabilistic}: $q$ acts as a likelihood, $D$ as the prior, and $\tilde q$ as the posterior. Its numerator $\Eq[N]=\int N q\,dy$ is the sum-product message~\citep{kschischang2001factor} from the factor $N$ to $x$ when the incoming message on $y$ is $q$. Part (b) is Jeffrey's rule~\citep{jeffrey1965logic,diaconis1982updating}. Both obey probability kinematics, as (c) shows, and differ only in how the evidence is specified~\citep{chan2003revision}. Part (d) is the classical translation between them (Theorem~5 of~\citep{chan2005revision}, as described in~\citep{peng2010uncertain}). So the question ``how accurate is $\pcd$?'' has no answer until one says what $q$ means.

\paragraph{Signed models.} Without $N\geq0$ the identities in (c), (d) remain algebraically true wherever the denominators are nonzero, but $\tilde q$ is a signed measure and the probabilistic reading is lost. This matters from Section~\ref{sec:fail} on.

The next statement says which reading fits three situations a neuron meets in practice.

\begin{proposition}[Evidence channels and missing inputs]\label{prop:chan}
Let the model have uniform one-dimensional input marginals.
\begin{enumerate}[label=(\alph*)]
\item Let observations $o_l$ of the inputs be independent given $y$, with likelihoods $\lambda_l(y_l)$, and let $o$ be independent of $x$ given $y$. Then the product of per-input posteriors $q=\prod_l q_l$, $q_l\propto\lambda_l$, gives $\pcd[q]=p(x\mid o)$. In general $\peu[q]\neq p(x\mid o)$.
\item If instead $q^\ast=p(y\mid o)$ is the joint posterior, then $\peu[q^\ast]=p(x\mid o)$, and $\pcd[q^\ast]$ counts the prior $D$ twice.
\item Let inputs $\yo$ be observed and inputs $\ym$ be missing, with belief a point mass on $\yo$ and flat on $\ym$. This belief is not a bounded density, so here $\Eq$ means integration over $\ym$ at fixed $\yo$; assume $\Eq[D]>0$. Then $\pcd$ is the model's conditional $p(x\mid \yo)$. $\peu=\int p(x\mid \yo,\ym)\,d\ym$ ignores the dependence of $\ym$ on $\yo$ and in general differs from it.
\end{enumerate}
\end{proposition}

Part (c) is the HCR paper's rule for missing variables, which uses only coefficients with zero index for them~\citep[Sec.~II-A]{duda2024biology}. So the HCR paper's stated target for Eq.~10 is the marginal-preserving form, while its missing-data rule is the evidence form, and the same contraction serves both. Part (b) has a consequence for depth: a density passed forward through a chain is a predictive marginal of a hidden variable, which already contains that variable's prior. The matching form at the next layer is then $\peu$, and the study's ground truth takes that view.

None of this needs polynomials. Any model that averages before normalising tilts its belief the same way.

\begin{proposition}[Any model that normalises after averaging]\label{prop:any}
Let $\pi\geq0$ on $\mathcal X\times\mathcal Y$ and $Z(y)=\int \pi\,dx>0$. Then
$\Eq[\pi]/\Eq[Z]=\int (\pi/Z)\,d\tilde q$ with $d\tilde q\propto Z\,dq$. If $\pi$ is integrable, this is virtual evidence in the joint model $p\propto \pi$, whose input density is proportional to $Z$.
\end{proposition}

So a softmax classifier whose exponentiated logits are averaged before normalising implicitly tilts the belief by $Z=\sum_x\exp(\text{logit}_x)$, the unnormalised input density of the energy-based reading of \citet{grathwohl2020classifier}; squared density models~\citep{rudi2021psd,tsuchida2023squared} behave the same way. Linearity of $N$ in the coefficients adds only one thing, the closed-form contraction with the moment vector.

\paragraph{The gap.} How far apart are the two answers? They differ by how much the tilt moves the conditional, which is a covariance. Let $\CV(D)=\sd_q(D)/\Eq[D]$. The first equality below is the classical covariance identity for a ratio of means~\citep{hartley1954unbiased,schochet2022design}, applied pointwise in $x$.

\begin{proposition}[Gap]\label{prop:gap}
Assume $\Eq[D]>0$ and $D\neq0$ $q$-a.s., and let $C_h=\sup_y\int|h_x(y)|\,dx$ over the support of $q$. If $h_x$ is integrable under $q$, then for every $x$,
\begin{equation}\label{eq:identity}
\pcd(x)-\peu(x)=\frac{\Cov_q(h_x,D)}{\Eq[D]}.
\end{equation}
If moreover $h_x$ is square-integrable under $q$,
\begin{equation}\label{eq:tvbound}
\TV(\pcd,\peu)=\frac{\CV(D)}2\int|\mathrm{corr}_q(h_x,D)|\,\sd_q(h_x)\,dx\;\leq\;\frac{\CV(D)}2\Big(\sum_{i\geq1}\Var_q(A_i/D)\Big)^{1/2}.
\end{equation}
Also $\TV(\pcd,\peu)\leq C_h\TV(\tilde q,q)\leq\frac12C_h\CV(D)$, with $C_h=1$ for a nonnegative conditional. The gap vanishes if $x$ and $y$ are independent under the model, or if $D$ is constant on the support of $q$.
\end{proposition}

The bound in $\CV(D)$ alone is the cheap diagnostic, and it is loose: in the study the measured gap $\dden=\TV(\pcd,\peu)$ never exceeded half of $\CV(D)$ over \V{ph.bound.n} evaluations (an evaluation is one seed at one setting; Section~\ref{sec:design}), with median ratio \V{ph.bound.med} (post hoc). Keeping the term in $\sd_q(h_x)$ makes the bound \V{tvb.gain} times tighter in median over \V{tvb.n} population cases, with median ratio \V{tvb.sharp} and maximum \V{tvb.sharpmax}. What remains is the correlation of $h_x$ with $D$, which Section~\ref{sec:gap} computes.

\subsection{Illustration at small \texorpdfstring{$k$}{k}}\label{sec:illus}

Does the contrast show in a real polynomial model? It does where the model's rows can hold the conditional and the belief is not too narrow, and it does not in the pair family. Take a Gaussian ground truth (type SUM, Section~\ref{sec:design}), population coefficients, $m=4$, and correlation below the negativity onset. An observation $o=s+\sobs e$ of the inputs' normal scores defines three problems. In the \emph{evidence} problem the belief is the product of per-input posteriors and the truth is $p(x\mid o)$. In the \emph{posterior} problem the belief is the joint posterior with the same truth. In the \emph{marginal} problem the inputs really follow the product belief, and the truth is $\Eq[p(x\mid y)]$. All truths are closed forms. This illustrates the proved statements and is not a study result.

At $k=2$, $\rs=0.4$, $\sobs=1$ the median TV to the truth is \V{sem.evidence.cd.0.4.1.0} for $\pcd$ against \V{sem.evidence.eu.0.4.1.0} for $\peu$ in the evidence problem, \V{sem.posterior.cd.0.4.1.0} against \V{sem.posterior.eu.0.4.1.0} in the posterior problem, and \V{sem.marginal.cd.0.4.1.0} against \V{sem.marginal.eu.0.4.1.0} in the marginal problem (Figure~\ref{fig:idea}, Table~\ref{tab:sem}). The order flips as Theorem~\ref{thm:sem} says, more weakly at $\sobs=0.5$, where the belief is narrower. With a missing input at the same correlation the medians are \V{sem.miss.cd.0.4} against \V{sem.miss.eu.0.4}.

Beyond $k=2$ the outcome depends on the family (post hoc, all settings below the onset; Tables~\ref{tab:sembig} and~\ref{tab:semflip}). In the triple family at $k=8$, $\sobs=1$ the same flip appears: $\pcd$ is closer in the evidence problem (\V{sem.flip.triple.8.1.0.evidence.cd} against \V{sem.flip.triple.8.1.0.evidence.eu}), and $\peu$ in the posterior and marginal problems. The full tensor and the triple family at $k=3,4$ show it too. In the pair family at $k=4$ and $8$ it is gone: errors of \V{sem.big.lo} to \V{sem.big.hi} swamp the contrast, and $\pcd$ is closer than $\peu$ in most observations of the evidence and posterior problems, and in \V{sem.big.marg.closer.lo} to \V{sem.big.marg.closer.hi} of observations in the marginal problem. The reason is structural. Pair rows $i\geq1$ are additive in $y$, so they cannot carry the conditional, and the contrast is lost to that model error. A narrower belief also shrinks the contrast: at $\sobs=0.5$ the full tensor keeps the flip, while in the triple family at $k=4,8$ the two forms differ by at most \V{sem.flip.weak.hi} on the posterior and marginal problems. So the flip is proved for a model without truncation error, and it is visible in truncated models that keep the conditional, up to $k=8$, the largest case we tried. We make no claim for larger $k$, and the semantic reading has no pre-registered test.

An exploratory pilot on one real data set, run after the study's results were seen, points the same way for the evidence side: $\pcd$ beat $\peu$ with missing and with noisy inputs, while the neuron trailed gradient boosting by a wide margin and unguarded $\pcd$ failed where $\Eq[D]$ was small or negative (Appendix~\ref{app:pilot}). It has no Jeffrey-type arm, so it shows one side of the flip only.

\section{How large the gap is}\label{sec:gap}

How does the gap grow with dependence and with $k$? Second order in the pair basis, first order in the triple basis, with closed forms that fit the measurements without a fitted constant. We compute the covariance of Proposition~\ref{prop:gap} under the flat belief $u$, the case of the pre-registered hypothesis H1c. A flat belief with correlated inputs is out of distribution for the neuron, so it is a test bed here and not a use case. Write $G=D-1$, $V=N-D=\sum_{i\geq1}f_iA_i$, and $\chi_D^2=\Eu[G^2]=\sum_{\mathbf{j}\neq0}a_{0\mathbf{j}}^2$, the squared distance of $D$ from flat. Since $\Eu[D]=1$, $\chi_D=\sd_u(D)$. Expanding $1/D$ around 1 orders the gap by powers of $G$, and the basis decides which powers survive orthogonality.

\begin{proposition}[Flat belief, exact expansion]\label{prop:flat}
If $D>0$ on the cube, with $T_1=\Eu[GV]$ and $\delta_h=\sup_y\TV(h_\cdot(y),1)$,
\begin{equation}\label{eq:flat}
\pcd-\peu=T_1-\Eu[G^2(h_x-1)],\qquad\big|\TV(\pcd,\peu)-\tfrac12\|T_1\|_1\big|\leq\chi_D^2\,\delta_h.
\end{equation}
\begin{enumerate}[label=(\alph*)]
\item \emph{Pair basis, flat input marginals.} Rows $i\geq1$ hold single-input terms and $G$ holds two-input terms, so $T_1\equiv0$ and $\TV(\pcd,\peu)\leq\chi_D^2\delta_h$. Further, $\pcd-\peu=-T_2+\Eu[G^3(h_x-1)]$ with $T_2=\Eu[G^2V]$, and the remainder has $L^1$ norm at most $2\delta_h\|G\|_\infty\chi_D^2$.
\item \emph{Triple basis.} Rows $i\geq1$ hold two-input terms, which meet those of $G$: $T_1=\sum_{i\geq1}f_i\sum_{\mathbf{j}}a_{0\mathbf{j}}a_{i\mathbf{j}}$ over two-input indices $\mathbf{j}$. The gap is first order.
\end{enumerate}
\end{proposition}

Numerically the pair orthogonality holds to \V{orth.pair}; with sampled coefficients single-input terms appear in $G$, but only at noise level. For the equicorrelated Gaussian copula, the leading terms have closed forms.

\begin{corollary}[Closed forms, equicorrelated Gaussian copula]\label{cor:closed}
Let inputs have correlation matrix with off-diagonal $r$, couplings $c_l$ to $x$, and population coefficients. Let $\Gamma_{pq}=\mathrm{E}[f_p(y_1)f_q(y_2)]$ for $p,q\geq1$, $B_{ip}(c)=\mathrm{E}[f_i(x)f_p(y)]$ at correlation $c$, $(\Gamma^{ll'}_{i})_{pq}=\mathrm{E}[f_i(x)f_p(y_l)f_q(y_{l'})]$, and $\omega_p=\int f_p(u)\,f(u)^\top \Gamma^2f(u)\,du$ with $f=(f_1,\dots,f_m)$. Then
\begin{equation}\label{eq:closed}
T_2=(k-1)\sum_{l}\sum_{i,p\geq1}f_i(x)\,B_{ip}(c_l)\,\omega_p\ \ \text{(pair)},\qquad
T_1=\sum_{l<l'}\sum_{i\geq1}f_i(x)\,\langle \Gamma,\Gamma_i^{ll'}\rangle\ \ \text{(triple)}.
\end{equation}
Odd $p$ have $\omega_p=0$, so the sign of $c_l$ drops out of $T_2$, and $T_2$ has no $f_1$ term: at this order the variance of $\pcd$ moves and its mean does not.
\end{corollary}

\begin{figure}[t]
\centering
\includegraphics{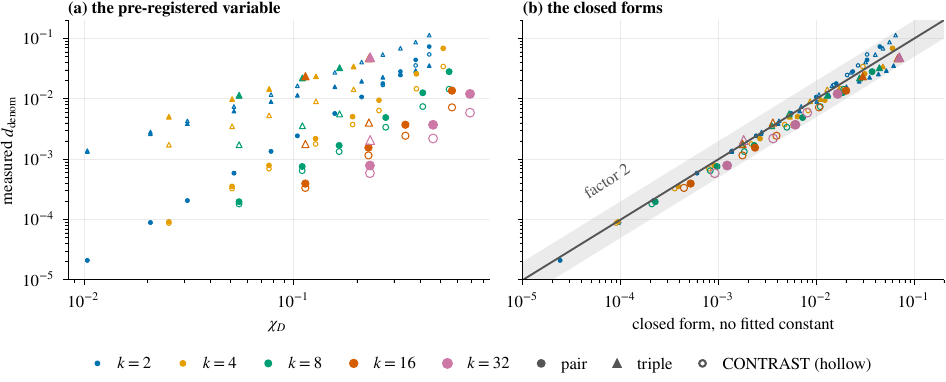}
\caption{Why a single power law in $\chi_D$ fits only one cell, and what replaces it (population coefficients, flat belief S2, $m=4$, the \V{lead.n} points of H1c). (a) Measured $\dden$ against $\chi_D$: the triple SUM cell collapses and the others separate by $k$. (b) The same points against the closed forms of Corollary~\ref{cor:closed}, with no fitted constant. The band is a factor 2. Post hoc.}
\label{fig:scaling}
\end{figure}

These formulas explain the H1c result, which asked whether $\dden$ under the flat belief S2 is one power law of $\chi_D$ in each (type, basis) cell. Reading off the leading orders (a reading, not a proved rate): in the pair basis $T_2\sim k^2\rs^2c^2\propto\chi_D^2c^2$, which carries $c^2$ and hence $k$, so no single function of $\chi_D$ fits. In the triple basis with type SUM, $T_1\sim k^2\rs c^2$ with $c^2$ a function of $k\rs$ over $k$, so where the coupling is not clipped $\dden$ is a function of $k\rs$, and $\chi_D$, of order $k\rs$, collapses the data. This is the one cell supported (share of points within a factor 2: \V{H1c.SUM/triple}). In CONTRAST, where half the inputs have the opposite coupling to $x$ (Section~\ref{sec:design}), the signs give $\sum_{l<l'}s_ls_{l'}=-k/2$, so $T_1\sim k\rs c^2$ and $k$ returns. Figure~\ref{fig:scaling} and Table~\ref{tab:lead} compare $\frac12\|T_2\|_1$ (pair) and $\frac12\|T_1\|_1$ (triple) with the \V{lead.n} points of H1c: with no fitted constant all lie within a factor 2, with ratios \V{lead.lo} to \V{lead.hi}, and \V{lead.small.lo} to \V{lead.small.hi} on the \V{lead.small.n} points with $\chi_D\leq0.2$. This post hoc comparison does not change the verdict of H1c.

The same term $T_1$ reappears under the belief that matters most in practice, the data distribution.

\begin{proposition}[Data belief and the pair structure]\label{prop:data}
Assume uniform marginals and population coefficients, and let $q$ be the true input density. Then $\Eq[D]=1+\chi_D^2$, and the true propagated density is uniform. In the pair basis $\pcd\equiv1$, which is that density. In the triple basis $\pcd-1=T_1/(1+\chi_D^2)$, a real deviation. More generally, in the pair basis for any belief,
\begin{equation}\label{eq:pairstruct}
\pcd(x)=1+\frac{1}{\Eq[D]}\sum_{l}\sum_{i,p\geq1}f_i(x)\,a_{i,(l,p)}\,\mathrm{E}_{q_l}[f_p],
\end{equation}
so a pair neuron sees the belief only through its one-dimensional marginals $q_l$ and the scalar $\Eq[D]$.
\end{proposition}

Thus $T_1$ is both the first-order gap under the flat belief and the whole error under the data belief. On the \V{db.n} triple population points above the Monte Carlo level, the measured distance $\dtot$ of $\pcd$ from the ground truth (Section~\ref{sec:design}), divided by $\frac12\|T_1\|_1/(1+\chi_D^2)$, lies between \V{db.lo} and \V{db.hi}. In the pair basis the denominator reaches the output only as a scalar, which limits what training $D$ can do to $\pcd$.

\section{When the reference and the message fail}\label{sec:fail}

\begin{figure}[t]
\centering
\includegraphics{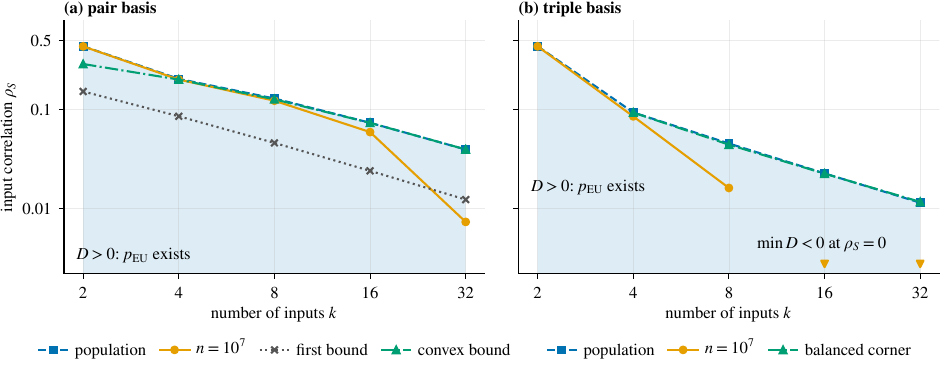}
\caption{Where the marginal-preserving reference exists (type SUM, flat belief S2, $m=4$). Shading: $D>0$ on the cube at population level, below the measured onset. (a) Pair basis with the first and convex lower bounds of Proposition~\ref{prop:pair}(a, b). (b) Triple basis with the balanced corner of Proposition~\ref{prop:corner}, which lies on the population curve. At $n=10^7$ noise moves the onset down, and triangles mark triple cells with $\min D<0$ already at $\rs=0$. Measured onsets are upper bounds (Section~\ref{sec:design}).}
\label{fig:exist}
\end{figure}

Everything above needs $D>0$ on the support of $q$. When does that fail? Figure~\ref{fig:exist} gives the answer: the region where $\peu$ exists shrinks like $\rs\sim1/k$, sooner in the triple basis, and sampling noise removes it sooner still. The reason is that a truncated product of nonnegative factors need not stay nonnegative. We work with the equicorrelated Gaussian copula and population coefficients, then add noise.

Take $r\geq0$ and write $s_l=\sqrt r\,W+\sqrt{1-r}\,\nu_l$ for the normal scores, let $g(w)=\mathrm{E}[f(y)\mid W=w]$ with $f=(f_1,\dots,f_m)$, and let $\xi_l(w)=g(w)^\top f(y_l)$. The untruncated density is $\EW\prod_l(1+\xi_l)$ with all degrees present, which is nonnegative. The neuron keeps
\begin{equation}\label{eq:latent}
D_{\mathrm{pair}}=1+\EW[e_2(\xi)],\qquad D_{\mathrm{triple}}=1+\EW[e_2(\xi)+e_3(\xi)],
\end{equation}
with $e_d$ the elementary symmetric polynomials. Truncating the product in this way gives up positivity.

\begin{proposition}[Pair basis: a positive semidefinite form]\label{prop:pair}
Let $r\geq0$ and $\Gamma=\EW[gg^\top]$, so $0\preceq \Gamma\preceq rI$; let $\varphi(u)=f(u)^\top \Gamma f(u)$, $F=\sum_lf(y_l)$ and $\Theta=m(m+2)=\sup_u|f(u)|^2$. Then
\begin{equation}\label{eq:quad}
D_{\mathrm{pair}}(y)=1+\tfrac12F^\top \Gamma F-\tfrac12\sum_l\varphi(y_l).
\end{equation}
\begin{enumerate}[label=(\alph*)]
\item $\min D\geq1-\frac k2\max_u\varphi(u)$, and $D>0$ on the cube whenever $r<2/(k\Theta)$, that is $\rs<\frac6\pi\arcsin\frac1{k\Theta}$.
\item $\min D\geq\min_\mu\Psi_k(\mu)$ with $\Psi_k(\mu)=1+\frac{k^2}2\bar f_\mu^\top \Gamma\bar f_\mu-\frac k2\int\varphi\,d\mu$, $\bar f_\mu=\int f\,d\mu$, a convex problem over probability measures on $[0,1]$.
\end{enumerate}
\end{proposition}

The form shows what drives negativity: the term $-\frac12\sum_l\varphi(y_l)$ is linear in $k$, and the first term penalises any $y$ whose $f(y_l)$ do not cancel in the directions of $\Gamma$. The bound (b), certified by a Frank--Wolfe duality gap on a grid of $1{,}001$ nodes, is close to the truth: within \V{on.gap4} of the measured onset for $k\geq4$ at $m=4$ (Table~\ref{tab:onset}; all degrees in Table~\ref{tab:onsetall}); it is loose at $k=2$. Its minimiser is not a corner but has interior atoms. The measured product $k\rs^\ast$ rises from \V{on.kr.pair.lo} to \V{on.kr.pair.hi} over $k\geq4$, and beyond the study's range the convex bound's product settles numerically near \V{on.conv.kr.2048} (post hoc). This is a computation, not a limit theorem. The first bound in (a), solved numerically, gives the column ``first'' of Table~\ref{tab:onset}; the explicit condition on $\rs$ in (a) is cruder still. The form needs only conditional independence given a latent variable, so $\Gamma\succeq0$, (b) and the first bound in (a) extend to exchangeable families with such a representation; $\Gamma\preceq rI$ is specific to the Gaussian case.

\begin{table}[t]
\centering
\caption{Negativity onset in $\rs$ with population coefficients, type SUM, $m=4$ (post hoc for the bounds). ``first'': the value of $\rs$ at which the first inequality of Proposition~\ref{prop:pair}(a), $1-\frac k2\max_u\varphi(u)$, reaches zero, solved numerically. ``convex'': the bound of Proposition~\ref{prop:pair}(b). ``corner'': value of Proposition~\ref{prop:corner}. Measured values come from interpolation on the study grid and are upper bounds, so small inversions are possible. Other degrees: Table~\ref{tab:onsetall}.}
\label{tab:onset}
\input{generated/tab_onset.tex}
\end{table}

One might expect the worst point to be a balanced corner of the cube, where half the inputs sit at $1$ and half at $0$. For the pair basis that is false, and for the triple basis it is nearly exact.

\begin{proposition}[Corners]\label{prop:corner}
Let $\xi_+=g^\top f(1)$, $\xi_-=g^\top f(0)$, and let $\kappa_{20}=\EW\xi_+^2$, $\kappa_{11}=\EW\xi_+\xi_-$, $\kappa_{30}=\EW\xi_+^3$, $\kappa_{21}=\EW\xi_+^2\xi_-$. At a corner with $j$ coordinates at $1$,
\begin{align}
D_{\mathrm{pair}}(j)&=1+\big[\tbinom j2+\tbinom{k-j}2\big]\kappa_{20}+j(k-j)\kappa_{11},\label{eq:cornerp}\\
D_{\mathrm{triple}}(j)&=D_{\mathrm{pair}}(j)+\big[\tbinom j3+\tbinom{k-j}3\big]\kappa_{30}+\big[\tbinom j2(k-j)+j\tbinom{k-j}2\big]\kappa_{21}.\label{eq:cornert}
\end{align}
With $\se^2=\frac12(\kappa_{20}+\kappa_{11})\geq0$ and $\so^2=\frac12(\kappa_{20}-\kappa_{11})\geq0$, the energies of the even and odd degrees, $D_{\mathrm{pair}}(j)=1+\frac12[k(k-1)\se^2+((2j-k)^2-k)\so^2]$. Any $\rs$ at which $\min_jD(j)<0$ is at or above the onset.
\end{proposition}

So the balanced corner $j=k/2$ cancels only the odd degrees. Even polynomials have the same sign at $0$ and $1$ and add $\frac12k(k-1)\se^2$: over a grid of $\rs$, all $m$ and even $k$ up to $4{,}096$, the balanced pair corner stays at or above \V{cor.min}, and no pair corner reached negative values in the study's range. The triple term changes this. With $\xi_+=\xe+\xo$ split into parts even and odd in $w$, it adds $\frac k2(\frac k2-1)\big[\frac{2k-2}{3}\EW\xe^3-2\EW\xe\xo^2\big]$, a negative term of the same order in $kr$ as the others. For $m\geq4$ the balanced corner then matches the measured triple onset within \V{on.tri.gap} (Table~\ref{tab:onset}); at $m=2$ no triple corner goes negative, so the formula gives no bound there.

\paragraph{The message.} $\pcd$ needs only $\Eq[D]\neq0$, a weaker condition than $D>0$ on the support, because averaging can cancel negative lobes. Proposition~\ref{prop:mass} says when it does.

\begin{proposition}[Mass of the message]\label{prop:mass}
Let $q$ be a product belief with $b_l=\mathrm{E}_{q_l}[f]$, and $\bar\xi_l(w)=g(w)^\top b_l$. Then $\Eq[D]=1+\EW[e_2(\bar\xi)]$ (pair) or $1+\EW[e_2(\bar\xi)+e_3(\bar\xi)]$ (triple). In the pair basis
\begin{equation}\label{eq:masspair}
\Eq[D]=1+\tfrac12\big(\textstyle\sum_lb_l\big)^\top \Gamma\big(\sum_lb_l\big)-\tfrac12\sum_lb_l^\top \Gamma b_l\;\geq\;1-\tfrac r2\sum_l|b_l|^2,
\end{equation}
and $\Eq[D]\geq1$ if all $q_l$ are equal. In the triple basis with equal $q_l$, $\Eq[D]=1+\binom k2\EW\bar\xi^2+\binom k3\EW\bar\xi^3$.
\end{proposition}

The formula matches \V{eqd.n} sampled population evaluations within \V{eqd.rel}, and it explains an asymmetry in the study. For a central belief the degree-2 moment is negative and $g_2$ is right-skewed, so $\EW\bar\xi^3<0$ and the cubic term drives $\Eq[D]$ negative at large $k$ (this last step is a heuristic reading). In the study $\Eq[D]\leq0$ occurred in \V{ph.eqd.nonpos} of \V{ph.eqd.rows} evaluations, all triple, at $k\geq\V{ph.eqd.mink}$, under Beta beliefs; with population coefficients it occurred too (\V{ph.eqd.pop.nonpos} of \V{ph.eqd.pop.rows}), so it is structural. The pair basis stayed at $\Eq[D]\geq\V{ph.eqd.pairmin}$ (population minimum \V{eqd.s3c} under the conflicting Beta belief S3c of Section~\ref{sec:design}). Pair safety is not universal: a point belief on a negative value of $D$ gives $\Eq[D]<0$ in any basis, and the bound in~\eqref{eq:masspair} is vacuous for narrow beliefs. The statement ``$\pcd$ stays defined past the onset'' holds for pair neurons with the beliefs tested, and not for triple neurons.

\paragraph{Noise.} Estimated coefficients move all of this, and the sample size needed for a positive $\hat D$ grows polynomially in $k$.

\begin{proposition}[Sampling noise at independence]\label{prop:noise}
Let the inputs be independent and uniform. Then $\mathrm{E}\hat D(y)=1$ and $\Var\hat D(y)=K(y)/n$ with $K(y)=\sum_{\mathbf{j}\neq0}f_{\mathbf{j}}(y)^2$, maximal at corners, where $K_{\mathrm c}=k\Theta+\binom k2\Theta^2$ (pair) or $K_{\mathrm c}=k\Theta+\binom k2\Theta^2+\binom k3\Theta^3$ (triple). In the Gaussian limit of $\sqrt n(\hat a_{0\mathbf{j}})$ at fixed $k$, write $1-\min\hat D=z\sqrt{K_{\mathrm c}/n}$. For fixed $m$, $\mathrm{E}z\leq C\sqrt{km}$, and for even $m$, $\mathrm{E}z\geq c\sqrt k$, for absolute constants $c,C>0$ (with $c$ valid for $k\geq k_0$, where $k_0$ depends on $m$); over corners $\mathrm{E}z\leq\sqrt{2k\ln2}$, and $z$ concentrates around its mean with $O(1)$ fluctuations. Hence, for even $m$, $\hat D>0$ needs $n\asymp z^2K_{\mathrm c}$, of order $k^3$ (pair) and $k^4$ (triple) up to factors in $m$.
\end{proposition}

This turns the rule of thumb $n\gtrsim z^2K$ into a statement on how $z$ grows. In the study $z$ grows from \V{ph.noise.zmin} to \V{ph.noise.zmax} (Table~\ref{tab:noisez}). In a Gaussian surrogate, a search over the cube at $k=32$ reaches \V{nz.pair32.cube} and over corners \V{nz.pair32.corner}, against \V{nz.pair32.meas} measured, consistent with $\min D$ being an upper bound. With the cube values, $z^2K_{\mathrm c}$ is \V{nz.need.pair32} for pair at $k=32$ and \V{nz.need.triple16} for triple at $k=16$, either side of the study's $n=10^7$. In the pair basis the noise minimisers put \V{nz.bound.lo}--\V{nz.bound.hi} of coordinates on the boundary, while the population pair minimum has interior atoms.

\section{A reliability atlas: the pre-registered study}\label{sec:study}

The theory above came after the data. The study itself asked plainer questions, fixed in advance: how large is the gap, where does the reference vanish, do the numbers transfer, and do depth and training change them? It cannot test the thesis of Section~\ref{sec:semantics}: its truth is the Jeffrey answer, so it measures distance to one of the two answers and never asks which form suits which belief. What it can give is an atlas of polynomial neurons, showing when the reference exists and what limits accuracy. Table~\ref{tab:hyp} gives every answer. In short: the reference vanishes early, as predicted; thresholds do not transfer; depth and training did not show the expected effects; and, post hoc, where $\peu$ is defined, in pair neurons under Beta beliefs, the gap is rarely the largest error.

\begin{table}[t]
\centering
\caption{All pre-registered outcomes, with the verdict words of the protocol. H2b has two pre-registered evaluations, so \V{hyp.total} hypotheses occupy \V{hyp.rows} lines. Fractions count significant cells among evaluable cells; ``opposite'' counts cells whose confidence interval lies entirely on the unpredicted side. H1c cells are shares of points within a factor 2, and its overall label is discussed in Section~\ref{sec:gapres}. H4b is also test M13 of the validation suite (Appendix~\ref{app:validation}). $D$ does not depend on the output coupling, so SUM and CONTRAST population cells coincide for negativity: the \V{H1a.n} cells of H1a are \V{H1a.distinct} distinct ones, the \V{H2a.n} of H2a are \V{H2a.distinct}, and the \V{H2b_population.n} of H2b (population) are \V{H2b_population.distinct}.}
\label{tab:hyp}
\input{generated/tab_hyp.tex}
\end{table}

\subsection{Design}\label{sec:design}

The design, hypotheses, decision rules, grids and budget rule were set out in a written protocol, fixed before any main run, with no recorded deviations. It was not entered in a public registry; its timing rests on the code lineage of the runs and on scheduler records, not on dated entries. A throughput run before that point, with its own seed, is excluded from all results. Analyses marked \emph{post hoc} were not pre-registered and enter no verdict. The design tests the marginal reading only; Sections~\ref{sec:semantics}--\ref{sec:fail} came later. Full definitions are in Appendix~\ref{app:defs}.

\paragraph{Ground truth.} Output and inputs are uniform; dependence comes from a $(k+1)$-dimensional copula. The Gaussian sweep has equicorrelated inputs with correlation $r=2\sin(\pi\rs/6)$, so $\rs$ is the Spearman correlation between inputs. In \textbf{SUM} each input has correlation $c=\min(0.5,\,0.8\sqrt{(1+(k-1)r)/k})$ with $x$. In \textbf{CONTRAST} half the inputs have $+c$ and half $-c$, with $c=\min(0.5,\,0.8\sqrt{(1-r)/k})$. Other experiments use $t$, Clayton and Gumbel copulas and random LKJ correlation matrices.

\paragraph{Beliefs.} Beliefs are product distributions unless stated. \textbf{S2} is flat. \textbf{S3a} is Beta$(3,3)$ per input, \textbf{S3b} is Beta$(4,2)$, \textbf{S3c} alternates Beta$(4,2)$ and Beta$(2,4)$ (conflicting evidence), and \textbf{S3d} is the narrow Beta$(6,6)$. \textbf{S4} is the true input copula (data belief). \textbf{S1}, the neuron's own $D$, is computed but not primary. The primary beliefs are S2, S3c and S4. S2 is not the data distribution, so part of its error against the ground truth is mismatch between prior and data.

\paragraph{Compared densities.} All four densities are computed from the same $M=\V{cfg.M}$ Monte Carlo samples of the belief, in $R=\V{cfg.R}$ replicate blocks: $\pcd$; $\peu$, \emph{defined} in the protocol's sense, that is $D\geq\V{cfg.floor}$ on the support cut $\{q\geq10^{-6}\max q\}$, which for Beta beliefs is weaker than mathematical definedness (Appendix~\ref{app:defs}); $\pec$, the conditional calibrated at each value of $y$ as $\max(N,\varepsilon)$ divided by its integral over $x$ ($\varepsilon\in\{0.1,0.01,0.001\}$, primary $0.01$); and the ground truth $\pg=\Eq[p(x\mid y)]$, the analytic conditional averaged over $q$. So the truth is the Jeffrey answer, under which $\peu$ is the matching form: all ``errors'' below carry that semantics. A joint normalising constant cancels in a conditional, so $\pec$ is also the conditional of the jointly clipped density.

\paragraph{Outcomes.} Outcomes are total-variation distances $\TV(p,p')=\frac12\int|p-p'|$ on raw, never renormalised densities:
\begin{align*}
\dden&=\TV(\pcd,\peu), & \dneg&=\TV(\peu,\pec), & \dmu&=\TV(\peu,\pg),\\
\dmc&=\TV(\pec,\pg), & \dtot&=\TV(\pcd,\pg), & \ddc&=\TV(\pcd,\pec).
\end{align*}
Densities can be signed, so TV here is an $L^1$ distance with a factor $\frac12$ and can exceed $1$. We also record variance and mean errors, $P(D<0)$ under uniform $y$, $\chi_D$, and $\min D$ from a multi-start projected optimiser with $\V{cfg.starts}$ starts. This $\min D$ is an upper bound on the true minimum, so measured onsets are conservative: true onsets can only be earlier.\looseness=-1 

\paragraph{Thresholds and verdicts.} A \emph{cell} is a combination of design factors (such as type, basis and $k$), a \emph{point} adds a value of $\rs$, and an \emph{evaluation} is one seed at one point; each seed re-estimates all coefficients. A threshold is the first crossing of a level by the seed-median curve, over the run of defined points from $\rs=0$. If the curve becomes undefined first, or the level is not reached by $\rs=0.9$, the entry is ``---''. The negativity onset is the first zero crossing of seed-median $\min D$, and $0$ if negative at $\rs=0$. Confidence intervals are percentile bootstraps over seeds~\citep{efron1979bootstrap,efron1993introduction}; they reflect estimation noise only, so very tight ones do not mean that the quantity is known beyond this design. A hypothesis over cells is \emph{supported} if at least $2/3$ of evaluable cells are significant in the predicted direction, \emph{not supported} if at least $2/3$ are not, and \emph{inconclusive} otherwise or with fewer than three evaluable cells. Some hypotheses have their own rule (Appendix~\ref{app:hyp}). ``Significant'' is used only in this pre-registered sense.

\paragraph{Experiments.} The Gaussian sweep crosses $k\in\{2,4,8,16,32\}$, both types, both bases, $m\in\{2,4,6\}$, $n=10^7$, 50 seeds and 19 values of $\rs$ from 0 to 0.9. The population sweep repeats these cells with exact coefficients. The convergence study doubles $M$, $n$ and two grid sizes on a subset. The copula-family and LKJ experiments ($k\leq8$) vary the ground truth. The chain experiment builds up to $L=4$ neurons, a neuron of the sweep followed by neurons with two independent inputs, on a Gaussian Markov ground truth with correlation $\alpha=\V{cfg.a}$ per layer, trained under three losses and two optimisers at fixed, untuned settings (Appendix~\ref{app:chains}). Altogether \V{jobs} runs finished with \V{failed} failures on NVIDIA A100-SXM4-40GB GPUs in float64, with TF32 disabled and deterministic algorithms, using \V{gpuh.jobs} GPU-hours of run time (\V{gpuh.billed} GPU-hours billed for everything, including validation and benchmarks).

\subsection{In pair neurons truncation, not the denominator, usually sets the error (post hoc)}\label{sec:trunc}

We begin with a post hoc reading of metrics that were all pre-registered, because it answers the practical question: would replacing $\pcd$ with $\peu$ help? Mostly not. Table~\ref{tab:trunc} sets the gap $\dden$ against the error of the truncated model $\dmu=\TV(\peu,\pg)$, where both exist, under the semantics that favours $\peu$. In the pair basis the model error exceeds the gap by a median factor of \V{tr.S4.pair.ratio} (S4) to \V{tr.S3d.pair.ratio} (S3d), and both forms sit at nearly the same distance from the truth, for example \V{tr.S3a.pair.cd} and \V{tr.S3a.pair.eu} under S3a. $\peu$ is closer in \V{tr.S3c.pair.euc} (S3c) to \V{tr.S3b.pair.euc} (S3b) of evaluations. A pair model with few coefficients cannot represent the conditional well enough for the difference to count.

There are two exceptions, both explained by Section~\ref{sec:gap}. Triple neurons under S2 and S4 have model error far below the gap, with ratios \V{tr.S2.triple.ratio} and \V{tr.S4.triple.ratio}, and $\peu$ closer in \V{tr.S2.triple.euc} of evaluations: here $T_1$ is first order and it is the whole error of $\pcd$. Under the Beta beliefs the triple ratio is \V{tr.S3b.triple.ratio} to \V{tr.S3d.triple.ratio}. The illustration of Section~\ref{sec:illus} sends the same message for the pair family with the evidence truth at $k=4,8$, where model error of \V{sem.big.lo} to \V{sem.big.hi} hides the choice of form. These results are for $m=4$, $\varepsilon=0.01$, $\rs>0$ and defined evaluations only, and the filter that defines them also caps the gap (Section~\ref{sec:gapres}), so they say nothing about the region past the onset, where $\peu$ does not exist. They concern synthetic ground truths, where the truncated part of the density is known, and a task could weigh the error differently.\looseness=-1 

\begin{table}[t]
\centering
\caption{Gap against truncation error, on defined evaluations with $\rs>0$, $m=4$, $\varepsilon=0.01$; truth is $\pg$, so $\peu$ is the matching form (post hoc). Distances are medians. ``ratio'': median of $\dmu/\dden$ per evaluation. ``EU closer'': share with $\peu$ nearer than $\pcd$ to $\pg$. ``EC beats EU'': share with $\pec$ nearer than $\peu$.}
\label{tab:trunc}
\input{generated/tab_trunc.tex}
\end{table}

\paragraph{A denominator-free route to the Jeffrey answer (post hoc).} If the Jeffrey answer is what is wanted, neither form need be used. Regressing $f_i(x)$ on features $\phi(y)$ gives a matrix $B$, and $c=B^\top\Eq[\phi]$ is a message with the same run-time form as $\pcd$, with no denominator, whose target is the conditional averaged under $q$. Table~\ref{tab:reg} compares it with both forms against $\pg$ over \V{reg.n} settings. Additive features do no better than $\pcd$. Pair features do much better, and fall below both forms in \V{reg.share} of settings, past the onset too. The limits are plain: it needs a feature Gram matrix, that is four-way moments the neuron does not store; it has no positivity guarantee; and it gives the wrong semantics for missing inputs, because it averages a conditional instead of conditioning.

\begin{table}[t]
\centering
\caption{Regression message against both forms (post hoc, not part of the study): median TV to $\pg$ over $\rs\in\{0.1,0.3,0.5,0.7\}$ and S2, S3a, S3c, S3d; type SUM, $m=4$, coefficients from $n=\V{reg.logn}$ samples. $\peu$ drops belief samples with $D<\V{cfg.floor}$.}
\label{tab:reg}
\input{generated/tab_reg.tex}
\end{table}

\subsection{The gap looks modest only where it can be measured}\label{sec:gapres}

Under S2 the $1\,\%$ threshold of $\dden$ falls from \V{thr.SUM.pair.2.d1} at $k=2$ to \V{thr.SUM.pair.16.d1} at $k=16$ (pair, Table~\ref{tab:thr}, Figure~\ref{fig:d}), while at $k=32$ the entry is empty. Empty does not mean small: $\peu$ became undefined (from $\rs=\V{thr.SUM.pair.32.und}$) before the curve reached $1\,\%$. In the triple basis at $k\geq16$ the reference is undefined already at $\rs=0$.

Selection explains this (post hoc). The filter that defines $\dden$ almost implies a modest gap: at $m=4$, $\rs>0$ within it $\CV(D)$ never exceeds \V{sel.cv}, so Proposition~\ref{prop:gap} caps the gap, and $\dden$ never exceeds \V{sel.dmax}. ``Modest where measured'' should therefore not be read as ``modest'': under S2 at $k=32$ only \V{sel.k32.pair} of pair evaluations survive the filter (more in the caption of Table~\ref{tab:bel}). Past the onset the outcomes are $\dtot$ and $\ddc$; in the pair basis median $\dtot$ is \V{past.pair.lo} to \V{past.pair.hi}, apart from S4 (Table~\ref{tab:past}, post hoc). Under the narrow belief S3d, at $k=16,32$, pair $\peu$ stays defined under the protocol's support cut up to $\rs=\V{ph.narrow.k32.rho}$ and $\dden$ never reaches $1\,\%$ (post hoc). That describes the cut and not an integral over the cube: a Beta density is positive on the open cube, so wherever $D$ changes sign $\Eq|N/D|$ is infinite (Appendix~\ref{app:defs}). Width is not everything: the off-centre S3b reaches $1\,\%$ earlier than S2 at $k\geq8$ for type SUM (Table~\ref{tab:bel}).

\textbf{H1c (\verdict{\V{H1c.verdict}})} asked for one power law in $\chi_D$. The share of points within a factor 2 was \V{H1c.SUM/pair} (SUM/pair), \V{H1c.CONTRAST/pair} (CONTRAST/pair), \V{H1c.SUM/triple} (SUM/triple) and \V{H1c.CONTRAST/triple} (CONTRAST/triple), which by the pre-registered rule makes these cells respectively \verdict{\V{H1c.verdict.SUM/pair}}, \verdict{\V{H1c.verdict.CONTRAST/pair}}, \verdict{\V{H1c.verdict.SUM/triple}} and \verdict{\V{H1c.verdict.CONTRAST/triple}}. The protocol gives this rule per (type, basis) and does not say how the cells combine. The overall label comes from the analysis code fixed with the protocol, which requires all cells to agree. Counting a \emph{supported} cell as significant under the general aggregation rule would instead give \V{H1c.cells.sup} of \V{H1c.cells}, that is \verdict{not supported}. We keep the recorded verdict; the cell pattern is what Corollary~\ref{cor:closed} explains. \textbf{H2c (\verdict{\V{H2c.verdict}}).} $\dden$ varied by at most a factor $1.5$ across $m$ at \V{H2c.share} of \V{H2c.n} points, between the rule's $80\,\%$ and $50\,\%$.

What the gap moves is mostly spread (post hoc). Among \V{ph.mv.n} points of the pair basis the relative variance error exceeds the standardised mean error at \V{ph.mv.share}. Under S2 the mean is almost untouched (at most \V{ph.mv.s2.mean}) while the variance moves by up to \V{ph.mv.s2.var}, as Corollary~\ref{cor:closed} says (Figure~\ref{fig:mv}).

\subsection{The reference vanishes early, and noise brings it forward}\label{sec:onset}

Here the expectations held. \textbf{H1a, H1b (\verdict{\V{H1a.verdict}}, \verdict{\V{H1b.verdict}}).} The onset falls strictly with $k$ in all \V{H1a.n} population cells, and at $n=10^7$ the difference between $k=32$ and $k=2$ has a confidence interval below zero in all \V{H1b.n} cells. For SUM in the pair basis it falls from \V{H1b.sp.k2} to \V{H1b.sp.k32}, a difference with $95\,\%$ CI $[\V{H1b.sp.lo},\V{H1b.sp.hi}]$. \textbf{H2a (\verdict{\V{H2a.verdict}}).} A higher degree gives an earlier onset in \V{H2a.sig} of \V{H2a.n} population cells (Table~\ref{tab:onsetm}). \textbf{H2b (\verdict{\V{H2b_population.verdict}} with population coefficients, \verdict{\V{H2b_gauss.verdict}} at $n=10^7$).} The triple basis is earlier than pair in \V{H2b_population.sig} of \V{H2b_population.n} and \V{H2b_gauss.sig} of \V{H2b_gauss.n} cells. These outcomes are what one would expect as terms are added, and Propositions~\ref{prop:pair} and~\ref{prop:corner} now give their mechanism.

Noise moves the onset at large $k$: \V{thr.SUM.pair.32.on} against \V{thr.SUM.pair.32.pop} (pair, $k=32$), and \V{thr.SUM.triple.16.on} against \V{thr.SUM.triple.16.pop} (triple, $k=16$), in line with Proposition~\ref{prop:noise}. Moving the floor on $D$ from $0$ to $0.05$ changes the share of S2 evaluations with $\min D$ above it by at most \V{floor.shift} of evaluations in any cell (post hoc reanalysis), so the pre-registered floor decides little.

\subsection{Estimation, not integration, limits the numbers}\label{sec:conv}

\textbf{H4a (\verdict{\V{H4a.verdict}}).} The protocol states H4a for the defined headline points at $m=4$, $\varepsilon=0.01$; the analysis code evaluates it on the pair basis at $k\in\{2,8,32\}$ under S2, S3c and S4. There the across-seed standard deviation of $\dden$ exceeded the Monte Carlo standard error at \V{H4a.share} of \V{H4a.n} points (Table~\ref{tab:hyp}); over all headline cells it did so at \V{ph.H4a.all.share} of \V{ph.H4a.all.n} (post hoc), so the verdict does not depend on this choice. \textbf{H4b (\verdict{\V{H4b.verdict}}).} Doubling $M$, $n$, or either grid must change every reported TV by less than $\max(3\,\mathrm{SE},10^{-3})$. There were \V{H4b.viol} violations among \V{H4b.total} paired values, all under $n\times2$ (Table~\ref{tab:conv}). Numerical integration is converged and, at large $k$, the estimator is not: \V{ph.h4b.triple} of violations are triple (post hoc). Such high-$k$ triple cells are stress tests: they stay in the tables, but say little about usable neurons.

\subsection{Clipping changes outputs more, and usually for the worse (post hoc)}\label{sec:clip}

Where $\dden$ and $\dneg$ both exist and at least one is above the noise floor (\V{ph.dvn.n} points), $\dneg>\dden$ at \V{ph.dvn.0.01} of points (\V{ph.dvn.0.1} for $\varepsilon=0.1$, \V{ph.dvn.0.001} for $\varepsilon=0.001$; post hoc). But $\dneg$ is movement, not harm. Against the truth, $\pec$ beats $\peu$ in only \V{clip.min} to \V{clip.max} of defined evaluations across beliefs and bases (Table~\ref{tab:trunc}), and under S4 it raises the pair distance from \V{clip.S4.eu} to \V{clip.S4.ec}. $\dneg$ under S2 barely depends on $\varepsilon$ (\V{clip.dneg.0.001}, \V{clip.dneg.0.01}, \V{clip.dneg.0.1} for $\varepsilon=0.001,0.01,0.1$), so it mostly reflects removing negative lobes and renormalising. Clipping is one calibration among several, and this says nothing about others.

Negativity removes $\peu$; it does not remove pair $\pcd$. Past the onset, pair $\pcd$ is closer to $\pg$ than $\pec$ in \V{ph.closer.undef.pair} of evaluations (\V{ph.closer.undef} pooled over both bases), with a strong dependence on the belief (Table~\ref{tab:past}). It does hurt triple $\pcd$ at $k\geq8$, where $\Eq[D]$ can be nonpositive (Proposition~\ref{prop:mass}), and $\dtot>1$ in \V{ph.tv1.n} evaluations (\V{ph.tv1.share}), \V{ph.tv1.triple} of them triple, with a largest value of \V{ph.tv1.max}: the output is then a strongly signed function, not a density. Under S2, $\pcd$ is the marginal and $\dtot$ is almost entirely mismatch between flat prior and truth (median gap to $\TV(\pg,\text{uniform})$ \V{ph.s2.gap}, maximum \V{ph.s2.maxgap}); Figure~\ref{fig:dec} shows the components. Under S4 the data bear out Proposition~\ref{prop:data}: pair median $\dtot=\V{ph.s4.pair}$, maximum \V{ph.s4.pairmax}; triple median \V{ph.s4.triple}, maximum \V{ph.s4.triplemax}.

\subsection{Transfer, chains and training}\label{sec:transfer}\label{sec:depth}

\paragraph{Transfer.} None of the three transfer hypotheses was supported. For \textbf{H3a (\verdict{\V{H3a.verdict}})}, \V{H3a.in} of \V{H3a.n} onset differences from the Gaussian had a CI inside $\pm0.05$ and \V{H3a.out} lay entirely outside. For \textbf{H3b (\verdict{\V{H3b.verdict}})}, $\dden$ at matched Kendall's $\tau$ was within a factor 2 of the Gaussian at \V{H3b.in} of \V{H3b.n} points. For \textbf{H3c (\verdict{\V{H3c.verdict}})}, the Gaussian power law held for \V{H3c.pair} (pair) of LKJ matrices, a result that rests on $k=2$. Numbers from the Gaussian sweep should not be carried to other dependence (Appendix~\ref{app:transfer}).

\paragraph{Chains and training.} In \textbf{H5a (\verdict{\V{H5a.verdict}})} the ratio $\dtot(L=4)/\dtot(L=1)$ exceeded 1 significantly in \V{H5a.sig} of \V{H5a.n} cells and was significantly below 1 in \V{H5a.opp}. Inputs of later neurons are independent, so their $D$ is nearly flat, and H5a does not test compounding of denominator error over correlated inputs. Training reduced $\dden$ significantly in \V{H5b.sig} of \V{H5b.n} evaluable cells (\textbf{H5b, \verdict{\V{H5b.verdict}}}; only $k=2$ is evaluable), $P(D<0)$ in \V{H5c.sig} of \V{H5c.n} (\textbf{H5c, \verdict{\V{H5c.verdict}}}) and held-out $\dtot$ in \V{H5d.sig} of \V{H5d.n} (\textbf{H5d, \verdict{\V{H5d.verdict}}}). The causes limit what these verdicts mean. About half of the end-to-end runs at $L=4$ kept the initial state (\V{ph.tr.e2e4.adam.kept} with Adam, \V{ph.tr.e2e4.lbfgs.kept} with L-BFGS; post hoc), and optimiser settings were fixed and untuned. The end-to-end objective fits only \V{e2e.targets} numbers ($3$ beliefs $\times$ $4$ moments) with \V{e2e.coef.lo} to \V{e2e.coef.hi} coefficients, and validation uses the same beliefs, so H5d says little about whether training can absorb the denominator. Under these settings, training did not reduce the three errors. Whether training can compensate remains open, and the HCR paper's expectation that it compensates for neglected calibration has not been tested (Appendix~\ref{app:chains}).

\section{Related work}\label{sec:related}

\paragraph{Uncertain evidence.} Jeffrey's rule~\citep{jeffrey1965logic,diaconis1982updating} and virtual evidence~\citep{pearl1988probabilistic} are the two classical updates, and \citet{chan2003revision,chan2005revision} show that both obey probability kinematics and translate into each other; \citet{peng2010uncertain} develop the Bayesian-network case. \citet{benmrad2015explication} separate likelihood (virtual) from probabilistic (Jeffrey) evidence in Bayesian networks, and \citet{jacobs2019mathematics} contrasts the two update rules and shows that they can lead to different conclusions. \citet{munk2022uncertain} show in modern probabilistic models how results depend on which interpretation of uncertain evidence is chosen, and \citet{yu2020soft} train Bayesian neural networks on soft evidence with approximations of Jeffrey conditionalisation. We add no new update rule. Our point is that a moment-vector message in a polynomial density neuron is one of these two rules, and that a formula presented as an approximation of one is exactly the other.

\paragraph{Messages and embeddings.} $\Eq[N]$ is one sum-product message~\citep{kschischang2001factor}, and a series message of this kind is sum-product by construction. The closest work on truncated series messages with estimated coefficients is the stochastic orthogonal-series message passing of \citet{noorshams2013belief}. Kernel mean embeddings use the same pattern of linear operators on moment-like vectors: the kernel sum and Bayes rules~\citep{fukumizu2013kernel} and kernel belief propagation~\citep{song2011kernel}, where estimated weights may also be negative. The kernel sum rule (Theorem~2 of~\citep{fukumizu2013kernel}) reads $m_{Q_Y}=C_{YX}C_{XX}^{-1}m_\Pi$, with $X$ the input and $Y$ the output in their notation. Since $C_{XX}$ applied to a function $f$ gives the embedding of the input density times $f$, the inverse acts as division by the input density. At population level the rule therefore corresponds to feeding the message $q/D$, that is Theorem~\ref{thm:sem}(d), the Jeffrey form; applying $C_{YX}$ with no inverse gives the evidence form. This is a correspondence, not a theorem, and the conversion is implicit in that literature. Expected predictions~\citep{khosravi2019expected} compute $\Eq$ of a model output under a generative input distribution in closed form, a different target from a normalised output density.

\paragraph{Positive and squared density models.} Squared expansions give nonnegativity by construction: \citet{gallant1987semi} in econometrics, and PSD models~\citep{rudi2021psd}, squared neural families~\citep{tsuchida2023squared} and subtractive mixtures via squaring~\citep{loconte2024subtractive} in machine learning. Some of these classes have closed forms for marginals, conditioning, or sum and product rules. The negativity of Section~\ref{sec:fail} is what those constructions avoid, at the cost of a quadratic form in place of a linear contraction. On corrections of series estimators see~\citep{gajek1986improving,glad2003correction}; on copula projection estimators see~\citep{ngounoubakam2025nonparametric,chatrabgoun2017legendre}.

\paragraph{Ratios and propagation.} The identity~\eqref{eq:identity} is classical in sampling theory~\citep{hartley1954unbiased,schochet2022design,cochran1977sampling,meng1993absolute}, the same structure appears in self-normalised importance sampling~\citep{owen2013monte,cardoso2022brsnis}, and delta-method expansions of a mean of a ratio are standard~\citep{oehlert1992note}. Most methods that push distributions through networks keep a parametric form: probabilistic backpropagation~\citep{hernandezlobato2015probabilistic}, natural-parameter networks~\citep{wang2016natural}, assumed density filtering~\citep{gast2018lightweight}, deterministic moment approximation~\citep{wu2019deterministic}, Gaussian inference~\citep{goulet2021tractable}, covariance propagation~\citep{wright2024analytic}, and lost inputs described by a Gaussian mixture~\citep{smieja2018processing}; see also~\citep{julier2004unscented,minka2001expectation}. HCR keeps a polynomial density, less restrictive in shape but possibly negative.

\paragraph{HCR and pre-registration.} HCR grew out of polynomial density fitting~\citep{duda2017rapid,duda2018hierarchical} and has been applied to time series~\citep{duda2018exploiting}, yield curves~\citep{duda2018modeling}, spreads~\citep{duda2020modelling}, income data~\citep{duda2018credibility} and redshifts~\citep{duda2024predicting}; Legendre KAN layers with an HCR reading are in~\citep{strawa2025improving}. We found no prior numerical study of the constant-denominator step. On pre-registration in machine learning see~\citep{nosek2018preregistration,forde2019scientific,bertinetto2021preface,lipton2019troubling,pineau2021improving,sandve2013ten,stodden2016enhancing}.

\section{Discussion}\label{sec:discussion}

\paragraph{Guidance.} The results reduce to seven working rules.
\begin{enumerate}
\item \emph{Decide what the belief is} (Theorem~\ref{thm:sem}). Per-input evidence and missing inputs call for $\pcd$. A distribution the inputs really follow, or a posterior or predictive marginal that already contains the prior, calls for $\peu$. To obtain the Jeffrey answer when $D>0$, feed the message $q/D$, which costs the moments of $q/D$; the practical route is the regression message of Section~\ref{sec:trunc} (post hoc), within its limits.
\item \emph{Read a flat belief correctly.} Under flat $q$, $\pcd$ is the model's output marginal in any basis. It is the answer to ``no information'', not the conditional averaged over a uniform input.
\item \emph{Check $\Eq[D]$ and $\CV(D)$} (Propositions~\ref{prop:gap},~\ref{prop:mass}). Both are cheap. A nonpositive or small $\Eq[D]$ should stop~propagation.
\item \emph{Do not spend effort on the denominator in pair neurons} under Beta-like beliefs where $\peu$ is defined (Table~\ref{tab:trunc}, post hoc): truncation dominates. In triple neurons under flat and data beliefs it can matter.
\item \emph{Check negativity.} For Gaussian-like dependence use the convex bound of Proposition~\ref{prop:pair} or the corners of Proposition~\ref{prop:corner}. Expect an onset near $\rs\sim1/k$, and earlier with degree or a triple basis. Do not transfer the numbers to heavy-tailed or asymmetric dependence (H3a, H3b) or to unequal correlations (H3c).
\item \emph{Size $n$ with $z^2K_{\mathrm c}$} (Proposition~\ref{prop:noise}). Prefer the pair basis with low degree at large $k$, or select coefficients, as the HCR paper suggests.
\item \emph{Be careful with clipping.} Clipping $N$ at each value of $y$ moved outputs away from the truth in most defined evaluations (Table~\ref{tab:trunc}, post hoc).
\end{enumerate}

\paragraph{Limitations.}
\begin{itemize}
\item \emph{Scope.} One neuron family, a Legendre basis, $m\leq6$ and, in the study, synthetic copulas only. The only real data are an exploratory pilot with one data set and one untrained neuron (Appendix~\ref{app:pilot}), so we cannot say in general how much a TV of a given size matters on a task. The basis keeps all coefficients of its family, and coefficient selection was not tested. The HCR paper's value-propagation mode does not use Eq.~10 and is outside this study.\looseness=-1 
\item \emph{The semantic reading} is post hoc and has no pre-registered test: the study's truth is the Jeffrey answer, and the evidence semantics is only illustrated, in full and triple models up to $k=8$, with the pair family losing the contrast to model error. The pilot has no Jeffrey-type arm. Explanations offered for tail dependence and mixing are consistent with the data but untested.
\item \emph{Theory.} Sections~\ref{sec:gap} and~\ref{sec:fail} assume population coefficients, uniform marginals and, for closed forms, equicorrelated Gaussian dependence; rates are at fixed $m$, the Gaussian limit in Proposition~\ref{prop:noise} is assumed, not proved uniformly in $k$, and the convex bound is computed on a grid.
\item \emph{Design.} The main sweep's equicorrelation favours large pair sums, the coupling $c$ is a design choice, and marginals are exactly uniform, whereas real pipelines use estimated quantiles. The chain has independent inputs after layer one, a fixed contractive kernel and no re-normalisation to quantiles, and training used one fixed setting per optimiser. The end-to-end objective fits \V{e2e.targets} numbers on the beliefs also used for validation, so H5d says little about whether training can absorb the denominator.
\item \emph{Measurement.} Measured onsets are upper bounds on a grid; TV on signed densities is an $L^1$ distance, not a probability bound. Confidence intervals are per cell with no multiplicity correction; verdicts rest on the counting rules.
\item \emph{Pre-registration.} The protocol was not entered in a public registry; run records and scheduler times are the only evidence of its timing.
\end{itemize}

\paragraph{Broader impact.} This is foundational work on a density model: a synthetic study and a small pilot on public census data aggregated to block groups, with no individual-level data. The practical risk is misuse of $\pcd$ or $\peu$ as calibrated uncertainty when the belief's meaning has not been considered; the guidance above is aimed at that.

\section{Conclusion}

A reader who propagates beliefs through a density model should ask first what the belief is, because the choice between normalising after and normalising before is a choice of question, not of accuracy. For polynomial neurons the rest can be quantified: the gap is second order in the dependence for pair bases and first order for triple bases, the reference disappears near $\rs\sim1/k$, and, post hoc, where $\peu$ is defined, in pair neurons under Beta beliefs truncation usually outweighs the gap, while in triple neurons under flat and data beliefs it does not. Thresholds did not transfer beyond the Gaussian sweep, and whether training can compensate for any of this remains open. The study is entirely synthetic; an exploratory pilot on one real data set agrees with the evidence side of the thesis and leaves the neuron well behind a standard baseline. Three experiments would settle most of what we leave unsettled: a pre-registered test of the two semantics, a pre-registered real task with missing or noisy inputs, and chains whose inputs stay correlated.\looseness=-1 

\section*{Code and data availability}\pdfbookmark[1]{Code and data availability}{availability}
The code is at \url{https://github.com/Przemyslaw11/EvidenceOrMarginal}. Appendix~\ref{app:repro} gives batches, seeds, hardware and software. All data in the study are synthetic and generated by the code; the pilot of Appendix~\ref{app:pilot} uses the public California housing data~\citep{pace1997sparse}.

\section*{Acknowledgements}\pdfbookmark[1]{Acknowledgements}{acknowledgements}
The author wishes to thank the PLGrid Infrastructure for its support and ACK~Cyfronet AGH for providing the computational resources on the Athena cluster.

\bibliographystyle{plainnat}
\bibliography{refs}

\appendix
\section{Notation and status of claims}\label{app:claims}

This appendix collects the symbols used across sections (Table~\ref{tab:notation}) and states, claim by claim, what kind of evidence stands behind each (Table~\ref{tab:claims}).

\begin{table}[htbp]
\centering\footnotesize
\caption{Notation shared across sections. Symbols used in one proof only are defined there.}
\label{tab:notation}
\begin{tabular}{@{}l>{\raggedright\arraybackslash}p{0.29\linewidth}l>{\raggedright\arraybackslash}p{0.32\linewidth}@{}}
\toprule
symbol & meaning & symbol & meaning \\
\midrule
$N,\ D,\ h_x$ & joint model, input density, conditional $N/D$ & $G,\ V$ & $D-1$ and $N-D$ \\
$A_i,\ a_{i\mathbf{j}}$ & row polynomial and its coefficients & $T_1,\ T_2$ & $\Eu[GV]$ and $\Eu[G^2V]$ \\
$q,\ \tilde q,\ u$ & belief, tilted belief, flat belief & $\chi_D,\ \delta_h$ & distance of $D$ from flat; $\sup_y\TV(h_\cdot(y),1)$ \\
$b_{\mathbf{j}},\ b_l$ & moments of $q$; moment vector of $q_l$ & $C_h$ & $\sup_y\int|h_x(y)|\,dx$ \\
$\pcd,\ \peu,\ \pec,\ \pg$ & message, marginal-preserving form, clipped form, ground truth & $W,\ g,\ \xi_l,\ \bar\xi_l$ & latent factor, $\mathrm{E}[f(y)\mid W]$, $g^\top f(y_l)$, $g^\top b_l$ \\
$k,\ m,\ n$ & inputs, degree, sample size & $\Gamma,\ \Gamma^{ll'}_i$ & pair and triple coefficient arrays \\
$r,\ \rs,\ c$ & input correlation, its Spearman value, coupling to $x$ & $\Theta,\ K_{\mathrm c},\ z$ & $m(m+2)$, corner variance factor, standardised minimum \\
$\sobs,\ o$ & observation noise, observation & $\kappa_{ab},\ \se^2,\ \so^2$ & corner moments, even and odd energies \\
$M,\ R$ & Monte Carlo samples, replicate blocks & $e_d,\ \omega_p$ & elementary symmetric polynomial; Corollary~\ref{cor:closed} \\
$\alpha,\ \gamma,\ L$ & chain correlations and depth & $j$ & coordinates at $1$ of a corner \\
$\dden,\ \dneg$ & gap $\TV(\pcd,\peu)$; clipping $\TV(\peu,\pec)$ & $\dmu,\ \dmc$ & errors $\TV(\peu,\pg)$, $\TV(\pec,\pg)$ \\
$\dtot,\ \ddc$ & $\TV(\pcd,\pg)$, $\TV(\pcd,\pec)$ & $\TV,\ \CV(D)$ & $\frac12\int|p-p'|$; $\sd_q(D)/\Eq[D]$ \\
S1--S4 & beliefs: own $D$, flat, Beta (S3a--d), data & SUM, CONTRAST & ground truths with equal or opposite couplings \\
$\varepsilon$ & clipping level of $\pec$ & floor & $D\geq\V{cfg.floor}$ needed for $\peu$ to be defined \\
\bottomrule
\end{tabular}
\end{table}

\begin{table}[htbp]
\centering\footnotesize
\caption{Status of each main claim.}
\label{tab:claims}
\begin{tabular}{@{}>{\raggedright\arraybackslash}p{0.43\linewidth}>{\raggedright\arraybackslash}p{0.54\linewidth}@{}}
\toprule
claim & status \\
\midrule
Semantics, tilt, conversion (Thm.~\ref{thm:sem}, Props.~\ref{prop:chan},~\ref{prop:any}) & Proved; classical mathematics, applied here. Post hoc. No pre-registered test. Illustrated in full and triple models up to $k=8$; the pair family loses the contrast. \\
Kernel sum rule as the Jeffrey form (Sec.~\ref{sec:related}) & Population-level correspondence, not a theorem. Post hoc. \\
Gap identity and bounds (Prop.~\ref{prop:gap}) & Proved; identity classical; bound checked on population cases. \\
Order of the gap, closed forms (Prop.~\ref{prop:flat}, Cor.~\ref{cor:closed}) & Proved under stated assumptions; the $k,\rs$ scaling is a leading-order reading; compared post hoc with H1c points. \\
Data belief and pair structure (Prop.~\ref{prop:data}) & Proved; checked post hoc on population evaluations. \\
Onset bounds, corners (Props.~\ref{prop:pair},~\ref{prop:corner}) & Proved for the equicorrelated Gaussian copula; convex value computed on a grid; closeness to the measured onset is numerical. The value of $k\rs^\ast$ at large $k$ is a post hoc computation, with no limit theorem. \\
Mass of the message (Prop.~\ref{prop:mass}) & Formula proved; sign reading for central beliefs heuristic. \\
Noise scaling (Prop.~\ref{prop:noise}) & Moments proved; extremes proved in the Gaussian limit, which is assumed; surrogate numerics. \\
Truncation against gap; clipping (Secs.~\ref{sec:trunc},~\ref{sec:clip}) & Post hoc analysis of pre-registered metrics; pair-basis, Beta-belief statement on defined evaluations, with named exceptions. \\
Regression message (Table~\ref{tab:reg}) & Post hoc computation on a Gaussian teacher, not part of the study; targets the Jeffrey answer only. \\
Real-data pilot (Appendix~\ref{app:pilot}) & Exploratory, not pre-registered, run after the results were seen; one data set, one untrained neuron, no tuning, no Jeffrey-type arm. \\
Onset and gap outcomes, H1--H5 & Pre-registered; verdicts in Table~\ref{tab:hyp}. \\
Explanations for tail dependence and mixing & Consistent with the data; untested. \\
\bottomrule
\end{tabular}
\end{table}

\section{Proofs}\label{app:proofs}

All statements of Sections~\ref{sec:semantics}--\ref{sec:fail} are proved here, in order. Each was also checked numerically against exact formulas and the study's data.

\begin{proof}[Proof of Theorem~\ref{thm:sem}]
(a) Let $c=1/\sup q$. The joint of $(x,y,E=1)$ is $N(x,y)\,c\,q(y)$, and its normaliser is $c\iint Nq\,dx\,dy=c\,\Eq[D]$. So $p(x,y\mid E=1)=N(x,y)q(y)/\Eq[D]$. Integrating over $y$ gives $\Eq[N]/\Eq[D]$; integrating over $x$ gives $Dq/\Eq[D]$.
(b) A joint density $J(x,y)$ with input marginal $q$ and conditional $h_x$ on the support of $q$ is $J=q\,h_x$ by definition of a conditional. Integrating over $y$ gives $\Eq[h_x]$.
(c) $\int h_x\tilde q\,dy=\Eq[h_xD]/\Eq[D]=\Eq[N]/\Eq[D]$. The second equality is the definition.
(d) Let $q'=(q/D)/\Eq[1/D]$, a probability density. Then $\mathrm{E}_{q'}[N]=\Eq[N/D]/\Eq[1/D]$ and $\mathrm{E}_{q'}[D]=1/\Eq[1/D]$, so the ratio is $\peu[q]$. And $\peu[\tilde q]=\pcd[q]$ is (c).
\end{proof}

\begin{proof}[Proof of Proposition~\ref{prop:chan}]
(a) Under the model $p(x\mid o)\propto\int N(x,y)\prod_l\lambda_l(y_l)\,dy$, since $o$ is independent of $x$ given $y$. With uniform one-dimensional marginals $q_l=\lambda_l/\int\lambda_l$ is the posterior of $y_l$ given $o_l$ alone, and $\prod_l\lambda_l\propto q$. So $p(x\mid o)\propto\Eq[N]$, and $\int\Eq[N]\,dx=\Eq[D]$ normalises. In general $\peu[q]=\Eq[h_x]$ differs by~\eqref{eq:identity}.
(b) $q^\ast\propto D\prod_l\lambda_l$. Then $\peu[q^\ast]\propto\int h_xD\prod\lambda_l\,dy=\int N\prod\lambda_l\,dy$. While $\pcd[q^\ast]\propto\int ND\prod\lambda_l\,dy$ contains $D$ as well as the prior already in $N$.
(c) The belief has moments $b_{\mathbf{j}}=f_{\mathbf{j}}(\yo)$ if $\mathbf{j}$ is zero on the missing inputs, and $b_{\mathbf{j}}=0$ otherwise. So $\Eq[N]=\int N(x,\yo,\ym)\,d\ym=p(x,\yo)$ and $\Eq[D]=p(\yo)$. The ratio is $p(x\mid \yo)=\int p(x\mid \yo,\ym)\,p(\ym\mid \yo)\,d\ym$, which equals the integral against the flat density when $\ym$ is independent of $\yo$ under the model, and in general not otherwise.
\end{proof}

\begin{proof}[Proof of Proposition~\ref{prop:any}]
$\int(\pi/Z)\,d\tilde q=\Eq[(\pi/Z)Z]/\Eq[Z]=\Eq[\pi]/\Eq[Z]$. If $\iint \pi<\infty$, then $p=\pi/\iint \pi$ is a joint density with input density $Z/\int Z$ and conditional $\pi/Z$, and Theorem~\ref{thm:sem}(a) applies, with bounded $q$.
\end{proof}

\begin{proof}[Proof of Proposition~\ref{prop:gap}]
$N=h_xD$ gives $\Eq[N]=\Eq[h_x]\Eq[D]+\Cov_q(h_x,D)$; divide by $\Eq[D]$. ($D$ is bounded and $h_x$ is $q$-integrable by assumption, so the covariance exists. Square-integrability makes the standard deviations below finite.) Writing the covariance as correlation times standard deviations and integrating over $x$ gives the equality in~\eqref{eq:tvbound}. For the inequality use $|\mathrm{corr}|\leq1$ and Cauchy--Schwarz on $[0,1]$: $\int\sd_q(h_x)\,dx\leq(\int\Var_q(h_x)\,dx)^{1/2}$. Since $h_x=\sum_if_i(x)A_i/D$, Parseval gives $\int\Var_q(h_x)\,dx=\sum_i\Var_q(A_i/D)$, and the term $i=0$ vanishes as $A_0/D=1$. For the second bound, $\pcd-\peu=\int h_x\,(\tilde q-q)\,dy$ by Theorem~\ref{thm:sem}(c), whose proof needs only $\Eq[D]\neq0$; then $\int|\pcd-\peu|\,dx\leq\int\!\int|h_x|\,dx\,|\tilde q-q|\,dy\leq2C_h\TV(\tilde q,q)$, and $\TV(\tilde q,q)=\frac12\Eq|D/\Eq[D]-1|\leq\frac12\CV(D)$ by Jensen. If $x$ and $y$ are independent, $h_x$ is constant in $y$; if $D$ is constant on the support of $q$, the covariance in~\eqref{eq:identity} vanishes.
\end{proof}

\begin{proof}[Proof of Proposition~\ref{prop:flat}]
$\Eu[D]=a_{00}=1$, so $\pcd=\Eu[h_xD]=\peu+\Eu[h_xG]$, and $\Eu[h_xG]=\Eu[(h_x-1)G]$ since $\Eu G=0$. Now $h_x-1=V/D$ and, from $D=1+G$, $1/D=1-G/D$, so
\begin{equation}\label{eq:rec}
h_x-1=V-G(h_x-1).
\end{equation}
Substituting gives~\eqref{eq:flat}. For the bound, $\int|\Eu[G^2(h_x-1)]|\,dx\leq\Eu[G^2\int|h_x-1|\,dx]\leq2\chi_D^2\delta_h$, and $|\TV(\pcd,\peu)-\frac12\|T_1\|_1|$ is at most half of that by the triangle inequality.

(a) In the pair basis $A_i$, $i\geq1$, is a constant plus single-input terms, and with flat input marginals $G$ is a sum of two-input terms. These are orthogonal under $u$, and $\Eu G=0$ removes the constant, so $T_1\equiv0$. Applying~\eqref{eq:rec} once more in $\Eu[G^2(h_x-1)]$ gives $\pcd-\peu=-T_2+\Eu[G^3(h_x-1)]$, and $\int|\Eu[G^3(h_x-1)]|\,dx\leq\|G\|_\infty\Eu[G^2\int|h_x-1|\,dx]\leq2\delta_h\|G\|_\infty\chi_D^2$.

(b) By orthonormality $\Eu[GA_i]=\sum_{\mathbf{j}\neq0}a_{0\mathbf{j}}a_{i\mathbf{j}}$ over common indices. With flat marginals only two-input indices are common.
\end{proof}

\begin{proof}[Proof of Corollary~\ref{cor:closed}]
Population coefficients with uniform marginals give $a_{i0}=0$ for $i\geq1$, $a_{0,(l,p)}=0$, and $a_{0,(l,p;l',q)}=\Gamma_{pq}$ for all pairs. So $G=\sum_{l<l'}f(y_l)^\top \Gamma f(y_{l'})$ and $V=\sum_{i,p\geq1}f_i(x)\sum_lB_{ip}(c_l)f_p(y_l)$. To compute $\Eu[G^2f_p(y_t)]$, expand $G^2$ over pairs of index pairs. Any variable that appears exactly once, other than $y_t$, integrates to zero, since each factor has degree at least 1. What remains are the $k-1$ squares of pairs containing $t$, each giving $\int f_p(u)f(u)^\top \Gamma\,\mathrm{E}[f(v)f(v)^\top]\,\Gamma f(u)\,du=\omega_p$, because $\mathrm{E}[ff^\top]=I$ under the flat law. Pairs without $t$ vanish as $\Eu f_p=0$. This gives $T_2$. $T_1$ is Proposition~\ref{prop:flat}(b) with $a_{i,(l,p;l',q)}=(\Gamma^{ll'}_i)_{pq}$.

Symmetry. Since $(y\mapsto1-y)$ preserves the equicorrelated law and $f_p(1-u)=(-1)^pf_p(u)$, $\Gamma_{pq}=0$ for $p+q$ odd. So $u\mapsto f(u)^\top \Gamma^2f(u)$ is even around $\frac12$, and $\omega_p=0$ for odd $p$. Mehler's expansion gives $B_{ip}(c)=0$ for $i+p$ odd, and $B_{ip}(-c)=(-1)^pB_{ip}(c)$. So for even $p$ the sign of $c$ drops, and only even $i$ remain: $f_1$ has no coefficient, and the mean, which is linear in the $f_1$ coefficient, does not move at this order.
\end{proof}

\begin{proof}[Proof of Proposition~\ref{prop:data}]
$\Eq[D]=\sum_{\mathbf{j}}a_{0\mathbf{j}}\Eq[f_{\mathbf{j}}]=\sum_{\mathbf{j}}a_{0\mathbf{j}}^2=1+\chi_D^2$, as $\Eq[f_{\mathbf{j}}]=a_{0\mathbf{j}}$ for population coefficients. The true propagated density is the marginal of $x$, which is uniform. For~\eqref{eq:pairstruct}: in the pair basis $\Eq[A_i]=a_{i0}+\sum_{l,p}a_{i,(l,p)}\mathrm{E}_{q_l}[f_p]$ with $a_{i0}=0$, and $\Eq[A_0]=\Eq[D]$, so $\pcd=1+\sum_{i\geq1}f_i\Eq[A_i]/\Eq[D]$. Under the data belief $\mathrm{E}_{q_l}[f_p]=0$, so $\pcd\equiv1$. In the triple basis $\Eq[A_i]$ gains $\sum a_{i\mathbf{j}}\Eq[f_{\mathbf{j}}]=\sum a_{i\mathbf{j}}a_{0\mathbf{j}}$ over two-input $\mathbf{j}$, which is the $f_i$ coefficient of $T_1$.
\end{proof}

\begin{proof}[Proof of Equation~\eqref{eq:latent} and Proposition~\ref{prop:pair}]
Given $W=w$ the $y_l$ are independent, each with density $1+\sum_{p\geq1}g_p(w)f_p(y_l)$ in the full basis. So for an index $\mathbf{j}$ with active degrees $p_1,\dots,p_d\geq1$, we have $a_{0\mathbf{j}}=\EW\prod_tg_{p_t}$. Since $\EW g_p=\mathrm{E}f_p(y)=0$, single-input terms vanish, and summing the two- and three-input terms gives~\eqref{eq:latent}. With $\xi_l=g^\top f(y_l)$, $\EW e_2(\xi)=\sum_{l<l'}f(y_l)^\top \Gamma f(y_{l'})=\frac12(F^\top \Gamma F-\sum_l\varphi(y_l))$, which is~\eqref{eq:quad}.

$\Gamma\succeq0$ as a second-moment matrix. For $\Gamma\preceq rI$, let $v\in\mathbb R^m$ and $U=v^\top f(y)$, a mean-zero function of $s=\sqrt rW+\sqrt{1-r}\nu$ with variance $|v|^2$. Expand $U=\sum_{n\geq1}\gamma_n\mathrm{He}_n(s)/\sqrt{n!}$. Then $\mathrm{E}[\mathrm{He}_n(s)\mid W]=r^{n/2}\mathrm{He}_n(W)$, so $v^\top \Gamma v=\mathrm{E}[(\mathrm{E}[U\mid W])^2]=\sum_{n\geq1}r^n\gamma_n^2\leq r|v|^2$.

(a) $F^\top \Gamma F\geq0$ gives the first bound. Also $\varphi(u)\leq r|f(u)|^2\leq r\sum_{p=1}^m(2p+1)=r\Theta$, as $|f_p|\leq\sqrt{2p+1}$. So $D>0$ if $k r\Theta/2<1$. Finally $r=2\sin(\pi\rs/6)$.
(b) With $\mu=\frac1k\sum_l\delta_{y_l}$, $F=k\bar f_\mu$ and $\sum_l\varphi(y_l)=k\int\varphi\,d\mu$, so $D(y)=\Psi_k(\mu)$. Minimising over all probability measures can only lower the value. $\Psi_k$ is a convex quadratic in $\bar f_\mu$ plus a linear term, and $\bar f_\mu$ is linear in $\mu$.
\end{proof}

\begin{proof}[Proof of Proposition~\ref{prop:corner}]
At a corner $\xi$ has $j$ entries $\xi_+$ and $k-j$ entries $\xi_-$, so $e_2=\binom j2\xi_+^2+\binom{k-j}2\xi_-^2+j(k-j)\xi_+\xi_-$, and $e_3$ likewise. The map $y\mapsto1-y$, $w\mapsto-w$ gives $g_p(-w)=(-1)^pg_p(w)$, hence $\xi_-(w)=\xi_+(-w)$. As $W$ is symmetric, $\EW\xi_-^2=\kappa_{20}$, $\EW\xi_-^3=\kappa_{30}$, $\EW\xi_+\xi_-^2=\kappa_{21}$. This gives~\eqref{eq:cornerp} and~\eqref{eq:cornert}. Writing $\xi_\pm=\xe\pm\xo$ with $\xe$ even and $\xo$ odd in $w$, $\se^2=\EW\xe^2$, $\so^2=\EW\xo^2$, $\kappa_{20}=\se^2+\so^2$ and $\kappa_{11}=\se^2-\so^2$, and algebra gives the stated form. For the triple remark, $\kappa_{30}=\EW\xe^3+3\EW\xe\xo^2$ and $\kappa_{21}=\EW\xe^3-\EW\xe\xo^2$; put $j=k/2$. A corner lies in the cube, so a negative corner value means $\min D<0$ at that $\rs$.
\end{proof}

\begin{proof}[Proof of Proposition~\ref{prop:mass}]
$D$ is affine in each $f(y_l)$ by~\eqref{eq:latent}, and $q$ is a product, so $\Eq$ replaces $f(y_l)$ by $b_l$, which gives the first formula. Equation~\eqref{eq:masspair} follows as in~\eqref{eq:quad}, with $\Gamma\succeq0$ and $\Gamma\preceq rI$. If $b_l=b$, then $\Eq[D]=1+\binom k2b^\top \Gamma b\geq1$. The triple formula follows from $e_2$ and $e_3$ at equal arguments.
\end{proof}

\begin{proof}[Proof of Proposition~\ref{prop:noise}]
With independent uniform inputs, $\hat a_{0\mathbf{j}}$ for $\mathbf{j}\neq0$ have mean $0$, variance $1/n$ and are uncorrelated, by orthonormality of $f_{\mathbf{j}}$. So $\Var\hat D(y)=\sum_{\mathbf{j}\neq0}f_{\mathbf{j}}(y)^2/n$. $K$ is a sum of products of $s_l=|f(y_l)|^2\leq\Theta$, attained at corners, where $|f_p|=\sqrt{2p+1}$.

Let $\zeta_{\mathbf{j}}$ be i.i.d.\ standard normal and $P(y)=K_{\mathrm c}^{-1/2}\sum_{\mathbf{j}\neq0}\zeta_{\mathbf{j}}f_{\mathbf{j}}(y)$, so $z=\sup_y(-P(y))$ and $-P$ has the law of $P$. Its variance is at most $1$, and $1$ at corners.

\emph{Corners.} The maximum of $2^k$ unit normals has mean at most $\sqrt{2k\ln2}$.

\emph{Upper bound on the cube.} Write $P=\sum_{d\leq d_{\max}}P_d$ with $P_d$ the part with $d$ active inputs, $d_{\max}\leq3$. Let $v_l=f(y_l)$ in the ball $\mathbb B$ of radius $\sqrt\Theta$ of $\mathbb R^m$, and regard $P_d$ as a function on $\mathbb B^k$. It is affine in each $v_l$ and homogeneous of degree $d$. Let $\mathcal N$ be a $\frac12\sqrt\Theta$-net of $\mathbb B$ with $|\mathcal N|\leq5^m$, so $\frac12\mathbb B\subset\mathrm{conv}\,\mathcal N$~\citep{vershynin2018high}. A function affine in each block attains its maximum over a product of convex hulls at a product of vertices, so $\sup_{\mathbb B^k}P_d=2^d\sup_{(\mathbb B/2)^k}P_d\leq2^d\max_{\mathcal N^k}P_d$. Each value is a centred normal with variance at most $1$, so $\mathrm{E}\max_{\mathcal N^k}P_d\leq\sqrt{2km\ln5}$. Summing over $d$ gives $\mathrm{E}z\leq C\sqrt{km}$.

\emph{Lower bound.} Let $m$ be even and take corners $y,y'$ that differ in a fraction $\delta$ of coordinates. Each coordinate gives $\langle f(y_l),f(y_l')\rangle=\Theta$ if equal, and $\sum_{p=1}^m(-1)^p(2p+1)=m$ if different, a ratio $1/(m+2)$. The covariance of $P$ at $y,y'$ is thus $\sum_d e_d(\text{inner products})/K_{\mathrm c}$, which for large $k$ tends to the top-degree value $(1-\delta+\delta/(m+2))^{d_{\max}}$. By the Gilbert--Varshamov bound there are $2^{c_1k}$ corners at pairwise distance at least $k/4$, so their pairwise correlations are at most a constant below $1$ for large $k$, and their pairwise $L^2$ distances at least a constant. Sudakov's minoration~\citep{vershynin2018high} gives $\mathrm{E}z\geq c\sqrt k$.

\emph{Concentration.} $z$ is a supremum of a Gaussian process with variance at most $1$, so by the Borell--TIS inequality~\citep{vershynin2018high} it has sub-Gaussian fluctuations of order $1$ around its mean. Then $\min\hat D>0$ is the event $z\sqrt{K_{\mathrm c}/n}<1$, which gives the stated order with $K_{\mathrm c}\sim\frac12k^2\Theta^2$ or $\frac16k^3\Theta^3$ and $z^2\asymp k$ up to $m$.
\end{proof}

\section{The illustration of Section~\ref{sec:illus}}\label{app:illus}

The tables below give the full grid behind Figure~\ref{fig:idea} and the same comparison at larger $k$: in the pair family, which loses the flip, and in the full and triple models, which keep it. All are post hoc illustrations, not study results.

\begin{table}[htbp]
\centering
\caption{Illustration at $k=2$: median TV to the truth over 200 observations, for the three problems of Section~\ref{sec:illus}.}
\label{tab:sem}
\input{generated/tab_sem.tex}
\end{table}

\begin{table}[htbp]
\centering
\caption{Illustration at $k=4,8$, pair basis, below the onset, 60 observations; ``closer'' is the share of observations with $\pcd$ nearer the truth than $\peu$. Pair rows $i\geq1$ are additive in $y$, so they cannot carry the conditional, and this model error hides the flip of Table~\ref{tab:sem}.}
\label{tab:sembig}
\input{generated/tab_sem_big.tex}
\end{table}

\begin{table}[htbp]
\centering
\caption{Post hoc illustration beyond the pair family, below the onset ($\min D$ over the cube between \V{sem.flip.mind.lo} and \V{sem.flip.mind.hi}), 60 observations; ``closer'' as in Table~\ref{tab:sembig}. At $\sobs=1$ every model flips as in Table~\ref{tab:sem}. At $\sobs=0.5$ the belief is narrower and the full tensor still flips, while the triple family at $k=4,8$ shows almost no difference on the posterior and marginal problems. The median of 60 observations carries a Monte Carlo error of a few thousandths. $k=8$ is the largest case tried.}
\label{tab:semflip}
\input{generated/tab_sem_flip.tex}
\end{table}

\section{An exploratory real-data pilot}\label{app:pilot}

This pilot is exploratory. It was not pre-registered, it was run after the study's results were seen, it uses one data set and one untrained, directly estimated neuron with no tuning, and it is not a statement about trained HCR networks. It enters no verdict.

\paragraph{Data and neuron.} California housing~\citep{pace1997sparse} has \V{pilot.inputs} inputs and the median house value as target. All variables are mapped to $[0,1]$ by their empirical distribution functions on the training split, with \V{pilot.train} training and \V{pilot.test} test rows; \V{pilot.seeds} seeds redraw the split. In the first split the inputs' Spearman correlations have median magnitude \V{pilot.rho.med} and maximum \V{pilot.rho.max}, so the data lie well past the onset of Section~\ref{sec:fail}. The neuron uses direct moment estimates in the pair and triple families with $m=2,3$.

\paragraph{Scenarios.} \emph{All seen}: the belief is a point. \emph{Half missing}: each test input is lost with probability $\frac12$, and the belief is a point on the rest and flat on the lost ones, the case of Proposition~\ref{prop:chan}(c). \emph{Noisy}: each input's normal score is seen with noise $\sobs=\V{pilot.tau}$, as in Section~\ref{sec:illus}. Marginals are uniform, so the per-input posterior is the normalised likelihood, and Proposition~\ref{prop:chan}(a) makes $\pcd$ the right form here too. Both of the last two scenarios are of evidence type. No Jeffrey-type arm was run, so the pilot shows one side of the flip only.

\paragraph{Forms and scores.} $\pcd$ is computed exactly from per-input moments. $\peu$ uses \V{pilot.samples} belief samples per row and drops those with $D<\V{pilot.floor}$; a row with no sample left returns the marginal. REG is the regression message of Section~\ref{sec:trunc} with ridge regression on the neuron's features. Boosting is gradient boosting on trees with default settings, trained on inputs corrupted in the same way. We report the mean squared error of the predicted mean clipped to $[0,1]$, and the mean log density with the density clipped at $\V{pilot.floor}$ and renormalised.

\paragraph{Results.} Table~\ref{tab:pilot} reads in both directions. $\pcd$ beats $\peu$ in the two evidence-type scenarios, as Proposition~\ref{prop:chan} predicts, in \V{pilot.missing.win} of \V{pilot.missing.win.n} seed and family pairs with half the inputs missing and \V{pilot.noisy.win} of \V{pilot.noisy.win.n} with noise, and the log density agrees. With all inputs seen the belief is a point, so the two forms coincide wherever $D$ is not small; the difference in that line comes from rows below the floor, where $\peu$ returns the marginal. The neuron trails boosting by a wide margin in every scenario. In clipped-mean error the regression message is ahead of both forms with all inputs seen and with noise, and level with $\pcd$ with half missing, despite its semantics; in log density it leads only with all inputs seen. Unguarded $\pcd$ fails on rows where $\Eq[D]$ is small or negative (Table~\ref{tab:pilotfail}): such rows are few, yet the raw mean squared error reaches \V{pilot.rawmax}, against a marginal baseline of \V{pilot.marg}. This is why Rule~3 of Section~\ref{sec:discussion} matters. So on this one data set the choice of form has a visible effect in the predicted direction, and the neuron as estimated here is not competitive.

\begin{table}[htbp]
\centering
\caption{Exploratory pilot, means over \V{pilot.seeds} seeds; ranges run over the four families. The largest standard deviation over seeds of any entry is \V{pilot.sd}. The marginal has log density $0$ by construction.}
\label{tab:pilot}
\input{generated/tab_pilot.tex}
\end{table}

\begin{table}[htbp]
\centering
\caption{Where the neuron fails in the pilot; ranges run over the four families. ``below the floor'': share of test points (all seen) or belief samples (other scenarios) with $D<\V{pilot.floor}$. ``CD ahead of EU'': seed and family pairs with lower clipped-mean error for $\pcd$. ``largest raw MSE of CD'': worst seed and family with no clipping of the predicted mean.}
\label{tab:pilotfail}
\input{generated/tab_pilot_fail.tex}
\end{table}

\section{Full definitions, grids and numerics}\label{app:defs}

This appendix completes Section~\ref{sec:design}: exact definitions of each measured quantity, the grids and ground truths of each experiment, and the numerical settings.

\paragraph{Support and floor.} The support of a belief is $\{q\geq10^{-6}\max q\}$. $\peu$ is \emph{defined} for an evaluation if no Monte Carlo sample in the support and no optimiser point in the support has $D<\V{cfg.floor}$. For Beta beliefs this is not mathematical definedness. A Beta density is positive on the open cube, so wherever $D$ changes sign $\Eq|N/D|$ is infinite, and what is reported is the integral restricted to the cut. Under S3d at $m=4$, $\rs>0$, $D$ changes sign on the cube in \V{cut.sign.2} of defined evaluations at $k=2$ and \V{cut.sign.32} at $k=32$. The cut acts on the joint density, so its reach changes with $k$: the share of Beta samples inside it falls to \V{cut.in.lo}--\V{cut.in.hi} at $k=32$, depending on the belief, and the share of defined evaluations is not monotone in $k$. It should not be read as a trend at $k\geq16$. Values among defined evaluations are safe: samples dropped below the floor outside the cut are at most \V{cut.drop} of $M$ (post hoc).

\paragraph{TV.} $\TV(p,p')=\frac12\int|p-p'|$ on \V{cfg.snodes} nodes in normal score $s\in[-8,8]$ with a cubic sub-interval rule, on raw signed densities. Under a monotone change of variable TV does not change; the normal score resolves the sharp conditional densities at high correlation. $\pec$ uses \V{cfg.gl}-node Gauss--Legendre integration.

\paragraph{Variance and mean errors.} $\mathrm{relvar}=(\Var_{\mathrm{CD}}-\Var_{\mathrm{ref}})/\Var_{\mathrm{ref}}$ in $x$, and $\mathrm{smean}=|\mathrm{Mean}_{\mathrm{CD}}-\mathrm{Mean}_{\mathrm{ref}}|/\sd_{\mathrm{ref}}$, with reference $\peu$ (primary) or $\pg$.

\paragraph{Noise.} With $R$ replicate blocks, $\mathrm{SE}=\sd(\text{replicates})/\sqrt R$, and the noise floor is the mean replicate-difference $L^1$ divided by $\sqrt{R-1}$. Values below three times this floor are flagged and reported, not subtracted.

\paragraph{Seed-median curve.} Undefined seed values count as $+\infty$; a grid point is defined if at least $50\,\%$ of seeds are defined. H1c, H2c and the H3c fit use medians over defined seeds; with undefined seeds as $+\infty$, no verdict changes (post hoc). ``Undefined from'' is the first grid point at which at least $50\,\%$ of seeds have $\peu$ undefined. Thresholds are per seed too, and bootstrap CIs ($2{,}000$ resamples) resample seeds, paired where conditions share seeds; ``not reached'' counts as $+\infty$ in the percentile.

\paragraph{Grids and ground truths.}\label{app:grids} Besides the Gaussian sweep the study uses $t_3$ and $t_{10}$ copulas~\citep{demarta2005t} in SUM, CONTRAST and exchangeable (EXCH) forms, and exchangeable Clayton and Gumbel copulas~\citep{mcneil2009multivariate}, all at matched Kendall's $\tau$ of the inputs~\citep{lindskog2003kendall}. Archimedean samples use the mixture representation of Marshall and Olkin~\citep{marshall1988families}, with Kanter's method~\citep{kanter1975stable} for the positive stable variables. Random correlation matrices come from the LKJ distribution~\citep{lewandowski2009generating}.

$\rs\in\{0,0.01,0.02,0.03,0.05,0.075,0.1,0.15,0.2,\dots,0.5\text{ step }0.05,0.6,0.7,0.8,0.9\}$; $m\in\{2,4,6\}$; $\varepsilon\in\{0.1,0.01,0.001\}$; $M=\V{cfg.M}$; $R=\V{cfg.R}$. Gaussian sweep: 50 seeds. Population sweep: exact coefficients, 5 belief seeds. Copula families: $k\in\{2,4,8\}$, 20 seeds. LKJ matrices: $k\in\{2,4,8\}$, $\eta\in\{1,4\}$, 200 matrices. Convergence study: $k\in\{2,8,32\}$, 5 seeds. Chains: Appendix~\ref{app:chains}. The budget rule for removing seeds was not triggered.

\paragraph{Numerics.}\label{app:numerics} $\min D$ uses \V{cfg.starts} starts (corners, edge midpoints, the 256 lowest-$D$ Monte Carlo samples of S2 and S4, and uniform random points), projected Adam for 400 steps, cosine learning rate from $0.005$ to $0.0001$, projected onto the support for Beta beliefs, and takes the minimum with the Monte Carlo minimum. These settings were chosen on $d=3$ coefficients against brute-force grids, not on the study's data. Random numbers come from the PCG64 generator, seeded from a fixed root seed and the indices of each condition, with common random numbers across conditions. Bootstraps use a fixed seed.

\section{Negativity and noise: further results}\label{app:neg}

\begin{figure}[htbp]
\centering
\includegraphics{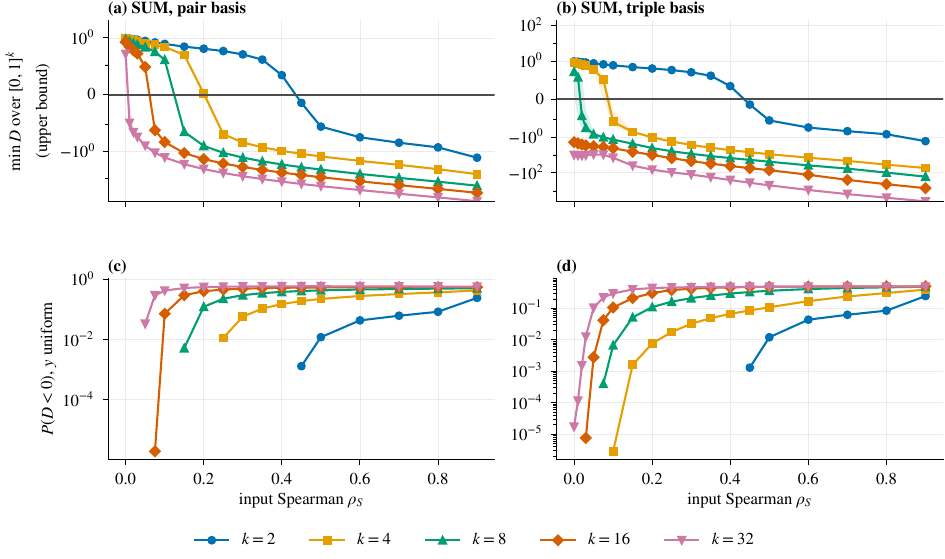}
\caption{Negativity of $D$ for type SUM, $m=4$, $n=10^7$, seed medians. (a, b) $\min D$ over $[0,1]^k$ on a symmetric logarithmic axis; the value is an upper bound on the true minimum. (c, d) $P(D<0)$ under uniform $y$ on a logarithmic axis; points with zero observed probability are not drawn.}
\label{fig:neg}
\end{figure}

\begin{figure}[htbp]
\centering
\includegraphics{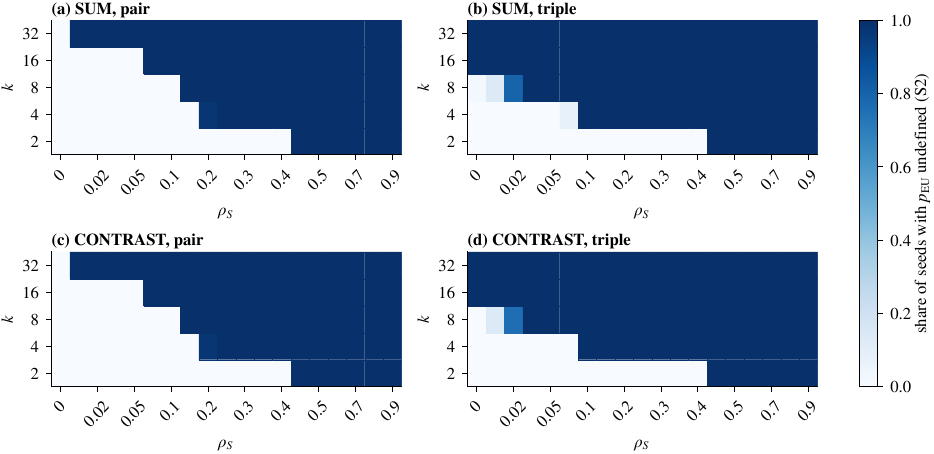}
\caption{Share of seeds with $\peu$ undefined under S2, $m=4$, $n=10^7$. The $\rs$ axis is categorical (grid values, unevenly spaced). Dark regions are where $\dden$ cannot be measured.}
\label{fig:undef}
\end{figure}

Tables~\ref{tab:onsetall}--\ref{tab:noise} extend Section~\ref{sec:fail} to all degrees and to the noise scale; Figures~\ref{fig:neg} and~\ref{fig:undef} show the measured negativity and where $\dden$ cannot be measured.

\begin{table}[htbp]
\centering
\caption{Onsets for all degrees (population, SUM, S2; post hoc for the bounds). ``first'' and ``convex'' as in Table~\ref{tab:onset}. ``search'': best upper value found by coordinate descent, capped by ``measured''. ``---'': no corner goes negative, or $k=2$, where the triple family equals the pair family for $D$. Over all $m$ the convex bound lies within \V{on.gap} of the measured onset for $k\geq4$. At $k=32$, $m=4$ its optimal measure has atoms at $0$, \V{on.atoms.k32}, \V{on.atoms.k32b} and $1$. Beyond the study's range its product $k\rs$ at $m=4$ is \V{on.conv.kr.8}, \V{on.conv.kr.32}, \V{on.conv.kr.128}, \V{on.conv.kr.512} and \V{on.conv.kr.2048} at $k=8,32,128,512,2{,}048$ (post hoc). The explicit condition on $\rs$ in Proposition~\ref{prop:pair}(a) gives \V{on.arc.2} at $k=2$ and \V{on.arc.32} at $k=32$.}
\label{tab:onsetall}
\input{generated/tab_onset_all.tex}
\end{table}

\begin{table}[htbp]
\centering
\caption{$z=(1-\min D)/\sigma_{\mathrm c}$ at $\rs=0$, $m=4$ (post hoc). ``measured'': from the study at $n=10^7$. ``corners'' and ``cube'': Gaussian surrogate of Proposition~\ref{prop:noise} by coordinate descent, medians over a few draws. ``on the boundary'': share of coordinates on the boundary in the cube minimiser.}
\label{tab:noisez}
\input{generated/tab_noise_z.tex}
\end{table}

\begin{table}[htbp]
\centering
\caption{Noise scale at $\rs=0$, $m=4$, $n=10^7$, type SUM, S2 (post hoc). $\sigma_{\mathrm c}=\sqrt{K_{\mathrm c}/n}$ from Proposition~\ref{prop:noise}; $|B_D|$ is the number of nonzero multi-indices of $D$; ``ratio'' is $(1-\min D)/\sigma_{\mathrm c}$. The triple basis at $k=2$ is omitted because it equals the pair basis for $D$.}
\label{tab:noise}
\input{generated/tab_noise.tex}
\end{table}

\FloatBarrier
\section{The gap and the errors: further results}\label{app:thr}

\begin{figure}[htbp]
\centering
\includegraphics{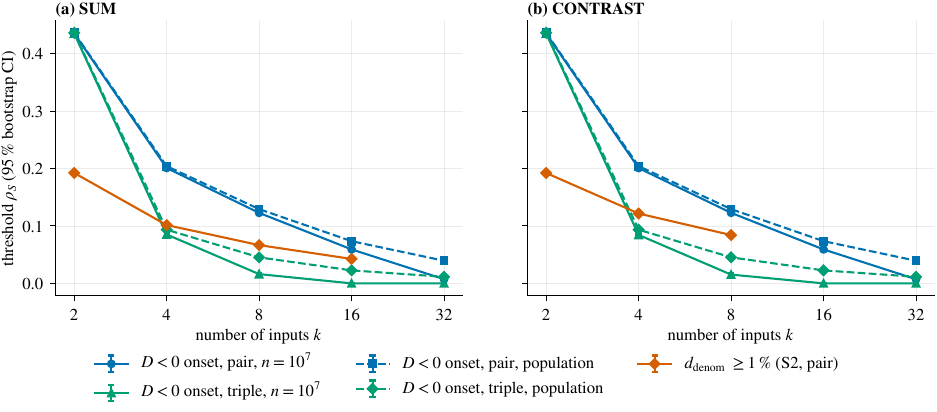}
\caption{Negativity onsets for both bases, at $n=10^7$ (solid) and with population coefficients (dashed), and the $1\,\%$ threshold of $\dden$ (S2, pair), against $k$, for $m=4$. Error bars are $95\,\%$ bootstrap CIs and are often smaller than the markers. Missing points mean the level was not reached before $\peu$ became undefined.}
\label{fig:thr}
\end{figure}

\begin{figure}[htbp]
\centering
\includegraphics{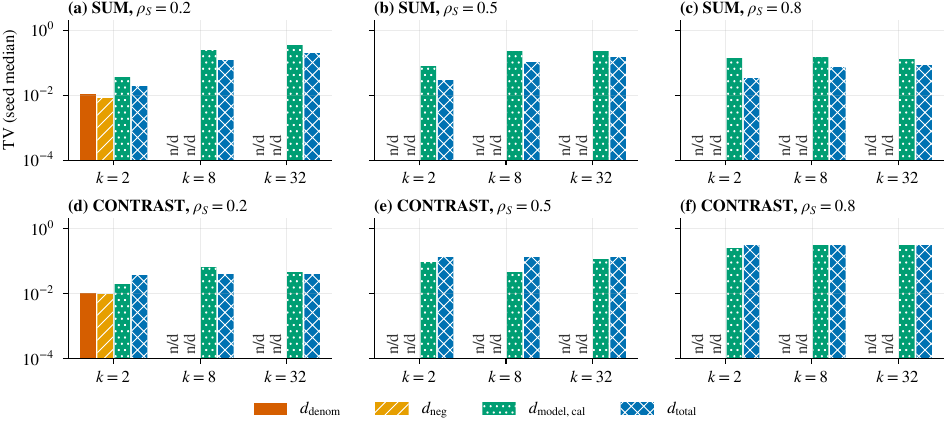}
\caption{Error components under the flat belief S2, pair basis, $m=4$, $n=10^7$, $\varepsilon=0.01$; seed medians at three correlations and three values of $k$. ``n/d'' marks quantities not defined because $\peu$ is undefined, not values near zero. $\dtot$ under S2 is dominated by the mismatch between the flat belief and the truth.}
\label{fig:dec}
\end{figure}

\begin{figure}[H]
\centering
\includegraphics{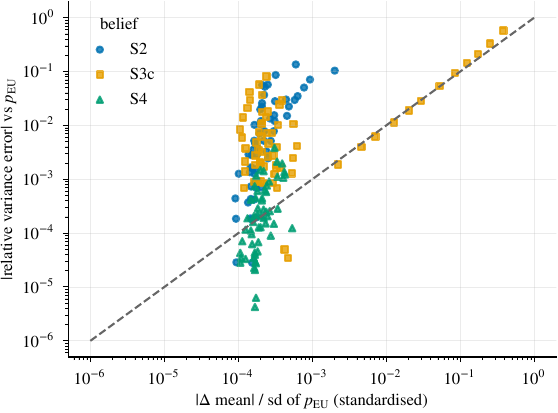}
\caption{Absolute relative variance error against standardised mean error, both relative to $\peu$, pair basis, $m=4$, $\varepsilon=0.01$, all $k$, types and $\rs>0$ where defined; seed medians. Points above the dashed line have larger variance error than mean error.}
\label{fig:mv}
\end{figure}

This appendix holds the threshold tables and error components behind Sections~\ref{sec:trunc},~\ref{sec:gapres} and~\ref{sec:clip}. Table~\ref{tab:thr} and Figure~\ref{fig:thr} are the summary; Tables~\ref{tab:thrfull}--\ref{tab:bel} add CONTRAST, confidence intervals, variance thresholds (Table~\ref{tab:relvar}) and all beliefs. Table~\ref{tab:past} shows what happens past the onset. Table~\ref{tab:decomp} and Figure~\ref{fig:dec} show how the error splits, and Figure~\ref{fig:mv} compares variance and mean errors.

\begin{table}[htbp]
\centering
\caption{Thresholds in $\rs$ for type SUM, flat belief S2, $m=4$, seed-median curves, $n=10^7$ except the population onset. ``---'' means the level was not reached before $\peu$ became undefined; it never means that the error is small. ``undefined from'' is the first grid value at which most seeds have $\peu$ undefined. Onsets are upper bounds.}
\label{tab:thr}
\input{generated/tab_thr_main.tex}
\end{table}

\begin{figure}[htbp]
\centering
\includegraphics{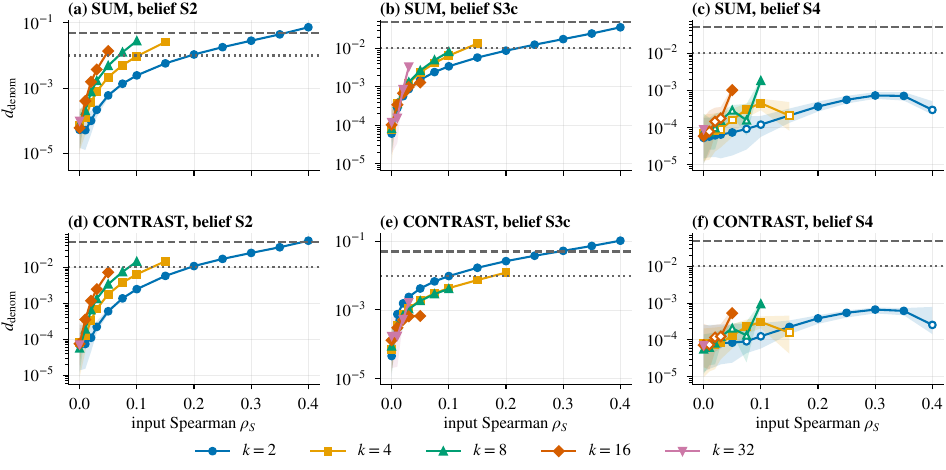}
\caption{$\dden$ against input Spearman correlation $\rs$ for the pair basis, $m=4$, $n=10^7$, $\varepsilon=0.01$, beliefs S2, S3c and S4. Lines are seed medians and bands $2.5$--$97.5\,\%$ seed intervals. A curve ends where most seeds have $\peu$ undefined; the gap carries no information about the error. Hollow markers are below three times the Monte Carlo noise floor. Dotted and dashed lines mark $1\,\%$ and $5\,\%$.}
\label{fig:d}
\end{figure}

\begin{table}[htbp]
\centering
\caption{All thresholds for S2, $m=4$, with $95\,\%$ bootstrap CIs. ``---'' as in Table~\ref{tab:thr}.}
\label{tab:thrfull}
\input{generated/tab_thr_full.tex}
\end{table}

\begin{table}[htbp]
\centering
\caption{Thresholds of the relative variance error (S2, $m=4$) and $\min D$ at $\rs=0$. ``---'' as in Table~\ref{tab:thr}.}
\label{tab:relvar}
\input{generated/tab_thr_relvar.tex}
\end{table}

\begin{table}[htbp]
\centering
\caption{$1\,\%$ threshold of $\dden$ and undefined onset (shown as threshold / onset) for all beliefs, pair basis, $m=4$. ``---'' in the onset means $\peu$ stayed defined up to $\rs=0.9$; only then does ``---'' in the threshold mean that the level was not reached in the range. ``Defined'' is relative to the protocol's support cut, whose reach changes with $k$ (Appendix~\ref{app:defs}), so do not read defined share or undefined onsets under Beta beliefs at $k\geq16$ as a trend. Selection (Section~\ref{sec:gapres}): among defined evaluations at $m=4$, $\rs>0$, $\dden$ has $99$th percentile \V{sel.q99}; over all $m$, $\CV(D)$ reaches \V{sel.cv.all} and $\dden$ \V{sel.dmax.all}. The $5\,\%$ level of $\dden$ is reached in \V{sel.d5.s2} of \V{sel.d5.s2n} S2 cells and \V{sel.d5.all} of \V{sel.d5.alln} cells over all beliefs. Under S3d the seed-median $\dden$ is at most \V{narrow.pair4} for pair at $m=4$, but \V{narrow.triple4} for triple at $m=4$ and \V{narrow.triple2} at $m=2$ (post hoc).}
\label{tab:bel}
\input{generated/tab_beliefs.tex}
\end{table}

\begin{table}[htbp]
\centering
\caption{Past the onset: evaluations with $\peu$ undefined, $\rs>0$, $m=4$, $\varepsilon=0.01$, medians (post hoc). ``CD closer than EC'': share with $\pcd$ nearer to $\pg$ than $\pec$. Here truth is the Jeffrey answer.}
\label{tab:past}
\input{generated/tab_past.tex}
\end{table}

\begin{table}[htbp]
\centering
\caption{Error components at $k=2$, pair basis, $m=4$, $n=10^7$, $\varepsilon=0.01$, seed medians.}
\label{tab:decomp}
\input{generated/tab_decomp.tex}
\end{table}

\FloatBarrier

\section{Decision rules and hypothesis details}\label{app:hyp}

The decision rules below are given in substance from the protocol. The tables and the figure that follow hold the detail behind H1c (Tables~\ref{tab:h1c} and~\ref{tab:lead}), H2a, H3a--c and H4b.

H1a, H2a, H2b (population): deterministic inequality per cell. H1b, H2b ($n=10^7$), H5a--d: aggregation rule of Section~\ref{sec:design}. H1c: power law per (type, basis), $\geq90\,\%$ of points within a factor 2 for \emph{supported}, $<50\,\%$ for \emph{not supported}. H2c: $\max/\min$ over $m$ at most $1.5$ at $\geq80\,\%$ of points for \emph{supported}, $\leq50\,\%$ for \emph{not supported}. H3a: \emph{supported} if every CI is inside $[-0.05,0.05]$, \emph{not supported} if any is entirely outside. H3b: same with the ratio in $[0.5,2]$. H3c: $\geq80\,\%$ for \emph{supported}, $<50\,\%$ for \emph{not supported}; the analysis drops matrices whose $\dden$ lies below the noise floor, as the protocol does for points in H1c, H2c and H3b. H4a: $\geq90\,\%$ for \emph{supported}, $<50\,\%$ for \emph{not supported}. H4b: every violation is listed in the released results. The two bases coincide for $D$ at $k=2$, which is why $k=2$ is not in H2b.

\begin{figure}[htbp]
\centering
\includegraphics{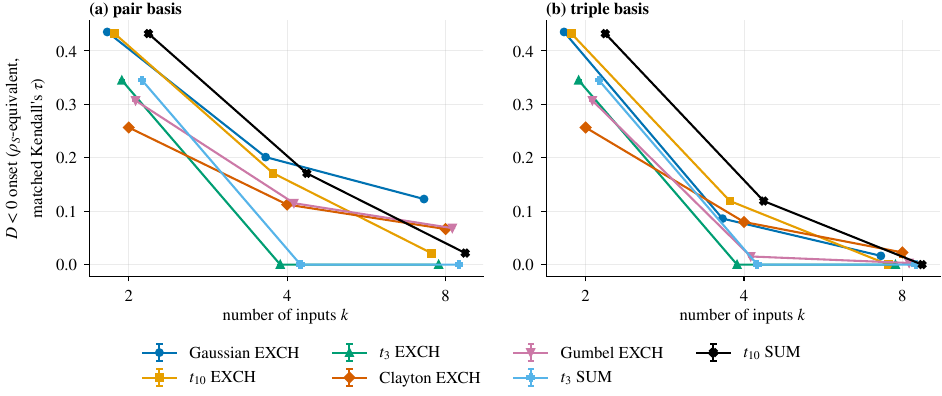}
\caption{Negativity onset by copula family at matched Kendall's $\tau$, expressed in $\rs$ units, $m=4$, $n=10^7$, 20 seeds, with $95\,\%$ bootstrap CIs (mostly smaller than the markers). Points are shifted slightly in $k$ for legibility. EXCH families and the $t$ SUM forms are shown; $t$ CONTRAST is in Table~\ref{tab:fam}. An onset of $0$ means $\min D<0$ at $\rs=0$.}
\label{fig:fam}
\end{figure}

\paragraph{Transfer in detail.}\label{app:transfer} None of the three transfer hypotheses was supported. \textbf{H3a (\verdict{\V{H3a.verdict}}).} The rule requires every onset difference from the Gaussian to have a CI inside $\pm0.05$. Of \V{H3a.n} comparisons, \V{H3a.in} were inside and \V{H3a.out} entirely outside (Table~\ref{tab:fam}, Figure~\ref{fig:fam}). Onsets were mostly earlier, with differences as low as \V{fam.t3.min} for $t_3$ and \V{fam.clayton.min} for Clayton. \textbf{H3b (\verdict{\V{H3b.verdict}}).} $\dden$ under S2 at matched Kendall's $\tau$ fell within a factor 2 of the Gaussian, with its CI, at \V{H3b.in} of \V{H3b.n} points and entirely outside at \V{H3b.out}; median ratios were \V{fam.t10.ratio} ($t_{10}$), \V{fam.clayton.ratio} (Clayton), \V{fam.t3.ratio} ($t_3$) and \V{fam.gumbel.ratio} (Gumbel). \textbf{H3c (\verdict{\V{H3c.verdict}}).} For LKJ matrices the Gaussian power law predicted $\dden$ within a factor 2 for \V{H3c.pair} (pair) and \V{H3c.triple} (triple) of the \V{H3c.n} defined matrices whose $\dden$ lies above the noise floor. This rests on $k=2$: at $k=4$ and $8$ the numbers of defined matrices were \V{lkj.defined.k4} and \V{lkj.defined.k8}, so Table~\ref{tab:lkj} lists $k=2$ only. Part of the $t$ result is consistent with structure: a $t$ copula is dependent at zero correlation, so $\rs=0$ is not independence. At $k=2$ in the pair basis $\chi_D$ at $\rs=0$ is \V{ph.t3.chi} for $t_3$ and \V{ph.t10.chi} for $t_{10}$, against \V{ph.g.chi} for the Gaussian, and $t_3$ already has $\min D={}$\V{ph.t3.minD4} at $k=4$ (post hoc). We did not test this further.

\begin{table}[htbp]
\centering
\caption{H1c fits $\dden=C\chi_D^\beta$ on population coefficients, S2, $m=4$. Cell verdicts follow the H1c rule.}
\label{tab:h1c}
\input{generated/tab_h1c.tex}
\end{table}

\begin{table}[htbp]
\centering
\caption{H1c points against the pre-registered power law and the closed forms (post hoc). ``power law'' and ``closed form'' give the share of points within a factor 2 of the fit and of the formula; ``ratio'' is measured $\dden$ over the formula.}
\label{tab:lead}
\input{generated/tab_lead.tex}
\end{table}

\begin{table}[htbp]
\centering
\caption{Population negativity onset by degree (H2a), type SUM. At $k=2$ the two bases coincide for $D$.}
\label{tab:onsetm}
\input{generated/tab_onset_m.tex}
\end{table}

\begin{table}[htbp]
\centering
\caption{Comparison with the Gaussian copula at matched Kendall's $\tau$ ($m=4$, pair and triple, $k\in\{2,4,8\}$, $n=10^7$, 20 seeds). Columns: smallest and largest onset difference from the Gaussian; number of $95\,\%$ CIs inside and entirely outside $\pm0.05$; median ratio of $\dden$ to Gaussian under S2.}
\label{tab:fam}
\input{generated/tab_families.tex}
\end{table}

\begin{table}[htbp]
\centering
\caption{H3c by $k$ and $\eta$. ``defined'' counts matrices with $\peu$ defined (pair basis, S2, $m=4$). Shares are among defined matrices with $\dden$ above the noise floor. The rule is at least $80\,\%$ for \emph{supported} and below $50\,\%$ for \emph{not supported}.}
\label{tab:lkj}
\input{generated/tab_lkj.tex}
\end{table}

\begin{table}[htbp]
\centering
\caption{Convergence test H4b on the subset of the convergence study ($k\in\{2,8,32\}$, 5 seeds). Paired values are reported TVs before and after doubling one setting; a violation is a change of at least $\max(3\,\mathrm{SE},10^{-3})$. The final column is the largest ratio of change to tolerance. The \V{H4b.viol} violations, \V{H4b.share} of all paired values, fall at $k=8$ (\V{H4b.k8}) and $k=32$ (\V{H4b.k32}). Post hoc: \V{ph.h4b.triple} of them are triple, with median relative change \V{ph.h4b.rel}, and the largest change runs from \V{ph.h4b.worst.from} to \V{ph.h4b.worst.to}, at $k=32$, triple, $m=6$, S3d, a signed output.}
\label{tab:conv}
\input{generated/tab_conv.tex}
\end{table}

\FloatBarrier
\section{Chains and training}\label{app:chains}\label{app:train}

This appendix defines the chain experiment and gives the detail behind Section~\ref{sec:depth}.

\paragraph{Ground truth.} The ground truth is a Markov chain in normal scores. The first hidden variable depends on $y$ through a Gaussian-copula neuron of the sweep; for $l\geq 2$, $s_{h_l}=\alpha\,s_{h_{l-1}}+\gamma\,s_{w_l}+\sqrt{1-\alpha^2-\gamma^2}\,\nu_l$ with $\alpha=\V{cfg.a}$, $\gamma=\V{cfg.b}$, fresh independent side inputs $w_l$ and noise $\nu_l$. Neuron $l\geq2$ is a triple-basis model of $(h_l;h_{l-1},w_l)$, that is $k=2$. Neuron 1 uses pair or triple. Depth is $L\leq 4$, with $k\in\{2,4,8\}$, $\rs\in\{0,0.2,0.4,0.6\}$, $m=4$ and 10 seeds. Estimation uses $n=10^6$; training, validation and test splits have $2\cdot10^5$, $10^5$ and $10^5$ samples. Between layers the only normalisation is $c_0=1$, the first coefficient of the propagated polynomial; there is no re-mapping of variables to quantiles.\looseness=-1 

\paragraph{Training.} Training uses three losses: value MSE as in Eq.~7 of~\citep{duda2024biology}, calibrated negative log-likelihood (NLL), and end-to-end squared $L^2$ distance between the $\pcd$ output and the projected target, summed over S2, S3c and S4. It uses two optimisers, Adam~\citep{kingma2015adam} and L-BFGS~\citep{liu1989limited}, each with fixed, untuned settings. Adam: learning rate $0.003$, cosine decay to 0, at most $2{,}000$ full-batch steps, validation every 50 steps, patience 10. L-BFGS: strong Wolfe line search, learning rate 1, 20 iterations per step, history 20, at most 50 steps, validation every step, patience 10. The best validation state is kept. $\varepsilon=0.01$ in the likelihood loss. The value loss uses rows $i\in\{0,1\}$; an all-rows variant is identical because its extra gradients vanish exactly (validation test M14, Appendix~\ref{app:validation}). The end-to-end loss uses $2^{16}$ samples per belief per split. The end-to-end run for depth $L$ trains the first $L$ neurons and evaluates them followed by the untrained remainder. A flat-$D$ control starts from $a_{0\mathbf{j}}=0$ for $\mathbf{j}\neq0$.

\begin{table}[htbp]
\centering
\caption{Depth results by belief and neuron-1 basis, medians over $k$, $\rs>0$ and seeds. Columns: $\dtot$ at each layer; variance of $\pg$ at $L=1$ and $L=4$; median H5a ratio over cells including $\rs=0$; numbers of H5a cells significantly above 1, significantly below 1, and total. The breakdown is post hoc; the verdict uses all \V{H5a.n} cells.}
\label{tab:depth}
\input{generated/tab_depth.tex}
\end{table}

\begin{table}[htbp]
\centering
\caption{Training runs from direct estimates, medians over runs (post hoc). ``kept initial'': share of runs whose best validation state is the initial one. ``coef.\ dist.'': coefficient distance moved. ``test loss'': relative change of the training loss on the test split, and the share of runs with a decrease. $\Delta P(D<0)$ is taken where $P(D<0)$ was positive. $\Delta\min D$ is the change in $\min D$.}
\label{tab:training}
\input{generated/tab_training.tex}
\end{table}

\begin{figure}[htbp]
\centering
\includegraphics{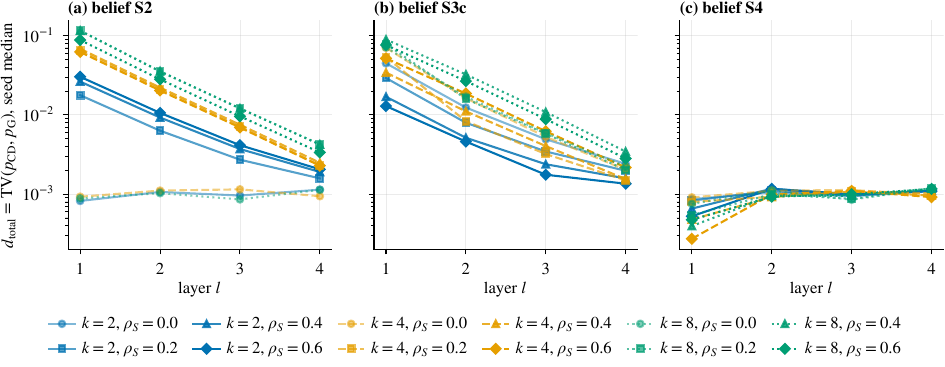}
\caption{$\dtot$ by layer for chains with direct estimates: seed medians over 10 seeds, neuron 1 in the pair basis, $m=4$, $n=10^6$, for $k\in\{2,4,8\}$ and $\rs\in\{0,0.2,0.4,0.6\}$. The decline under S2 and S3c reflects the target approaching the uniform density, not improved accuracy. Values near $10^{-3}$ are at Monte Carlo noise level.}
\label{fig:depth}
\end{figure}

\begin{figure}[htbp]
\centering
\includegraphics{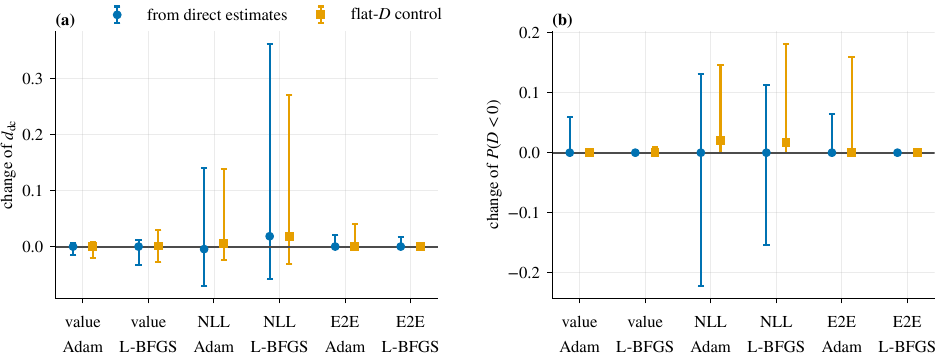}
\caption{Effect of training on neuron 1 under S2. (a) Change of $\ddc=\TV(\pcd,\pec)$, shown because it exists everywhere; this is not the pre-registered $\dden$ of H5b. (b) Change of $P(D<0)$. Points are medians and bars $10$--$90\,\%$ ranges over all $k$, $\rs$, bases and seeds, from direct estimates and from the flat-$D$ control. E2E: end-to-end training, the runs with $L=4$.}
\label{fig:train}
\end{figure}

\paragraph{Results in detail.} \textbf{H5a (\verdict{\V{H5a.verdict}}).} The ratio $\dtot(L=4)/\dtot(L=1)$ exceeded 1 significantly in \V{H5a.sig} of \V{H5a.n} cells and was significantly below 1 in \V{H5a.opp}. Table~\ref{tab:depth} and Figure~\ref{fig:depth} give median ratios of \V{H5a.ratio.S2} (S2), \V{H5a.ratio.S3c} (S3c) and \V{H5a.ratio.S4} (S4). This is not an improvement in itself, because the target moves: under S2 the variance of $\pg$ rises towards the uniform value $1/12$ (from \V{ph.depth.var.S2.pair.1} to \V{ph.depth.var.S2.pair.4}), and $\dtot$ falls because the true output forgets the belief. A variance-type deviation shrinks by $\alpha^6=\V{ph.a6}$ over three layers, close to the S2 ratio, which is consistent with contraction under a mixing kernel~\citep{boyen1998tractable}, not something we tested. Under S4 the pair ratio rises to \V{ph.h5a.S4.pair.ratio} (\V{ph.h5a.S4.pair.sig} of \V{ph.h5a.S4.pair.n} cells significant), while the triple ratio falls to \V{ph.h5a.S4.triple.ratio} (post hoc); the significant cells are small (paragraph H5a below). Inputs of later neurons are independent, so their $D$ is nearly flat (median $\min D$ \V{ph.layer.minD}): H5a does not test compounding of denominator error over correlated inputs. \textbf{H5b (\verdict{\V{H5b.verdict}}), H5c (\verdict{\V{H5c.verdict}}), H5d (\verdict{\V{H5d.verdict}}).} Training reduced $\dden$ of neuron 1 significantly in \V{H5b.sig} of \V{H5b.n} evaluable cells (\V{H5b.opp} opposite), and only $k=2$ is evaluable. It reduced $P(D<0)$ in \V{H5c.sig} of \V{H5c.n} cells (\V{H5c.opp} opposite), and held-out $\dtot$ in \V{H5d.sig} of \V{H5d.n} cells (\V{H5d.opp} opposite); the training runs are summarised in Table~\ref{tab:training} and Figure~\ref{fig:train}. Three facts limit these verdicts. About half of the end-to-end runs at $L=4$ kept the initial state (\V{ph.tr.e2e4.adam.kept} with Adam, \V{ph.tr.e2e4.lbfgs.kept} with L-BFGS), and their loss before training, \V{e2e.L4}, is already near the noise level of the target, \V{e2e.noise}, so those runs could not show learning; they supply \V{ph.H5b.e2e.n} of the \V{H5b.n} H5b cells and \V{ph.H5c.e2e.n} of the \V{H5c.n} H5c cells. Optimiser settings were fixed and untuned. And $\dden$ can be evaluated only at $k=2$. We conclude that training did not reduce these errors under these settings. The data do not show that training cannot compensate for the approximations, and the HCR paper's expectation that training compensates for neglected calibration remains untested.

\paragraph{H5a.} The significant H5a cells all have tiny errors: $\dtot(L=4)\leq\V{ph.h5a.sig.maxL4}$, at most \V{ph.h5a.sig.floor} times the noise floor (so below the flagging level), and \V{ph.h5a.sig.rho0} are at $\rs=0$ (post hoc). Lowest $\min D$ of later layers is \V{ph.layer.minDlow}.

\paragraph{H5b by loss (post hoc).} NLL \V{ph.H5b.nll.sig} of \V{ph.H5b.nll.n}, value \V{ph.H5b.val.sig} of \V{ph.H5b.val.n}, end-to-end \V{ph.H5b.e2e.sig} of \V{ph.H5b.e2e.n}.

\paragraph{H5c by loss (post hoc).} NLL \V{ph.H5c.nll.sig} of \V{ph.H5c.nll.n} (\V{ph.H5c.nll.opp} opposite), value \V{ph.H5c.val.sig} of \V{ph.H5c.val.n} (\V{ph.H5c.val.opp} opposite), end-to-end \V{ph.H5c.e2e.sig} of \V{ph.H5c.e2e.n} (\V{ph.H5c.e2e.opp} opposite). NLL lowers the negative mass (median change of $P(D<0)$ $-\V{ph.tr.nll.adam.dP}$ with Adam, $-\V{ph.tr.nll.lbfgs.dP}$ with L-BFGS, where it was positive) while deepening the minimum ($\min D$ falls by \V{ph.tr.nll.adam.dminD} and \V{ph.tr.nll.lbfgs.dminD}).

\paragraph{H5d.} The count rises with $k$ (\V{ph.H5d.k.2.sig}, \V{ph.H5d.k.4.sig}, \V{ph.H5d.k.8.sig} of \V{ph.H5d.k.2.n} each at $k=2,4,8$) and is larger in the triple basis (\V{ph.H5d.basis.triple.sig} against \V{ph.H5d.basis.pair.sig} of \V{ph.H5d.basis.pair.n}). At $L=1$, median $\dtot$ falls on the training beliefs (S2 by \V{ph.e2e.1_adam.S2} with Adam and \V{ph.e2e.1_lbfgs.S2} with L-BFGS, S3c by \V{ph.e2e.1_adam.S3c} and \V{ph.e2e.1_lbfgs.S3c}), and rises on held-out beliefs (by \V{ph.e2e.1_adam.held} and \V{ph.e2e.1_lbfgs.held}), so the optimiser did learn (post hoc). The cause of the held-out result is the design: the objective fits \V{e2e.targets} numbers ($3$ beliefs $\times$ $4$ moments) with \V{e2e.coef.lo} to \V{e2e.coef.hi} coefficients, and validation uses the same beliefs. So H5d says little about whether training can absorb the denominator, and a fair test would need a distribution of training beliefs.

There were no non-finite losses in the \V{ph.tr.runs} training runs from direct estimates. The flat-$D$ control starts with $\dden=0$; after NLL training it has median $P(D<0)=\V{ph.flat.nll.P}$ and $\chi_D=\V{ph.flat.nll.chi}$, so likelihood training moves a flat model into the negative regime.

\FloatBarrier
\section{Validation}\label{app:validation}

Fourteen validation tests check the implementation (Table~\ref{tab:val}). All code-validation tests (M1--M12, M14) pass on an A100 with a single version of the code. M13 is not a code test: it is hypothesis H4b on the study's data and appears in Section~\ref{sec:conv} as \verdict{\V{H4b.verdict}}.

\begin{table}[htbp]
\centering
\caption{Validation tests. M13 ``fail'' is the H4b outcome.}
\label{tab:val}
\input{generated/tab_validation.tex}
\end{table}

For instance, M6 checks Eq.~\eqref{eq:identity} in sample: the maximum residual is $\V{val.M6}$ over \V{val.M6.n} defined cases. M9 finds a largest difference of $\V{val.M9}$ between the optimiser and reference minima. M4 reproduces \V{val.M4.n} TV values of the $d=3$ study, with worst difference over tolerance \V{val.M4.worst}.

\paragraph{Note on M7.} The Student-$t$ part of M7 was rewritten after the main runs. The earlier check compared $1-F$ with SciPy~\citep{virtanen2020scipy} and could not resolve tails near $F=1$; it failed on CPU and GPU alike with SciPy \V{sw.scipy}. The reference is now 50-digit mpmath~\citep{mpmath2026}, whose version was not pinned: relative error $\V{val.M7.low}$ in the lower tail, absolute error $\V{val.M7.up}$ on the upper side, and relative error $\V{val.M7.q}$ for the quantile's tail probability. The study's code did not change; the test changed.

\section{Time and energy}\label{app:energy}

This separate descriptive benchmark never enters the error tables. It gives the cost of what the moment contraction avoids. It compares one forward pass of four parameter-matched layers of width 64: HCR-CD (Eq.~\eqref{eq:exact}, pair, $m=4$), HCR-CAL (calibrated conditional, $S=64$ samples per belief, \V{cfg.gl} $x$-nodes), an MLP layer with GELU, and a KAN layer~\citep{liu2025kan} (grid 5, order 3). Settings are $k\in\{2,8,32\}$ and batch size $16{,}384$ or $256$, in \V{en.runs} runs (\V{en.runs.dtype} per precision) of 30 repeats of about 2\,s, with NVML power sampling at 10\,ms on one A100 type. Table~\ref{tab:energy} gives the main setting, and Tables~\ref{tab:e32} and~\ref{tab:e64} give all settings.

\begin{table}[htbp]
\centering
\caption{One forward pass at $k=8$, batch size $16{,}384$, median over the \V{en.row.repeats} repeats of \V{en.runs.dtype} runs per precision, energy including idle power. A descriptive microbenchmark of unoptimised code; see the caveats in the text. The layers do different jobs, and HCR-CAL uses $S=64$ samples per belief, so the lines are not like-for-like.}
\label{tab:energy}
\input{generated/tab_energy_main.tex}
\end{table}

At this setting HCR-CD uses \V{en.cd.over.mlp.32} times the energy of the MLP layer in float32 and \V{en.cd.over.mlp.64} times in float64, KAN uses \V{en.kan.over.cd.32} and \V{en.kan.over.cd.64} times that of HCR-CD, and HCR-CAL uses \V{en.cal.over.cd.32} and \V{en.cal.over.cd.64} times. We draw no general claim that HCR is more or less efficient than the other layers. The comparison is parameter-matched, forward only and inference-shaped, on one GPU type with clocks not locked, and all implementations are unoptimised. The layers compute different things: an MLP returns values, and HCR-CD returns a density. The factor for HCR-CAL reflects $S=64$ samples and many $x$-nodes per belief, a design choice, and not a like-for-like cost.

Measurement quality was acceptable for this purpose. The energy counter agrees with integrated sampled power at median ratio \V{en.counter.med} (1st--99th percentile \V{en.counter.lo}--\V{en.counter.hi}) over \V{en.repeats} repeats, which matters because built-in sensors have known sampling problems~\citep{burtscher2014measuring}, including readings on the A100 that cover only part of the runtime~\citep{yang2024accurate}. Idle power dominates \V{en.idle.n} of \V{en.idle.of} configurations. Run-to-run differences follow the physical GPU, with one GPU \V{en.rel.hi1}--\V{en.rel.hi2}\,\% above the others, in two runs only. The float64 MLP at $k=32$, batch size $16{,}384$, draws \V{en.mlp.limit.W}\,W, at the 400\,W power limit, so \V{en.limit.repeats} repeats ran at $97.5\,\%$ of the limit or above. Also, float64 is faster than float32 for the MLP at $k=2$, batch size $16{,}384$ (\V{en.mlpfast.64}\,ms against \V{en.mlpfast.32}\,ms), an effect of cuBLAS kernel selection that a separate probe confirmed. On reporting energy in machine learning, see~\citep{henderson2020towards}.

\begin{table}[htbp]
\centering
\caption{Float32 forward pass: median over \V{en.row.repeats} repeats (\V{en.runs.dtype} runs) of time, energy, energy net of idle, power, SM clock, and the ratio of the energy counter to sampled energy.}
\label{tab:e32}
\input{generated/tab_energy_float32.tex}
\end{table}

\begin{table}[htbp]
\centering
\caption{Float64, as Table~\ref{tab:e32}.}
\label{tab:e64}
\input{generated/tab_energy_float64.tex}
\end{table}

\FloatBarrier
\section{The earlier three-input study}\label{app:d3}

This separate study is not pre-registered and motivated the present design; it uses $d=3$ variables, that is $k=2$, with $m=4$. With a flat belief at $\rs=0.2$, $\dden$ was \V{d3.flat.sum} (SUM) and \V{d3.flat.con} (CONTRAST), and for a narrow belief at $\rs=0.3$ it was \V{d3.narrow.sum} and \V{d3.narrow.con}. Over \V{d3.closer.n} evaluations where $\peu$ was undefined, $\pcd$ was closer to $\pg$ than $\pec$ in \V{d3.closer}. A Taylor-corrected denominator with $0$, $1$ or $2$ terms (SUM, S2) gave TV to $\peu$ of \V{d3.taylor.r1k0}, \V{d3.taylor.r1k1}, \V{d3.taylor.r1k2} at $\rs=0.1$, and \V{d3.taylor.r4k0}, \V{d3.taylor.r4k1}, \V{d3.taylor.r4k2} at $\rs=0.4$: it helps at small $\rs$ and hurts once the series condition fails. These values come from a different setup and should not be compared directly with the main results.

\section{Reproducibility}\label{app:repro}

\paragraph{Batches.} The study ran in \V{batches.n} batches (Table~\ref{tab:batches}). The only computational change between them was that the Student-$t$ distribution came to be evaluated on the device; the other changes concerned scheduling and cost accounting. All $t$ runs came after that change. Validation and the energy benchmark ran on later versions of the code than the main runs.\looseness=-1 

\begin{table}[htbp]
\centering
\caption{Runs and GPU-hours by experiment, and runs by batch.}
\label{tab:batches}
\input{generated/tab_batches.tex}
\end{table}

\paragraph{Seeds.} Listed in Section~\ref{sec:design} and Appendix~\ref{app:defs}; the throughput run of Section~\ref{sec:design} is excluded. \textbf{Hardware.} NVIDIA A100-SXM4-40GB, driver \V{sw.driver}, float64, TF32 disabled, deterministic algorithms; the computations of Sections~\ref{sec:semantics}--\ref{sec:fail}, the regression message and the pilot run on a CPU in minutes. \textbf{Software.} PyTorch~\citep{paszke2019pytorch} \V{sw.torch}, NumPy \V{sw.numpy}, SciPy \V{sw.scipy}, all under BSD-style licences. The study uses no external data set. The pilot uses California housing through scikit-learn \V{pilot.sklearn}, in a separate environment.

\paragraph{Regeneration.} The released code identifies each version and regenerates every number, table and figure. Every result in the text comes from generated macros and tables, and a separate pass re-derives them from the study's outputs.

\end{document}

%% file: generated/numbers.tex
\expandafter\def\csname v@H1a.distinct\endcsname{2}
\expandafter\def\csname v@H1a.n\endcsname{4}
\expandafter\def\csname v@H1a.verdict\endcsname{supported}
\expandafter\def\csname v@H1b.n\endcsname{4}
\expandafter\def\csname v@H1b.sp.hi\endcsname{-0.428}
\expandafter\def\csname v@H1b.sp.k2\endcsname{0.435}
\expandafter\def\csname v@H1b.sp.k32\endcsname{0.007}
\expandafter\def\csname v@H1b.sp.lo\endcsname{-0.429}
\expandafter\def\csname v@H1b.verdict\endcsname{supported}
\expandafter\def\csname v@H1c.CONTRAST/pair\endcsname{0.52}
\expandafter\def\csname v@H1c.CONTRAST/triple\endcsname{0.48}
\expandafter\def\csname v@H1c.SUM/pair\endcsname{0.55}
\expandafter\def\csname v@H1c.SUM/triple\endcsname{1.00}
\expandafter\def\csname v@H1c.cells\endcsname{4}
\expandafter\def\csname v@H1c.cells.sup\endcsname{1}
\expandafter\def\csname v@H1c.verdict\endcsname{inconclusive}
\expandafter\def\csname v@H1c.verdict.CONTRAST/pair\endcsname{inconclusive}
\expandafter\def\csname v@H1c.verdict.CONTRAST/triple\endcsname{not supported}
\expandafter\def\csname v@H1c.verdict.SUM/pair\endcsname{inconclusive}
\expandafter\def\csname v@H1c.verdict.SUM/triple\endcsname{supported}
\expandafter\def\csname v@H2a.distinct\endcsname{10}
\expandafter\def\csname v@H2a.n\endcsname{20}
\expandafter\def\csname v@H2a.sig\endcsname{20}
\expandafter\def\csname v@H2a.verdict\endcsname{supported}
\expandafter\def\csname v@H2b_gauss.n\endcsname{8}
\expandafter\def\csname v@H2b_gauss.sig\endcsname{8}
\expandafter\def\csname v@H2b_gauss.verdict\endcsname{supported}
\expandafter\def\csname v@H2b_population.distinct\endcsname{4}
\expandafter\def\csname v@H2b_population.n\endcsname{8}
\expandafter\def\csname v@H2b_population.sig\endcsname{8}
\expandafter\def\csname v@H2b_population.verdict\endcsname{supported}
\expandafter\def\csname v@H2c.n\endcsname{221}
\expandafter\def\csname v@H2c.share\endcsname{60\,\%}
\expandafter\def\csname v@H2c.verdict\endcsname{inconclusive}
\expandafter\def\csname v@H3a.in\endcsname{21}
\expandafter\def\csname v@H3a.n\endcsname{48}
\expandafter\def\csname v@H3a.out\endcsname{27}
\expandafter\def\csname v@H3a.verdict\endcsname{not supported}
\expandafter\def\csname v@H3b.in\endcsname{73}
\expandafter\def\csname v@H3b.n\endcsname{236}
\expandafter\def\csname v@H3b.out\endcsname{156}
\expandafter\def\csname v@H3b.verdict\endcsname{not supported}
\expandafter\def\csname v@H3c.n\endcsname{272}
\expandafter\def\csname v@H3c.pair\endcsname{39\,\%}
\expandafter\def\csname v@H3c.triple\endcsname{34\,\%}
\expandafter\def\csname v@H3c.verdict\endcsname{not supported}
\expandafter\def\csname v@H4a.n\endcsname{114}
\expandafter\def\csname v@H4a.share\endcsname{93\,\%}
\expandafter\def\csname v@H4a.verdict\endcsname{supported}
\expandafter\def\csname v@H4b.k32\endcsname{4{,}962}
\expandafter\def\csname v@H4b.k8\endcsname{498}
\expandafter\def\csname v@H4b.share\endcsname{0.47\,\%}
\expandafter\def\csname v@H4b.total\endcsname{1{,}165{,}059}
\expandafter\def\csname v@H4b.verdict\endcsname{not supported}
\expandafter\def\csname v@H4b.viol\endcsname{5{,}460}
\expandafter\def\csname v@H5a.n\endcsname{72}
\expandafter\def\csname v@H5a.opp\endcsname{51}
\expandafter\def\csname v@H5a.ratio.S2\endcsname{0.05}
\expandafter\def\csname v@H5a.ratio.S3c\endcsname{0.06}
\expandafter\def\csname v@H5a.ratio.S4\endcsname{1.25}
\expandafter\def\csname v@H5a.sig\endcsname{11}
\expandafter\def\csname v@H5a.verdict\endcsname{not supported}
\expandafter\def\csname v@H5b.n\endcsname{18}
\expandafter\def\csname v@H5b.opp\endcsname{1}
\expandafter\def\csname v@H5b.sig\endcsname{5}
\expandafter\def\csname v@H5b.verdict\endcsname{not supported}
\expandafter\def\csname v@H5c.n\endcsname{78}
\expandafter\def\csname v@H5c.opp\endcsname{19}
\expandafter\def\csname v@H5c.sig\endcsname{24}
\expandafter\def\csname v@H5c.verdict\endcsname{not supported}
\expandafter\def\csname v@H5d.n\endcsname{192}
\expandafter\def\csname v@H5d.opp\endcsname{84}
\expandafter\def\csname v@H5d.sig\endcsname{27}
\expandafter\def\csname v@H5d.verdict\endcsname{not supported}
\expandafter\def\csname v@batches.n\endcsname{5}
\expandafter\def\csname v@cfg.M\endcsname{2^{20}}
\expandafter\def\csname v@cfg.R\endcsname{16}
\expandafter\def\csname v@cfg.a\endcsname{0.6}
\expandafter\def\csname v@cfg.b\endcsname{0.5}
\expandafter\def\csname v@cfg.floor\endcsname{0.01}
\expandafter\def\csname v@cfg.gl\endcsname{192}
\expandafter\def\csname v@cfg.snodes\endcsname{801}
\expandafter\def\csname v@cfg.starts\endcsname{2{,}000}
\expandafter\def\csname v@clip.S4.ec\endcsname{0.024}
\expandafter\def\csname v@clip.S4.eu\endcsname{0.0004}
\expandafter\def\csname v@clip.dneg.0.001\endcsname{0.024}
\expandafter\def\csname v@clip.dneg.0.01\endcsname{0.025}
\expandafter\def\csname v@clip.dneg.0.1\endcsname{0.028}
\expandafter\def\csname v@clip.max\endcsname{39\,\%}
\expandafter\def\csname v@clip.min\endcsname{0\,\%}
\expandafter\def\csname v@coef.pair\endcsname{8{,}581}
\expandafter\def\csname v@coef.triple\endcsname{357{,}765}
\expandafter\def\csname v@cor.min\endcsname{0.32}
\expandafter\def\csname v@cut.drop\endcsname{\ensuremath{3\times 10^{-4}}}
\expandafter\def\csname v@cut.in.hi\endcsname{0.79}
\expandafter\def\csname v@cut.in.lo\endcsname{0.52}
\expandafter\def\csname v@cut.sign.2\endcsname{14\,\%}
\expandafter\def\csname v@cut.sign.32\endcsname{100\,\%}
\expandafter\def\csname v@d3.closer\endcsname{32\,\%}
\expandafter\def\csname v@d3.closer.n\endcsname{145}
\expandafter\def\csname v@d3.flat.con\endcsname{0.041}
\expandafter\def\csname v@d3.flat.sum\endcsname{0.020}
\expandafter\def\csname v@d3.narrow.con\endcsname{0.0039}
\expandafter\def\csname v@d3.narrow.sum\endcsname{0.0005}
\expandafter\def\csname v@d3.taylor.r1k0\endcsname{0.0116}
\expandafter\def\csname v@d3.taylor.r1k1\endcsname{0.0022}
\expandafter\def\csname v@d3.taylor.r1k2\endcsname{0.0009}
\expandafter\def\csname v@d3.taylor.r4k0\endcsname{0.0353}
\expandafter\def\csname v@d3.taylor.r4k1\endcsname{0.0412}
\expandafter\def\csname v@d3.taylor.r4k2\endcsname{0.0758}
\expandafter\def\csname v@db.hi\endcsname{1.01}
\expandafter\def\csname v@db.lo\endcsname{0.97}
\expandafter\def\csname v@db.n\endcsname{85}
\expandafter\def\csname v@e2e.L4\endcsname{\ensuremath{2.4\times 10^{-4}}}
\expandafter\def\csname v@e2e.coef.hi\endcsname{6{,}364}
\expandafter\def\csname v@e2e.coef.lo\endcsname{61}
\expandafter\def\csname v@e2e.noise\endcsname{\ensuremath{1.8\times 10^{-4}}}
\expandafter\def\csname v@e2e.targets\endcsname{12}
\expandafter\def\csname v@en.cal.over.cd.32\endcsname{647}
\expandafter\def\csname v@en.cal.over.cd.64\endcsname{828}
\expandafter\def\csname v@en.cd.over.mlp.32\endcsname{2.0}
\expandafter\def\csname v@en.cd.over.mlp.64\endcsname{1.9}
\expandafter\def\csname v@en.counter.hi\endcsname{1.061}
\expandafter\def\csname v@en.counter.lo\endcsname{0.953}
\expandafter\def\csname v@en.counter.med\endcsname{0.999}
\expandafter\def\csname v@en.idle.n\endcsname{8}
\expandafter\def\csname v@en.idle.of\endcsname{48}
\expandafter\def\csname v@en.kan.over.cd.32\endcsname{3.5}
\expandafter\def\csname v@en.kan.over.cd.64\endcsname{4.1}
\expandafter\def\csname v@en.limit.repeats\endcsname{71}
\expandafter\def\csname v@en.mlp.limit.W\endcsname{398}
\expandafter\def\csname v@en.mlpfast.32\endcsname{0.145}
\expandafter\def\csname v@en.mlpfast.64\endcsname{0.0667}
\expandafter\def\csname v@en.rel.hi1\endcsname{5}
\expandafter\def\csname v@en.rel.hi2\endcsname{6}
\expandafter\def\csname v@en.repeats\endcsname{4{,}320}
\expandafter\def\csname v@en.row.repeats\endcsname{90}
\expandafter\def\csname v@en.runs\endcsname{6}
\expandafter\def\csname v@en.runs.dtype\endcsname{3}
\expandafter\def\csname v@eqd.n\endcsname{800}
\expandafter\def\csname v@eqd.rel\endcsname{0.5\,\%}
\expandafter\def\csname v@eqd.s3c\endcsname{0.77}
\expandafter\def\csname v@failed\endcsname{0}
\expandafter\def\csname v@fam.clayton.min\endcsname{$-$0.179}
\expandafter\def\csname v@fam.clayton.ratio\endcsname{5.34}
\expandafter\def\csname v@fam.gumbel.ratio\endcsname{16.60}
\expandafter\def\csname v@fam.t10.ratio\endcsname{2.08}
\expandafter\def\csname v@fam.t3.min\endcsname{$-$0.201}
\expandafter\def\csname v@fam.t3.ratio\endcsname{7.22}
\expandafter\def\csname v@floor.shift\endcsname{0.045}
\expandafter\def\csname v@gpuh.billed\endcsname{196.25}
\expandafter\def\csname v@gpuh.jobs\endcsname{176.7}
\expandafter\def\csname v@hyp.inc\endcsname{2}
\expandafter\def\csname v@hyp.not\endcsname{8}
\expandafter\def\csname v@hyp.rows\endcsname{16}
\expandafter\def\csname v@hyp.sup\endcsname{5}
\expandafter\def\csname v@hyp.total\endcsname{15}
\expandafter\def\csname v@jobs\endcsname{2{,}900}
\expandafter\def\csname v@lead.hi\endcsname{1.76}
\expandafter\def\csname v@lead.lo\endcsname{0.57}
\expandafter\def\csname v@lead.n\endcsname{112}
\expandafter\def\csname v@lead.small.hi\endcsname{1.21}
\expandafter\def\csname v@lead.small.lo\endcsname{0.72}
\expandafter\def\csname v@lead.small.n\endcsname{64}
\expandafter\def\csname v@lkj.defined.k4\endcsname{0}
\expandafter\def\csname v@lkj.defined.k8\endcsname{0}
\expandafter\def\csname v@narrow.pair4\endcsname{0.0032}
\expandafter\def\csname v@narrow.triple2\endcsname{0.0211}
\expandafter\def\csname v@narrow.triple4\endcsname{0.0124}
\expandafter\def\csname v@nz.bound.hi\endcsname{77\,\%}
\expandafter\def\csname v@nz.bound.lo\endcsname{75\,\%}
\expandafter\def\csname v@nz.need.pair32\endcsname{\ensuremath{8.6\times 10^{6}}}
\expandafter\def\csname v@nz.need.triple16\endcsname{\ensuremath{1.1\times 10^{8}}}
\expandafter\def\csname v@nz.pair32.corner\endcsname{4.4}
\expandafter\def\csname v@nz.pair32.cube\endcsname{5.5}
\expandafter\def\csname v@nz.pair32.meas\endcsname{4.4}
\expandafter\def\csname v@on.arc.2\endcsname{0.0398}
\expandafter\def\csname v@on.arc.32\endcsname{0.0025}
\expandafter\def\csname v@on.atoms.k32\endcsname{0.18}
\expandafter\def\csname v@on.atoms.k32b\endcsname{0.82}
\expandafter\def\csname v@on.conv.kr.128\endcsname{1.319}
\expandafter\def\csname v@on.conv.kr.2048\endcsname{1.325}
\expandafter\def\csname v@on.conv.kr.32\endcsname{1.263}
\expandafter\def\csname v@on.conv.kr.512\endcsname{1.324}
\expandafter\def\csname v@on.conv.kr.8\endcsname{1.015}
\expandafter\def\csname v@on.gap\endcsname{4.4\,\%}
\expandafter\def\csname v@on.gap4\endcsname{1.9\,\%}
\expandafter\def\csname v@on.kr.pair.hi\endcsname{1.26}
\expandafter\def\csname v@on.kr.pair.lo\endcsname{0.82}
\expandafter\def\csname v@on.tri.gap\endcsname{6.1\,\%}
\expandafter\def\csname v@orth.pair\endcsname{\ensuremath{2\times 10^{-17}}}
\expandafter\def\csname v@past.pair.hi\endcsname{0.10}
\expandafter\def\csname v@past.pair.lo\endcsname{0.05}
\expandafter\def\csname v@ph.H4a.all.n\endcsname{293}
\expandafter\def\csname v@ph.H4a.all.share\endcsname{96\,\%}
\expandafter\def\csname v@ph.H5b.e2e.n\endcsname{8}
\expandafter\def\csname v@ph.H5b.e2e.sig\endcsname{0}
\expandafter\def\csname v@ph.H5b.nll.n\endcsname{6}
\expandafter\def\csname v@ph.H5b.nll.sig\endcsname{4}
\expandafter\def\csname v@ph.H5b.val.n\endcsname{4}
\expandafter\def\csname v@ph.H5b.val.sig\endcsname{1}
\expandafter\def\csname v@ph.H5c.e2e.n\endcsname{26}
\expandafter\def\csname v@ph.H5c.e2e.opp\endcsname{7}
\expandafter\def\csname v@ph.H5c.e2e.sig\endcsname{0}
\expandafter\def\csname v@ph.H5c.nll.n\endcsname{26}
\expandafter\def\csname v@ph.H5c.nll.opp\endcsname{4}
\expandafter\def\csname v@ph.H5c.nll.sig\endcsname{21}
\expandafter\def\csname v@ph.H5c.val.n\endcsname{26}
\expandafter\def\csname v@ph.H5c.val.opp\endcsname{8}
\expandafter\def\csname v@ph.H5c.val.sig\endcsname{3}
\expandafter\def\csname v@ph.H5d.basis.pair.n\endcsname{96}
\expandafter\def\csname v@ph.H5d.basis.pair.sig\endcsname{4}
\expandafter\def\csname v@ph.H5d.basis.triple.sig\endcsname{23}
\expandafter\def\csname v@ph.H5d.k.2.n\endcsname{64}
\expandafter\def\csname v@ph.H5d.k.2.sig\endcsname{0}
\expandafter\def\csname v@ph.H5d.k.4.sig\endcsname{6}
\expandafter\def\csname v@ph.H5d.k.8.sig\endcsname{21}
\expandafter\def\csname v@ph.a6\endcsname{0.047}
\expandafter\def\csname v@ph.bound.med\endcsname{0.037}
\expandafter\def\csname v@ph.bound.n\endcsname{149{,}880}
\expandafter\def\csname v@ph.closer.undef\endcsname{55\,\%}
\expandafter\def\csname v@ph.closer.undef.pair\endcsname{59\,\%}
\expandafter\def\csname v@ph.depth.var.S2.pair.1\endcsname{0.0713}
\expandafter\def\csname v@ph.depth.var.S2.pair.4\endcsname{0.0829}
\expandafter\def\csname v@ph.dvn.0.001\endcsname{67\,\%}
\expandafter\def\csname v@ph.dvn.0.01\endcsname{68\,\%}
\expandafter\def\csname v@ph.dvn.0.1\endcsname{72\,\%}
\expandafter\def\csname v@ph.dvn.n\endcsname{873}
\expandafter\def\csname v@ph.e2e.1_adam.S2\endcsname{0.036}
\expandafter\def\csname v@ph.e2e.1_adam.S3c\endcsname{0.021}
\expandafter\def\csname v@ph.e2e.1_adam.held\endcsname{0.006}
\expandafter\def\csname v@ph.e2e.1_lbfgs.S2\endcsname{0.031}
\expandafter\def\csname v@ph.e2e.1_lbfgs.S3c\endcsname{0.021}
\expandafter\def\csname v@ph.e2e.1_lbfgs.held\endcsname{0.025}
\expandafter\def\csname v@ph.eqd.mink\endcsname{8}
\expandafter\def\csname v@ph.eqd.nonpos\endcsname{22{,}297}
\expandafter\def\csname v@ph.eqd.pairmin\endcsname{0.71}
\expandafter\def\csname v@ph.eqd.pop.nonpos\endcsname{2{,}230}
\expandafter\def\csname v@ph.eqd.pop.rows\endcsname{39{,}900}
\expandafter\def\csname v@ph.eqd.rows\endcsname{399{,}000}
\expandafter\def\csname v@ph.flat.nll.P\endcsname{2.1\,\%}
\expandafter\def\csname v@ph.flat.nll.chi\endcsname{0.52}
\expandafter\def\csname v@ph.g.chi\endcsname{0.0015}
\expandafter\def\csname v@ph.h4b.rel\endcsname{2.8\,\%}
\expandafter\def\csname v@ph.h4b.triple\endcsname{93\,\%}
\expandafter\def\csname v@ph.h4b.worst.from\endcsname{26.4}
\expandafter\def\csname v@ph.h4b.worst.to\endcsname{18.1}
\expandafter\def\csname v@ph.h5a.S4.pair.n\endcsname{12}
\expandafter\def\csname v@ph.h5a.S4.pair.ratio\endcsname{1.91}
\expandafter\def\csname v@ph.h5a.S4.pair.sig\endcsname{8}
\expandafter\def\csname v@ph.h5a.S4.triple.ratio\endcsname{0.04}
\expandafter\def\csname v@ph.h5a.sig.floor\endcsname{2.43}
\expandafter\def\csname v@ph.h5a.sig.maxL4\endcsname{0.0012}
\expandafter\def\csname v@ph.h5a.sig.rho0\endcsname{4}
\expandafter\def\csname v@ph.layer.minD\endcsname{0.97}
\expandafter\def\csname v@ph.layer.minDlow\endcsname{0.93}
\expandafter\def\csname v@ph.mv.n\endcsname{181}
\expandafter\def\csname v@ph.mv.s2.mean\endcsname{0.0020}
\expandafter\def\csname v@ph.mv.s2.var\endcsname{13.5\,\%}
\expandafter\def\csname v@ph.mv.share\endcsname{78\,\%}
\expandafter\def\csname v@ph.narrow.k32.rho\endcsname{0.9}
\expandafter\def\csname v@ph.noise.zmax\endcsname{4.4}
\expandafter\def\csname v@ph.noise.zmin\endcsname{1.0}
\expandafter\def\csname v@ph.s2.gap\endcsname{\ensuremath{1.6\times 10^{-4}}}
\expandafter\def\csname v@ph.s2.maxgap\endcsname{0.018}
\expandafter\def\csname v@ph.s4.pair\endcsname{\ensuremath{2.9\times 10^{-4}}}
\expandafter\def\csname v@ph.s4.pairmax\endcsname{\ensuremath{9.2\times 10^{-4}}}
\expandafter\def\csname v@ph.s4.triple\endcsname{0.015}
\expandafter\def\csname v@ph.s4.triplemax\endcsname{0.28}
\expandafter\def\csname v@ph.t10.chi\endcsname{0.09}
\expandafter\def\csname v@ph.t3.chi\endcsname{0.30}
\expandafter\def\csname v@ph.t3.minD4\endcsname{$-$0.67}
\expandafter\def\csname v@ph.tr.e2e4.adam.kept\endcsname{54\,\%}
\expandafter\def\csname v@ph.tr.e2e4.lbfgs.kept\endcsname{50\,\%}
\expandafter\def\csname v@ph.tr.nll.adam.dP\endcsname{0.055}
\expandafter\def\csname v@ph.tr.nll.adam.dminD\endcsname{7.3}
\expandafter\def\csname v@ph.tr.nll.lbfgs.dP\endcsname{0.043}
\expandafter\def\csname v@ph.tr.nll.lbfgs.dminD\endcsname{8.1}
\expandafter\def\csname v@ph.tr.runs\endcsname{2{,}880}
\expandafter\def\csname v@ph.tv1.max\endcsname{473}
\expandafter\def\csname v@ph.tv1.n\endcsname{3{,}245}
\expandafter\def\csname v@ph.tv1.share\endcsname{0.8\,\%}
\expandafter\def\csname v@ph.tv1.triple\endcsname{95\,\%}
\expandafter\def\csname v@pilot.floor\endcsname{0.01}
\expandafter\def\csname v@pilot.inputs\endcsname{8}
\expandafter\def\csname v@pilot.marg\endcsname{0.083}
\expandafter\def\csname v@pilot.missing.win\endcsname{40}
\expandafter\def\csname v@pilot.missing.win.n\endcsname{40}
\expandafter\def\csname v@pilot.noisy.win\endcsname{37}
\expandafter\def\csname v@pilot.noisy.win.n\endcsname{40}
\expandafter\def\csname v@pilot.rawmax\endcsname{\ensuremath{3\times 10^{4}}}
\expandafter\def\csname v@pilot.rho.max\endcsname{0.88}
\expandafter\def\csname v@pilot.rho.med\endcsname{0.12}
\expandafter\def\csname v@pilot.samples\endcsname{128}
\expandafter\def\csname v@pilot.sd\endcsname{0.003}
\expandafter\def\csname v@pilot.seeds\endcsname{10}
\expandafter\def\csname v@pilot.sklearn\endcsname{1.9.1}
\expandafter\def\csname v@pilot.tau\endcsname{0.7}
\expandafter\def\csname v@pilot.test\endcsname{2{,}500}
\expandafter\def\csname v@pilot.train\endcsname{15{,}000}
\expandafter\def\csname v@reg.logn\endcsname{2^{20}}
\expandafter\def\csname v@reg.n\endcsname{64}
\expandafter\def\csname v@reg.share\endcsname{96.9\,\%}
\expandafter\def\csname v@sel.cv\endcsname{0.75}
\expandafter\def\csname v@sel.cv.all\endcsname{0.996}
\expandafter\def\csname v@sel.d5.all\endcsname{9}
\expandafter\def\csname v@sel.d5.alln\endcsname{120}
\expandafter\def\csname v@sel.d5.s2\endcsname{3}
\expandafter\def\csname v@sel.d5.s2n\endcsname{20}
\expandafter\def\csname v@sel.dmax\endcsname{0.13}
\expandafter\def\csname v@sel.dmax.all\endcsname{0.176}
\expandafter\def\csname v@sel.k32.pair\endcsname{5\,\%}
\expandafter\def\csname v@sel.q99\endcsname{0.062}
\expandafter\def\csname v@sem.big.hi\endcsname{0.19}
\expandafter\def\csname v@sem.big.lo\endcsname{0.07}
\expandafter\def\csname v@sem.big.marg.closer.hi\endcsname{88\,\%}
\expandafter\def\csname v@sem.big.marg.closer.lo\endcsname{48\,\%}
\expandafter\def\csname v@sem.evidence.cd.0.4.1.0\endcsname{0.008}
\expandafter\def\csname v@sem.evidence.eu.0.4.1.0\endcsname{0.028}
\expandafter\def\csname v@sem.flip.mind.hi\endcsname{0.45}
\expandafter\def\csname v@sem.flip.mind.lo\endcsname{0.03}
\expandafter\def\csname v@sem.flip.triple.8.1.0.evidence.cd\endcsname{0.033}
\expandafter\def\csname v@sem.flip.triple.8.1.0.evidence.eu\endcsname{0.044}
\expandafter\def\csname v@sem.flip.weak.hi\endcsname{0.0035}
\expandafter\def\csname v@sem.marginal.cd.0.4.1.0\endcsname{0.024}
\expandafter\def\csname v@sem.marginal.eu.0.4.1.0\endcsname{0.010}
\expandafter\def\csname v@sem.miss.cd.0.4\endcsname{0.019}
\expandafter\def\csname v@sem.miss.eu.0.4\endcsname{0.081}
\expandafter\def\csname v@sem.posterior.cd.0.4.1.0\endcsname{0.034}
\expandafter\def\csname v@sem.posterior.eu.0.4.1.0\endcsname{0.008}
\expandafter\def\csname v@sw.driver\endcsname{595.91.07}
\expandafter\def\csname v@sw.numpy\endcsname{1.26.4}
\expandafter\def\csname v@sw.scipy\endcsname{1.13.1}
\expandafter\def\csname v@sw.torch\endcsname{2.4.0+cu121}
\expandafter\def\csname v@thr.SUM.pair.16.d1\endcsname{0.043}
\expandafter\def\csname v@thr.SUM.pair.2.d1\endcsname{0.192}
\expandafter\def\csname v@thr.SUM.pair.32.on\endcsname{0.007}
\expandafter\def\csname v@thr.SUM.pair.32.pop\endcsname{0.039}
\expandafter\def\csname v@thr.SUM.pair.32.und\endcsname{0.010}
\expandafter\def\csname v@thr.SUM.triple.16.on\endcsname{0.000}
\expandafter\def\csname v@thr.SUM.triple.16.pop\endcsname{0.023}
\expandafter\def\csname v@tr.S2.triple.euc\endcsname{100\,\%}
\expandafter\def\csname v@tr.S2.triple.ratio\endcsname{0.23}
\expandafter\def\csname v@tr.S3a.pair.cd\endcsname{0.0472}
\expandafter\def\csname v@tr.S3a.pair.eu\endcsname{0.0471}
\expandafter\def\csname v@tr.S3b.pair.euc\endcsname{69\,\%}
\expandafter\def\csname v@tr.S3b.triple.ratio\endcsname{5.5}
\expandafter\def\csname v@tr.S3c.pair.euc\endcsname{24\,\%}
\expandafter\def\csname v@tr.S3d.pair.ratio\endcsname{420}
\expandafter\def\csname v@tr.S3d.triple.ratio\endcsname{32}
\expandafter\def\csname v@tr.S4.pair.ratio\endcsname{2.0}
\expandafter\def\csname v@tr.S4.triple.ratio\endcsname{0.027}
\expandafter\def\csname v@tvb.gain\endcsname{4.3}
\expandafter\def\csname v@tvb.n\endcsname{121}
\expandafter\def\csname v@tvb.sharp\endcsname{0.20}
\expandafter\def\csname v@tvb.sharpmax\endcsname{0.85}
\expandafter\def\csname v@val.M4.n\endcsname{2{,}088}
\expandafter\def\csname v@val.M4.worst\endcsname{0.33}
\expandafter\def\csname v@val.M6\endcsname{2.22\times 10^{-16}}
\expandafter\def\csname v@val.M6.n\endcsname{36}
\expandafter\def\csname v@val.M7.low\endcsname{6.2\times 10^{-15}}
\expandafter\def\csname v@val.M7.q\endcsname{1.6\times 10^{-14}}
\expandafter\def\csname v@val.M7.up\endcsname{2.8\times 10^{-16}}
\expandafter\def\csname v@val.M9\endcsname{8.2\times 10^{-5}}

%% file: generated/tab_onset.tex
\footnotesize
\begin{tabular}{rrrrrr}
\toprule
& \multicolumn{3}{c}{pair basis} & \multicolumn{2}{c}{triple basis} \\ \cmidrule(lr){2-4}\cmidrule(l){5-6}
$k$ & first & convex & measured & corner & measured \\
\midrule
2 & 0.152 & 0.289 & 0.437 & --- & --- \\
4 & 0.085 & 0.201 & 0.204 & 0.093 & 0.094 \\
8 & 0.046 & 0.127 & 0.129 & 0.044 & 0.045 \\
16 & 0.024 & 0.073 & 0.074 & 0.022 & 0.023 \\
32 & 0.012 & 0.039 & 0.039 & 0.012 & 0.011 \\
\bottomrule
\end{tabular}

%% file: generated/tab_hyp.tex
\footnotesize
\begin{tabular}{@{}l>{\raggedright\arraybackslash}p{0.31\linewidth}l>{\raggedright\arraybackslash}p{0.40\linewidth}@{}}
\toprule
 & claim & verdict & evidence \\
\midrule
H1a & population onset falls with $k$ & \emph{supported} & 4/4 cells strictly decreasing \\
H1b & same at $n=10^7$ & \emph{supported} & 4/4; SUM/pair 0.435 ($k=2$) vs 0.007 ($k=32$) \\
H1c & $d_{\mathrm{denom}}$ one power law of $\chi_D$ & \emph{inconclusive} & within factor 2: SUM/pair 0.55, CONTRAST/pair 0.52, SUM/triple 1.00, CONTRAST/triple 0.48 \\
H2a & higher $m$, earlier onset & \emph{supported} & 20/20 \\
H2b & triple earlier than pair (population) & \emph{supported} & 8/8 \\
H2b & triple earlier than pair ($n=10^7$) & \emph{supported} & 8/8 \\
H2c & $d_{\mathrm{denom}}$ depends weakly on $m$ & \emph{inconclusive} & 60\,\% of 221 points; supported needs $\geq 80\,\%$ \\
H3a & onset within $\pm 0.05$ of Gaussian & \emph{not supported} & 21/48 CIs inside, 27 entirely outside \\
H3b & $d_{\mathrm{denom}}$ within factor 2 of Gaussian & \emph{not supported} & 73/236 CIs inside, 156 entirely outside \\
H3c & Gaussian power law predicts LKJ & \emph{not supported} & pair 39\,\%, triple 34\,\% of 272 matrices; $k=2$ only \\
H4a & seed spread exceeds MC error & \emph{supported} & 93\,\% of 114 points; median ratio 2.1 \\
H4b & doubling changes TV $<\max(3\,\mathrm{SE},10^{-3})$ & \emph{not supported} & 5{,}460 violations, all under $n\times 2$ \\
H5a & $d_{\mathrm{total}}$ grows from $L=1$ to 4 & \emph{not supported} & 11/72; 51 significantly below 1 \\
H5b & training reduces $d_{\mathrm{denom}}$ & \emph{not supported} & 5/18; 1 opposite \\
H5c & training reduces $P(D<0)$ & \emph{not supported} & 24/78; 19 opposite \\
H5d & end-to-end training reduces held-out $d_{\mathrm{total}}$ & \emph{not supported} & 27/192; 84 opposite \\
\bottomrule
\end{tabular}

%% file: generated/tab_trunc.tex
\footnotesize
\begin{tabular}{llrrrrrrr}
\toprule
basis & belief & evaluations & $d_{\mathrm{denom}}$ & $\mathrm{TV}(p_{\mathrm{EU}},p_{\mathrm{G}})$ & $\mathrm{TV}(p_{\mathrm{CD}},p_{\mathrm{G}})$ & \multicolumn{1}{c}{ratio} & EU closer & EC beats EU \\
\midrule
pair & S2 & 2{,}902 & 0.0024 & 0.0119 & 0.0144 & 5.3 & 50\,\% & 16\,\% \\
 & S3a & 5{,}037 & 0.0004 & 0.0471 & 0.0472 & 120 & 52\,\% & 24\,\% \\
 & S3b & 6{,}200 & 0.0041 & 0.1063 & 0.1096 & 28 & 69\,\% & 38\,\% \\
 & S3c & 3{,}249 & 0.0019 & 0.0690 & 0.0675 & 30 & 24\,\% & 8\,\% \\
 & S3d & 7{,}022 & 0.0002 & 0.0824 & 0.0830 & 420 & 67\,\% & 39\,\% \\
 & S4 & 2{,}902 & 0.0002 & 0.0004 & 0.0003 & 2.0 & 41\,\% & 0\,\% \\
\midrule
triple & S2 & 1{,}804 & 0.0117 & 0.0027 & 0.0109 & 0.23 & 100\,\% & 28\,\% \\
 & S3a & 3{,}155 & 0.0015 & 0.0103 & 0.0118 & 7.6 & 55\,\% & 2\,\% \\
 & S3b & 3{,}337 & 0.0085 & 0.0280 & 0.0313 & 5.5 & 66\,\% & 32\,\% \\
 & S3c & 2{,}051 & 0.0021 & 0.0139 & 0.0207 & 7.6 & 29\,\% & 5\,\% \\
 & S3d & 4{,}399 & 0.0006 & 0.0234 & 0.0233 & 32 & 62\,\% & 7\,\% \\
 & S4 & 1{,}804 & 0.0127 & 0.0003 & 0.0126 & 0.027 & 100\,\% & 5\,\% \\
\bottomrule
\end{tabular}

%% file: generated/tab_reg.tex
\footnotesize
\begin{tabular}{rrrrr}
\toprule
$k$ & $p_{\mathrm{CD}}$ & $p_{\mathrm{EU}}$ & additive features & pair features \\
\midrule
2 & 0.019 & 0.023 & 0.017 & 0.009 \\
4 & 0.046 & 0.056 & 0.041 & 0.008 \\
8 & 0.076 & 0.073 & 0.074 & 0.013 \\
16 & 0.089 & 0.086 & 0.091 & 0.023 \\
\bottomrule
\end{tabular}

%% file: generated/tab_sem.tex
\footnotesize
\begin{tabular}{rrrrrrrr}
\toprule
& & \multicolumn{2}{c}{evidence} & \multicolumn{2}{c}{posterior} & \multicolumn{2}{c}{marginal} \\ \cmidrule(lr){3-4}\cmidrule(lr){5-6}\cmidrule(l){7-8}
$\rho_S$ & $\sigma_{\mathrm{obs}}$ & CD & EU & CD & EU & CD & EU \\
\midrule
0.1 & 0.5 & 0.013 & 0.013 & 0.014 & 0.013 & 0.014 & 0.013 \\
 & 1.0 & 0.008 & 0.010 & 0.011 & 0.008 & 0.010 & 0.009 \\
0.2 & 0.5 & 0.011 & 0.013 & 0.014 & 0.011 & 0.014 & 0.012 \\
 & 1.0 & 0.008 & 0.015 & 0.017 & 0.008 & 0.015 & 0.009 \\
0.3 & 0.5 & 0.011 & 0.016 & 0.015 & 0.011 & 0.014 & 0.011 \\
 & 1.0 & 0.008 & 0.024 & 0.030 & 0.008 & 0.023 & 0.009 \\
0.4 & 0.5 & 0.010 & 0.016 & 0.015 & 0.010 & 0.014 & 0.010 \\
 & 1.0 & 0.008 & 0.028 & 0.034 & 0.008 & 0.024 & 0.010 \\
\bottomrule
\end{tabular}

%% file: generated/tab_sem_big.tex
\footnotesize\setlength{\tabcolsep}{4.5pt}
\begin{tabular}{rrrrrrrrrrrr}
\toprule
& & & \multicolumn{3}{c}{evidence} & \multicolumn{3}{c}{posterior} & \multicolumn{3}{c}{marginal} \\ \cmidrule(lr){4-6}\cmidrule(lr){7-9}\cmidrule(l){10-12}
$k$ & $\rho_S$ & $\sigma_{\mathrm{obs}}$ & CD & EU & closer & CD & EU & closer & CD & EU & closer \\
\midrule
4 & 0.10 & 0.5 & 0.131 & 0.133 & 93\,\% & 0.131 & 0.132 & 88\,\% & 0.133 & 0.133 & 88\,\% \\
 &  & 1.0 & 0.078 & 0.092 & 93\,\% & 0.075 & 0.081 & 73\,\% & 0.080 & 0.088 & 75\,\% \\
 & 0.15 & 0.5 & 0.132 & 0.136 & 90\,\% & 0.129 & 0.132 & 85\,\% & 0.133 & 0.132 & 83\,\% \\
 &  & 1.0 & 0.072 & 0.078 & 92\,\% & 0.074 & 0.073 & 53\,\% & 0.088 & 0.089 & 63\,\% \\
8 & 0.05 & 0.5 & 0.139 & 0.141 & 73\,\% & 0.139 & 0.139 & 63\,\% & 0.142 & 0.143 & 68\,\% \\
 &  & 1.0 & 0.073 & 0.081 & 88\,\% & 0.073 & 0.074 & 67\,\% & 0.082 & 0.083 & 68\,\% \\
 & 0.10 & 0.5 & 0.186 & 0.187 & 63\,\% & 0.181 & 0.178 & 60\,\% & 0.173 & 0.176 & 63\,\% \\
 &  & 1.0 & 0.099 & 0.113 & 67\,\% & 0.092 & 0.101 & 55\,\% & 0.119 & 0.124 & 48\,\% \\
\bottomrule
\end{tabular}

%% file: generated/tab_sem_flip.tex
\footnotesize\setlength{\tabcolsep}{4pt}
\begin{tabular}{lrrrrrrrrrrrr}
\toprule
& & & & \multicolumn{3}{c}{evidence} & \multicolumn{3}{c}{posterior} & \multicolumn{3}{c}{marginal} \\ \cmidrule(lr){5-7}\cmidrule(lr){8-10}\cmidrule(l){11-13}
model & $k$ & $\rho_S$ & $\sigma_{\mathrm{obs}}$ & CD & EU & closer & CD & EU & closer & CD & EU & closer \\
\midrule
full & 3 & 0.15 & 0.5 & 0.024 & 0.030 & 85\,\% & 0.029 & 0.025 & 25\,\% & 0.028 & 0.025 & 28\,\% \\
 &  &  & 1.0 & 0.010 & 0.027 & 92\,\% & 0.028 & 0.011 & 3\,\% & 0.023 & 0.011 & 3\,\% \\
 & 4 & 0.15 & 0.5 & 0.026 & 0.034 & 87\,\% & 0.031 & 0.026 & 18\,\% & 0.028 & 0.023 & 17\,\% \\
 &  &  & 1.0 & 0.013 & 0.042 & 97\,\% & 0.053 & 0.013 & 0\,\% & 0.038 & 0.016 & 2\,\% \\
triple & 3 & 0.15 & 0.5 & 0.033 & 0.043 & 92\,\% & 0.037 & 0.033 & 45\,\% & 0.036 & 0.035 & 52\,\% \\
 &  &  & 1.0 & 0.015 & 0.039 & 97\,\% & 0.049 & 0.015 & 12\,\% & 0.040 & 0.020 & 25\,\% \\
 & 4 & 0.09 & 0.5 & 0.059 & 0.065 & 87\,\% & 0.060 & 0.060 & 52\,\% & 0.061 & 0.064 & 58\,\% \\
 &  &  & 1.0 & 0.023 & 0.041 & 88\,\% & 0.047 & 0.024 & 5\,\% & 0.039 & 0.026 & 10\,\% \\
 & 8 & 0.04 & 0.5 & 0.088 & 0.090 & 78\,\% & 0.086 & 0.087 & 57\,\% & 0.084 & 0.088 & 53\,\% \\
 &  &  & 1.0 & 0.033 & 0.044 & 88\,\% & 0.046 & 0.033 & 8\,\% & 0.045 & 0.037 & 15\,\% \\
\bottomrule
\end{tabular}

%% file: generated/tab_pilot.tex
\scriptsize\setlength{\tabcolsep}{3.5pt}
\begin{tabular}{lllllllll}
\toprule
& \multicolumn{5}{c}{MSE of the clipped mean} & \multicolumn{3}{c}{mean clipped log density} \\ \cmidrule(lr){2-6}\cmidrule(l){7-9}
scenario & CD & EU & REG & boosting & marginal & CD & EU & REG \\
\midrule
all inputs seen & 0.046--0.059 & 0.042--0.056 & 0.019--0.021 & 0.012 & 0.083 & 0.18--0.32 & 0.21--0.36 & 0.55--0.68 \\
half missing & 0.057--0.064 & 0.071--0.075 & 0.057--0.058 & 0.038 & 0.083 & 0.14--0.21 & 0.02--0.08 & 0.07--0.12 \\
noisy & 0.058--0.064 & 0.065--0.070 & 0.050--0.057 & 0.046 & 0.083 & 0.13--0.17 & 0.04--0.13 & 0.01--0.17 \\
\bottomrule
\end{tabular}

%% file: generated/tab_pilot_fail.tex
\footnotesize
\begin{tabular}{lllrr}
\toprule
scenario & below the floor & rows with $\mathrm{E}_q[D]\leq0$ & CD ahead of EU & largest raw MSE of CD \\
\midrule
all inputs seen & 0.6--1.2\,\% & 0.6--1.2\,\% & 0/40 & \ensuremath{4\times 10^{3}} \\
half missing & 17.3--24.3\,\% & 0.5--0.9\,\% & 40/40 & \ensuremath{6\times 10^{3}} \\
noisy & 12.9--22.9\,\% & 1.1--1.5\,\% & 37/40 & \ensuremath{3\times 10^{4}} \\
\bottomrule
\end{tabular}

%% file: generated/tab_onset_all.tex
\footnotesize
\begin{tabular}{rrrrrrrrr}
\toprule
& & \multicolumn{5}{c}{pair} & \multicolumn{2}{c}{triple} \\ \cmidrule(lr){3-7}\cmidrule(l){8-9}
$m$ & $k$ & first & convex & search & measured & $k\rho_S^\ast$ & corner & measured \\
\midrule
2 & 2 & 0.247 & 0.406 & 0.545 & 0.545 & 1.09 & --- & --- \\
 & 4 & 0.140 & 0.261 & 0.261 & 0.261 & 1.04 & --- & 0.199 \\
 & 8 & 0.075 & 0.158 & 0.158 & 0.158 & 1.26 & --- & 0.087 \\
 & 16 & 0.039 & 0.090 & 0.090 & 0.090 & 1.44 & --- & 0.041 \\
 & 32 & 0.020 & 0.049 & 0.049 & 0.049 & 1.57 & --- & 0.021 \\
\midrule
4 & 2 & 0.152 & 0.289 & 0.437 & 0.437 & 0.87 & --- & --- \\
 & 4 & 0.085 & 0.201 & 0.204 & 0.204 & 0.82 & 0.093 & 0.094 \\
 & 8 & 0.046 & 0.127 & 0.129 & 0.129 & 1.03 & 0.044 & 0.045 \\
 & 16 & 0.024 & 0.073 & 0.074 & 0.074 & 1.18 & 0.022 & 0.023 \\
 & 32 & 0.012 & 0.039 & 0.039 & 0.039 & 1.26 & 0.012 & 0.011 \\
\midrule
6 & 2 & 0.120 & 0.229 & 0.398 & 0.398 & 0.80 & --- & --- \\
 & 4 & 0.068 & 0.167 & 0.168 & 0.168 & 0.67 & 0.070 & 0.070 \\
 & 8 & 0.036 & 0.111 & 0.116 & 0.116 & 0.92 & 0.033 & 0.035 \\
 & 16 & 0.019 & 0.068 & 0.068 & 0.068 & 1.09 & 0.017 & 0.018 \\
 & 32 & 0.010 & 0.038 & 0.038 & 0.038 & 1.21 & 0.009 & 0.009 \\
\bottomrule
\end{tabular}

%% file: generated/tab_noise_z.tex
\footnotesize
\begin{tabular}{lrrrrrr}
\toprule
basis & $k$ & $z$ measured & $z$ corners & $z$ cube & on the boundary & $\sqrt{2k\ln 2}$ \\
\midrule
pair & 2 & 1.0 & 1.2 & 1.2 & 100\,\% & 1.7 \\
 & 4 & 1.8 & 1.5 & 1.7 & 75\,\% & 2.4 \\
 & 8 & 2.5 & 2.0 & 2.4 & 75\,\% & 3.3 \\
 & 16 & 3.5 & 2.9 & 3.8 & 75\,\% & 4.7 \\
 & 32 & 4.4 & 4.4 & 5.5 & 77\,\% & 6.7 \\
triple & 4 & 1.8 & 1.5 & 1.8 & 100\,\% & 2.4 \\
 & 8 & 2.6 & 2.4 & 2.7 & 88\,\% & 3.3 \\
 & 16 & 3.3 & 3.7 & 3.7 & 91\,\% & 4.7 \\
\bottomrule
\end{tabular}

%% file: generated/tab_noise.tex
\footnotesize
\begin{tabular}{lrrrrrrr}
\toprule
basis & $k$ & $|B_D|$ & $\sigma_{\mathrm{c}}$ & $1-\min D$ & ratio & $\chi_D$ & $\sqrt{|B_D|/n}$ \\
\midrule
pair & 2 & 24 & 0.008 & 0.008 & 1.0 & \ensuremath{1.52\times 10^{-3}} & \ensuremath{1.55\times 10^{-3}} \\
 & 4 & 112 & 0.019 & 0.035 & 1.8 & \ensuremath{3.32\times 10^{-3}} & \ensuremath{3.35\times 10^{-3}} \\
 & 8 & 480 & 0.040 & 0.099 & 2.5 & \ensuremath{6.87\times 10^{-3}} & \ensuremath{6.93\times 10^{-3}} \\
 & 16 & 1{,}984 & 0.083 & 0.292 & 3.5 & \ensuremath{1.41\times 10^{-2}} & \ensuremath{1.41\times 10^{-2}} \\
 & 32 & 8{,}064 & 0.169 & 0.746 & 4.4 & \ensuremath{2.84\times 10^{-2}} & \ensuremath{2.84\times 10^{-2}} \\
triple & 4 & 368 & 0.077 & 0.134 & 1.8 & \ensuremath{5.99\times 10^{-3}} & \ensuremath{6.07\times 10^{-3}} \\
 & 8 & 4{,}064 & 0.281 & 0.734 & 2.6 & \ensuremath{2.01\times 10^{-2}} & \ensuremath{2.02\times 10^{-2}} \\
 & 16 & 37{,}824 & 0.884 & 2.948 & 3.3 & \ensuremath{6.15\times 10^{-2}} & \ensuremath{6.15\times 10^{-2}} \\
 & 32 & 325{,}504 & 2.624 & 11.066 & 4.2 & \ensuremath{1.80\times 10^{-1}} & \ensuremath{1.80\times 10^{-1}} \\
\bottomrule
\end{tabular}

%% file: generated/tab_thr_main.tex
\footnotesize
\begin{tabular}{lrrrrrr}
\toprule
basis & $k$ & $d_{\mathrm{denom}}\geq 1\,\%$ & $d_{\mathrm{denom}}\geq 5\,\%$ & onset, $n=10^7$ & onset, pop. & undefined from \\
\midrule
pair & 2 & 0.192 & 0.359 & 0.435 & 0.437 & 0.450 \\
 & 4 & 0.102 & --- & 0.201 & 0.204 & 0.200 \\
 & 8 & 0.067 & --- & 0.123 & 0.129 & 0.150 \\
 & 16 & 0.043 & --- & 0.059 & 0.074 & 0.075 \\
 & 32 & --- & --- & 0.007 & 0.039 & 0.010 \\
triple & 2 & 0.085 & --- & 0.435 & 0.437 & 0.450 \\
 & 4 & 0.020 & --- & 0.085 & 0.094 & 0.100 \\
 & 8 & 0.009 & --- & 0.016 & 0.045 & 0.020 \\
 & 16 & --- & --- & 0.000 & 0.023 & 0.000 \\
 & 32 & --- & --- & 0.000 & 0.011 & 0.000 \\
\bottomrule
\end{tabular}

%% file: generated/tab_thr_full.tex
\scriptsize\setlength{\tabcolsep}{4pt}
\begin{tabular}{llrlllrr}
\toprule
type & basis & $k$ & $d_{\mathrm{denom}}\geq 1\,\%$ & $d_{\mathrm{denom}}\geq 5\,\%$ & onset, $n=10^7$ & onset, pop. & undefined from \\
\midrule
SUM & pair & 2 & 0.192 (0.192--0.193) & 0.359 (0.359--0.360) & 0.435 (0.435--0.436) & 0.437 & 0.450 \\
 &  & 4 & 0.102 (0.101--0.102) & --- & 0.201 (0.201--0.201) & 0.204 & 0.200 \\
 &  & 8 & 0.067 (0.067--0.067) & --- & 0.123 (0.123--0.123) & 0.129 & 0.150 \\
 &  & 16 & 0.043 (0.043--0.043) & --- & 0.059 (0.059--0.060) & 0.074 & 0.075 \\
 &  & 32 & --- & --- & 0.007 (0.007--0.008) & 0.039 & 0.010 \\
 & triple & 2 & 0.085 (0.085--0.086) & --- & 0.435 (0.435--0.436) & 0.437 & 0.450 \\
 &  & 4 & 0.020 (0.020--0.020) & --- & 0.085 (0.083--0.087) & 0.094 & 0.100 \\
 &  & 8 & 0.009 (0.009--0.009) & --- & 0.016 (0.015--0.017) & 0.045 & 0.020 \\
 &  & 16 & --- & --- & 0.000 (0.000--0.000) & 0.023 & 0.000 \\
 &  & 32 & --- & --- & 0.000 (0.000--0.000) & 0.011 & 0.000 \\
\midrule
CONTRAST & pair & 2 & 0.192 (0.192--0.192) & 0.388 (0.387--0.388) & 0.435 (0.435--0.435) & 0.437 & 0.450 \\
 &  & 4 & 0.122 (0.121--0.122) & --- & 0.201 (0.201--0.201) & 0.204 & 0.200 \\
 &  & 8 & 0.084 (0.084--0.084) & --- & 0.123 (0.123--0.123) & 0.129 & 0.150 \\
 &  & 16 & --- & --- & 0.059 (0.059--0.060) & 0.074 & 0.075 \\
 &  & 32 & --- & --- & 0.007 (0.007--0.008) & 0.039 & 0.010 \\
 & triple & 2 & 0.065 (0.065--0.065) & 0.235 (0.235--0.235) & 0.435 (0.435--0.435) & 0.437 & 0.450 \\
 &  & 4 & 0.054 (0.054--0.054) & --- & 0.084 (0.082--0.086) & 0.094 & 0.100 \\
 &  & 8 & --- & --- & 0.015 (0.013--0.018) & 0.045 & 0.020 \\
 &  & 16 & --- & --- & 0.000 (0.000--0.000) & 0.023 & 0.000 \\
 &  & 32 & --- & --- & 0.000 (0.000--0.000) & 0.011 & 0.000 \\
\bottomrule
\end{tabular}

%% file: generated/tab_thr_relvar.tex
\footnotesize
\begin{tabular}{llrllr}
\toprule
type & basis & $k$ & $|\mathrm{relvar}|\geq 5\,\%$ & $|\mathrm{relvar}|\geq 20\,\%$ & $\min D$ at $\rho_S=0$ \\
\midrule
SUM & pair & 2 & 0.283 (0.282--0.283) & --- & 0.992 \\
 &  & 4 & 0.147 (0.146--0.147) & --- & 0.965 \\
 &  & 8 & 0.095 (0.095--0.095) & --- & 0.901 \\
 &  & 16 & --- & --- & 0.708 \\
 &  & 32 & --- & --- & 0.254 \\
 & triple & 2 & 0.246 (0.245--0.246) & --- & 0.992 \\
 &  & 4 & 0.048 (0.048--0.048) & --- & 0.866 \\
 &  & 8 & --- & --- & 0.266 \\
 &  & 16 & --- & --- & $-$1.948 \\
 &  & 32 & --- & --- & $-$10.066 \\
CONTRAST & pair & 2 & 0.299 (0.299--0.299) & --- & 0.992 \\
 &  & 4 & --- & --- & 0.966 \\
 &  & 8 & --- & --- & 0.895 \\
 &  & 16 & --- & --- & 0.711 \\
 &  & 32 & --- & --- & 0.253 \\
 & triple & 2 & 0.139 (0.139--0.139) & --- & 0.992 \\
 &  & 4 & --- & --- & 0.863 \\
 &  & 8 & --- & --- & 0.250 \\
 &  & 16 & --- & --- & $-$2.068 \\
 &  & 32 & --- & --- & $-$10.202 \\
\bottomrule
\end{tabular}

%% file: generated/tab_beliefs.tex
\footnotesize
\begin{tabular}{lrllllll}
\toprule
type & $k$ & S2 & S3a & S3b & S3c & S3d & S4 \\
\midrule
SUM & 2 & 0.192 / 0.45 & --- / 0.45 & --- / 0.45 & 0.217 / 0.45 & --- / 0.60 & --- / 0.45 \\
 & 4 & 0.102 / 0.20 & --- / 0.25 & 0.039 / 0.25 & 0.125 / 0.20 & --- / 0.35 & --- / 0.20 \\
 & 8 & 0.067 / 0.15 & --- / 0.15 & 0.014 / 0.20 & --- / 0.15 & --- / 0.35 & --- / 0.15 \\
 & 16 & 0.043 / 0.075 & --- / 0.15 & 0.009 / 0.90 & --- / 0.075 & --- / --- & --- / 0.075 \\
 & 32 & --- / 0.010 & 0.054 / --- & 0.027 / --- & --- / 0.050 & --- / --- & --- / 0.010 \\
\midrule
CONTRAST & 2 & 0.192 / 0.45 & --- / 0.45 & --- / 0.45 & 0.102 / 0.45 & --- / 0.60 & --- / 0.45 \\
 & 4 & 0.122 / 0.20 & --- / 0.25 & --- / 0.25 & 0.177 / 0.25 & --- / 0.35 & --- / 0.20 \\
 & 8 & 0.084 / 0.15 & --- / 0.15 & --- / 0.20 & --- / 0.15 & --- / 0.35 & --- / 0.15 \\
 & 16 & --- / 0.075 & --- / 0.15 & --- / 0.90 & --- / 0.075 & --- / --- & --- / 0.075 \\
 & 32 & --- / 0.010 & --- / --- & --- / --- & --- / 0.050 & --- / --- & --- / 0.010 \\
\bottomrule
\end{tabular}

%% file: generated/tab_past.tex
\footnotesize
\begin{tabular}{llrrrrrr}
\toprule
basis & belief & evaluations & share undefined & $d_{\mathrm{total}}$ & $d_{\mathrm{dc}}$ & $d_{\mathrm{model,cal}}$ & CD closer than EC \\
\midrule
pair & S2 & 6{,}098 & 68\,\% & 0.098 & 0.086 & 0.146 & 64\,\% \\
 & S3a & 3{,}963 & 44\,\% & 0.054 & 0.008 & 0.050 & 30\,\% \\
 & S3b & 2{,}800 & 31\,\% & 0.067 & 0.011 & 0.057 & 25\,\% \\
 & S3c & 5{,}751 & 64\,\% & 0.067 & 0.052 & 0.086 & 64\,\% \\
 & S3d & 1{,}978 & 22\,\% & 0.046 & 0.001 & 0.041 & 25\,\% \\
 & S4 & 6{,}098 & 68\,\% & 0.000 & 0.017 & 0.017 & 97\,\% \\
\midrule
triple & S2 & 7{,}196 & 80\,\% & 0.079 & 0.051 & 0.068 & 41\,\% \\
 & S3a & 5{,}845 & 65\,\% & 0.078 & 0.042 & 0.073 & 50\,\% \\
 & S3b & 5{,}663 & 63\,\% & 0.112 & 0.053 & 0.086 & 38\,\% \\
 & S3c & 6{,}949 & 77\,\% & 0.033 & 0.053 & 0.061 & 71\,\% \\
 & S3d & 4{,}601 & 51\,\% & 0.102 & 0.036 & 0.105 & 54\,\% \\
 & S4 & 7{,}196 & 80\,\% & 0.022 & 0.039 & 0.021 & 54\,\% \\
\bottomrule
\end{tabular}

%% file: generated/tab_decomp.tex
\footnotesize
\begin{tabular}{llrrrrrr}
\toprule
belief & type & $\rho_S$ & $d_{\mathrm{denom}}$ & $d_{\mathrm{neg}}$ & $d_{\mathrm{model,uncal}}$ & $d_{\mathrm{model,cal}}$ & $d_{\mathrm{total}}$ \\
\midrule
S2 & SUM & 0.1 & 0.0025 & 0.0111 & 0.0131 & 0.0241 & 0.0106 \\
 &  & 0.2 & 0.0108 & 0.0076 & 0.0287 & 0.0358 & 0.0180 \\
 &  & 0.4 & 0.0736 & 0.0376 & 0.1002 & 0.0627 & 0.0265 \\
 & CONTRAST & 0.1 & 0.0025 & 0.0123 & 0.0129 & 0.0026 & 0.0154 \\
 &  & 0.2 & 0.0108 & 0.0095 & 0.0267 & 0.0195 & 0.0374 \\
 &  & 0.4 & 0.0547 & 0.0238 & 0.0397 & 0.0617 & 0.0906 \\
\midrule
S3c & SUM & 0.1 & 0.0034 & 0.0064 & 0.0399 & 0.0445 & 0.0368 \\
 &  & 0.2 & 0.0087 & 0.0064 & 0.0372 & 0.0415 & 0.0294 \\
 &  & 0.4 & 0.0355 & 0.0061 & 0.0485 & 0.0438 & 0.0171 \\
 & CONTRAST & 0.1 & 0.0097 & 0.0391 & 0.0650 & 0.0673 & 0.0710 \\
 &  & 0.2 & 0.0262 & 0.0551 & 0.0854 & 0.0971 & 0.1022 \\
 &  & 0.4 & 0.1030 & 0.1116 & 0.0985 & 0.1371 & 0.1526 \\
\midrule
S4 & SUM & 0.1 & 0.0001 & 0.0120 & 0.0003 & 0.0119 & 0.0003 \\
 &  & 0.2 & 0.0004 & 0.0103 & 0.0004 & 0.0100 & 0.0003 \\
 &  & 0.4 & 0.0003 & 0.0037 & 0.0004 & 0.0037 & 0.0002 \\
 & CONTRAST & 0.1 & 0.0001 & 0.0118 & 0.0003 & 0.0118 & 0.0003 \\
 &  & 0.2 & 0.0004 & 0.0104 & 0.0004 & 0.0101 & 0.0003 \\
 &  & 0.4 & 0.0003 & 0.0044 & 0.0004 & 0.0045 & 0.0004 \\
\bottomrule
\end{tabular}

%% file: generated/tab_h1c.tex
\footnotesize
\begin{tabular}{llrrrrl}
\toprule
type & basis & points & $\beta$ & $C$ & within $2\times$ & cell verdict \\
\midrule
CONTRAST & pair & 33 & 1.59 & 0.040 & 0.52 & inconclusive \\
 & triple & 23 & 1.00 & 0.083 & 0.48 & not supported \\
SUM & pair & 33 & 1.76 & 0.070 & 0.55 & inconclusive \\
 & triple & 23 & 0.88 & 0.107 & 1.00 & supported \\
\bottomrule
\end{tabular}

%% file: generated/tab_lead.tex
\footnotesize
\begin{tabular}{lrrrrrr}
\toprule
type, basis & points & $\beta$ (H1c) & power law & closed form & ratio range & median ratio \\
\midrule
SUM, pair & 33 & 1.76 & 55\,\% & 100\,\% & 0.61--1.65 & 0.88 \\
CONTRAST, pair & 33 & 1.59 & 52\,\% & 100\,\% & 0.61--1.65 & 0.87 \\
SUM, triple & 23 & 0.88 & 100\,\% & 100\,\% & 0.57--0.99 & 0.79 \\
CONTRAST, triple & 23 & 1.00 & 48\,\% & 100\,\% & 1.00--1.76 & 1.07 \\
\bottomrule
\end{tabular}

%% file: generated/tab_onset_m.tex
\footnotesize
\begin{tabular}{rrrrrrr}
\toprule
& \multicolumn{3}{c}{pair} & \multicolumn{3}{c}{triple} \\ \cmidrule(lr){2-4}\cmidrule(l){5-7}
$k$ & $m=2$ & $m=4$ & $m=6$ & $m=2$ & $m=4$ & $m=6$ \\
\midrule
2 & 0.545 & 0.437 & 0.398 & 0.545 & 0.437 & 0.398 \\
4 & 0.261 & 0.204 & 0.168 & 0.199 & 0.094 & 0.070 \\
8 & 0.158 & 0.129 & 0.116 & 0.087 & 0.045 & 0.035 \\
16 & 0.090 & 0.074 & 0.068 & 0.041 & 0.023 & 0.018 \\
32 & 0.049 & 0.039 & 0.038 & 0.021 & 0.011 & 0.009 \\
\bottomrule
\end{tabular}

%% file: generated/tab_families.tex
\footnotesize
\begin{tabular}{lrrrrrr}
\toprule
family & cells & min diff. & max diff. & CI inside & CI outside & median ratio \\
\midrule
Clayton & 6 & $-$0.179 & +0.006 & 2 & 4 & 5.34 \\
Gumbel & 6 & $-$0.128 & $-$0.013 & 1 & 5 & 16.60 \\
$t_{10}$ & 18 & $-$0.102 & +0.036 & 15 & 3 & 2.08 \\
$t_3$ & 18 & $-$0.201 & $-$0.016 & 3 & 15 & 7.22 \\
\bottomrule
\end{tabular}

%% file: generated/tab_lkj.tex
\footnotesize
\begin{tabular}{rrrrrr}
\toprule
$k$ & $\eta$ & defined & drawn & within $2\times$, pair & within $2\times$, triple \\
\midrule
2 & 1 & 108 & 200 & 38\,\% & 49\,\% \\
2 & 4 & 165 & 200 & 39\,\% & 24\,\% \\
\bottomrule
\end{tabular}

%% file: generated/tab_conv.tex
\footnotesize
\begin{tabular}{lrrrrD{.}{.}{2}}
\toprule
variant & paired values & violations & median change & max change & \multicolumn{1}{c}{worst ratio} \\
\midrule
$M\times 2$ & 291{,}249 & 0 & \ensuremath{1.9\times 10^{-5}} & \ensuremath{4.6\times 10^{-1}} & 0.86 \\
$n\times 2$ & 291{,}258 & 5{,}460 & \ensuremath{4.6\times 10^{-5}} & \ensuremath{8.3\times 10^{0}} & 36.21 \\
$s$-nodes $\times 2$ & 291{,}276 & 0 & \ensuremath{2.7\times 10^{-9}} & \ensuremath{3.7\times 10^{-5}} & 0.00 \\
$x$-nodes $\times 2$ & 291{,}276 & 0 & \ensuremath{4.0\times 10^{-10}} & \ensuremath{6.4\times 10^{-6}} & 0.00 \\
\bottomrule
\end{tabular}

%% file: generated/tab_depth.tex
\footnotesize
\begin{tabular}{llrrrrrrrr}
\toprule
& & \multicolumn{4}{c}{$d_{\mathrm{total}}$ at layer} & \multicolumn{2}{c}{$\mathrm{Var}\,p_{\mathrm{G}}$} & \multicolumn{2}{c}{H5a} \\ \cmidrule(lr){3-6}\cmidrule(lr){7-8}\cmidrule(l){9-10}
belief & basis & $L=1$ & $L=2$ & $L=3$ & $L=4$ & $L=1$ & $L=4$ & ratio & above/below/all \\
\midrule
S2 & pair & 0.0648 & 0.0211 & 0.0071 & 0.0024 & 0.0713 & 0.0829 & 0.05 & 1/9/12 \\
 & triple & 0.0653 & 0.0213 & 0.0071 & 0.0025 & 0.0713 & 0.0829 & 0.05 & 1/9/12 \\
S3c & pair & 0.0516 & 0.0111 & 0.0043 & 0.0018 & 0.0642 & 0.0826 & 0.04 & 0/12/12 \\
 & triple & 0.0126 & 0.0071 & 0.0030 & 0.0015 & 0.0642 & 0.0826 & 0.10 & 0/12/12 \\
S4 & pair & 0.0005 & 0.0010 & 0.0010 & 0.0011 & 0.0833 & 0.0833 & 1.91 & 8/0/12 \\
 & triple & 0.0748 & 0.0252 & 0.0086 & 0.0031 & 0.0833 & 0.0833 & 0.04 & 1/9/12 \\
\bottomrule
\end{tabular}

%% file: generated/tab_training.tex
\footnotesize
\begin{tabular}{llrrrrrr}
\toprule
loss & optimiser & kept initial & coef.\ dist. & test loss & test loss down & $\Delta P(D<0)$ & $\Delta\min D$ \\
\midrule
value & Adam & 23\,\% & 0.132 & $-$0.3\,\% & 52\,\% & 0.005 & $-$0.47 \\
 & L-BFGS & 4\,\% & 0.131 & $-$1.1\,\% & 82\,\% & 0.000 & $-$0.34 \\
NLL & Adam & 0\,\% & 0.649 & $-$10.9\,\% & 100\,\% & $-$0.055 & $-$7.35 \\
 & L-BFGS & 0\,\% & 0.660 & $-$11.1\,\% & 100\,\% & $-$0.043 & $-$8.06 \\
end-to-end, $L=1$ & Adam & 3\,\% & 0.432 & $-$97.0\,\% & 96\,\% & 0.109 & $-$13.67 \\
 & L-BFGS & 5\,\% & 0.216 & $-$97.7\,\% & 94\,\% & 0.060 & $-$1.21 \\
end-to-end, $L=2$ & Adam & 10\,\% & 0.167 & $-$91.7\,\% & 88\,\% & 0.035 & $-$2.12 \\
 & L-BFGS & 9\,\% & 0.059 & $-$91.8\,\% & 88\,\% & 0.001 & $-$0.04 \\
end-to-end, $L=3$ & Adam & 19\,\% & 0.154 & $-$52.2\,\% & 71\,\% & 0.033 & $-$1.34 \\
 & L-BFGS & 18\,\% & 0.045 & $-$50.2\,\% & 71\,\% & 0.001 & $-$0.02 \\
end-to-end, $L=4$ & Adam & 54\,\% & 0.000 & 0.0\,\% & 31\,\% & 0.003 & 0.00 \\
 & L-BFGS & 50\,\% & 0.000 & 0.0\,\% & 33\,\% & 0.000 & 0.00 \\
\bottomrule
\end{tabular}

%% file: generated/tab_validation.tex
\footnotesize
\begin{tabular}{@{}lp{0.72\linewidth}l@{}}
\toprule
test & what it checks & result \\
\midrule
M1 & orthonormality of the basis, explicit $f_1,\dots,f_4$ & pass \\
M2 & index families, coefficient counts, block vs dense estimator & pass \\
M3 & estimator vs population moments, SE scaling, Gauss--Hermite convergence & pass \\
M4 & reproduces the TV values of the $d=3$ study & pass \\
M5 & flat-$D$ control: $p_{\mathrm{CD}}=p_{\mathrm{EU}}$ & pass \\
M6 & covariance identity in sample & pass \\
M7 & integrals of all densities; copula samplers and conditionals; Student-$t$ CDF and quantile against 50-digit mpmath & pass \\
M8 & TV in $s$ vs $x$; grid refinement with spiky $p_{\mathrm{G}}$ & pass \\
M9 & min-$D$ optimiser vs brute-force grids and closed forms & pass \\
M10 & positive definiteness of all correlation matrices; LKJ moments & pass \\
M11 & bitwise reproducibility; a run interrupted and resumed & pass \\
M12 & seed variance matches estimated SE; common random numbers & pass \\
M13 & convergence under doubling = hypothesis H4b, on the study's data & \textbf{fail} \\
M14 & chain: $L=1$ vs engine, closed form vs MC, autograd vs finite differences, teacher Markov property & pass \\
\bottomrule
\end{tabular}

%% file: generated/tab_energy_main.tex
\footnotesize
\begin{tabular}{lrD{.}{.}{4}D{.}{.}{4}D{.}{.}{4}D{.}{.}{4}}
\toprule
& & \multicolumn{2}{c}{float32} & \multicolumn{2}{c}{float64} \\ \cmidrule(lr){3-4}\cmidrule(l){5-6}
layer & parameters & \multicolumn{1}{c}{ms} & \multicolumn{1}{c}{J} & \multicolumn{1}{c}{ms} & \multicolumn{1}{c}{J} \\
\midrule
HCR-CD & 39{,}232 & 0.484 & 0.137 & 0.666 & 0.203 \\
HCR-CAL & 39{,}232 & 308 & 88.7 & 580 & 168 \\
MLP & 39{,}265 & 0.289 & 0.0682 & 0.297 & 0.105 \\
KAN & 39{,}528 & 1.91 & 0.482 & 3.32 & 0.829 \\
\bottomrule
\end{tabular}

%% file: generated/tab_energy_float32.tex
\scriptsize
\begin{tabular}{rrlrD{.}{.}{4}D{.}{.}{5}D{.}{.}{6}rrr}
\toprule
$k$ & batch & layer & params & \multicolumn{1}{c}{ms} & \multicolumn{1}{c}{J} & \multicolumn{1}{c}{J net} & W & MHz & counter \\
\midrule
2 & 256 & HCR-CAL & 3{,}904 & 4.52 & 1.24 & 0.983 & 275 & 1{,}410 & 0.997 \\
 &  & HCR-CD & 3{,}904 & 0.147 & 0.00891 & 0.000430 & 61 & 1{,}095 & 1.017 \\
 &  & KAN & 4{,}158 & 1.06 & 0.0655 & 0.00319 & 62 & 1{,}095 & 0.995 \\
 &  & MLP & 3{,}883 & 0.0657 & 0.00399 & 0.000234 & 62 & 1{,}095 & 0.997 \\
 & 16{,}384 & HCR-CAL & 3{,}904 & 285 & 82.5 & 66.1 & 290 & 1{,}410 & 1.003 \\
 &  & HCR-CD & 3{,}904 & 0.178 & 0.0457 & 0.0354 & 256 & 1{,}410 & 0.980 \\
 &  & KAN & 4{,}158 & 1.07 & 0.0946 & 0.0338 & 92 & 1{,}260 & 0.992 \\
 &  & MLP & 3{,}883 & 0.145 & 0.0165 & 0.00814 & 114 & 1{,}410 & 0.993 \\
8 & 256 & HCR-CAL & 39{,}232 & 4.87 & 1.37 & 1.09 & 283 & 1{,}410 & 1.009 \\
 &  & HCR-CD & 39{,}232 & 0.154 & 0.0113 & 0.00231 & 74 & 1{,}185 & 0.993 \\
 &  & KAN & 39{,}528 & 1.08 & 0.0688 & 0.00538 & 64 & 1{,}095 & 1.017 \\
 &  & MLP & 39{,}265 & 0.0685 & 0.00498 & 0.000946 & 73 & 1{,}170 & 0.998 \\
 & 16{,}384 & HCR-CAL & 39{,}232 & 308 & 88.7 & 71.0 & 289 & 1{,}410 & 0.992 \\
 &  & HCR-CD & 39{,}232 & 0.484 & 0.137 & 0.109 & 283 & 1{,}410 & 1.011 \\
 &  & KAN & 39{,}528 & 1.91 & 0.482 & 0.372 & 253 & 1{,}410 & 1.002 \\
 &  & MLP & 39{,}265 & 0.289 & 0.0682 & 0.0515 & 236 & 1{,}410 & 1.003 \\
32 & 256 & HCR-CAL & 549{,}184 & 10.4 & 3.13 & 2.53 & 300 & 1{,}410 & 1.003 \\
 &  & HCR-CD & 549{,}184 & 0.152 & 0.0342 & 0.0254 & 226 & 1{,}410 & 0.994 \\
 &  & KAN & 549{,}504 & 1.06 & 0.115 & 0.0528 & 108 & 1{,}410 & 0.988 \\
 &  & MLP & 549{,}181 & 0.0713 & 0.0127 & 0.00859 & 179 & 1{,}410 & 1.004 \\
 & 16{,}384 & HCR-CAL & 549{,}184 & 663 & 202 & 164 & 305 & 1{,}410 & 1.000 \\
 &  & HCR-CD & 549{,}184 & 5.93 & 1.89 & 1.55 & 318 & 1{,}410 & 1.004 \\
 &  & KAN & 549{,}504 & 16.9 & 4.77 & 3.79 & 282 & 1{,}410 & 0.996 \\
 &  & MLP & 549{,}181 & 1.93 & 0.682 & 0.571 & 353 & 1{,}410 & 1.009 \\
\bottomrule
\end{tabular}

%% file: generated/tab_energy_float64.tex
\scriptsize
\begin{tabular}{rrlrD{.}{.}{4}D{.}{.}{5}D{.}{.}{6}rrr}
\toprule
$k$ & batch & layer & params & \multicolumn{1}{c}{ms} & \multicolumn{1}{c}{J} & \multicolumn{1}{c}{J net} & W & MHz & counter \\
\midrule
2 & 256 & HCR-CAL & 3{,}904 & 8.76 & 2.49 & 1.97 & 284 & 1{,}410 & 1.000 \\
 &  & HCR-CD & 3{,}904 & 0.143 & 0.00908 & 0.000623 & 63 & 1{,}095 & 0.980 \\
 &  & KAN & 4{,}158 & 1.03 & 0.0628 & 0.00257 & 61 & 1{,}095 & 0.989 \\
 &  & MLP & 3{,}883 & 0.0693 & 0.00432 & 0.000278 & 63 & 1{,}095 & 0.989 \\
 & 16{,}384 & HCR-CAL & 3{,}904 & 557 & 159 & 126 & 285 & 1{,}410 & 0.989 \\
 &  & HCR-CD & 3{,}904 & 0.328 & 0.0830 & 0.0638 & 253 & 1{,}410 & 0.983 \\
 &  & KAN & 4{,}158 & 1.04 & 0.130 & 0.0697 & 126 & 1{,}410 & 0.984 \\
 &  & MLP & 3{,}883 & 0.0667 & 0.0124 & 0.00851 & 186 & 1{,}410 & 1.006 \\
8 & 256 & HCR-CAL & 39{,}232 & 9.11 & 2.56 & 2.03 & 281 & 1{,}410 & 0.983 \\
 &  & HCR-CD & 39{,}232 & 0.147 & 0.0111 & 0.00251 & 76 & 1{,}260 & 1.001 \\
 &  & KAN & 39{,}528 & 1.01 & 0.0653 & 0.00515 & 63 & 1{,}095 & 1.003 \\
 &  & MLP & 39{,}265 & 0.0691 & 0.00490 & 0.000862 & 71 & 1{,}095 & 0.990 \\
 & 16{,}384 & HCR-CAL & 39{,}232 & 580 & 168 & 134 & 289 & 1{,}410 & 0.989 \\
 &  & HCR-CD & 39{,}232 & 0.666 & 0.203 & 0.164 & 304 & 1{,}410 & 1.003 \\
 &  & KAN & 39{,}528 & 3.32 & 0.829 & 0.634 & 250 & 1{,}410 & 1.001 \\
 &  & MLP & 39{,}265 & 0.297 & 0.105 & 0.0879 & 355 & 1{,}410 & 1.005 \\
32 & 256 & HCR-CAL & 549{,}184 & 15.0 & 4.69 & 3.81 & 313 & 1{,}410 & 1.009 \\
 &  & HCR-CD & 549{,}184 & 0.241 & 0.0527 & 0.0386 & 219 & 1{,}410 & 0.996 \\
 &  & KAN & 549{,}504 & 1.03 & 0.146 & 0.0872 & 145 & 1{,}410 & 0.997 \\
 &  & MLP & 549{,}181 & 0.0837 & 0.0185 & 0.0136 & 223 & 1{,}410 & 1.006 \\
 & 16{,}384 & HCR-CAL & 549{,}184 & 955 & 300 & 243 & 313 & 1{,}410 & 0.993 \\
 &  & HCR-CD & 549{,}184 & 7.41 & 2.15 & 1.71 & 289 & 1{,}410 & 1.000 \\
 &  & KAN & 549{,}504 & 31.1 & 8.18 & 6.35 & 263 & 1{,}410 & 0.999 \\
 &  & MLP & 549{,}181 & 2.80 & 1.11 & 0.948 & 398 & 1{,}388 & 1.002 \\
\bottomrule
\end{tabular}

%% file: generated/tab_batches.tex
\footnotesize
\begin{tabular}{lrrrrrrr}
\toprule
experiment & runs & GPU-h & batch 1 & batch 2 & batch 3 & batch 4 & batch 5 \\
\midrule
Gaussian sweep & 1{,}000 & 66.0 & 6 & 137 & 857 & --- & --- \\
population sweep & 100 & 11.6 & --- & 1 & 99 & --- & --- \\
convergence study & 240 & 23.2 & --- & --- & 240 & --- & --- \\
copula families & 1{,}080 & 42.5 & --- & --- & 1 & 42 & 1{,}037 \\
LKJ matrices & 240 & 2.8 & --- & 2 & 24 & 1 & 213 \\
chains & 240 & 30.6 & --- & --- & --- & --- & 240 \\
\midrule
total & 2{,}900 & 176.7 & 6 & 140 & 1{,}221 & 43 & 1{,}490 \\
\bottomrule
\end{tabular}